\documentclass{article}
\usepackage{floatrow}
\newfloatcommand{capbtabbox}{table}[][\FBwidth]
\usepackage{microtype}
\usepackage{graphicx}
\usepackage{multicol}
\usepackage{subcaption}
\usepackage{booktabs}
\usepackage{arydshln}% for professional tables

\newcommand{\enc}{\mathbf{E}} 
\newcommand{\hyp}{\mathbf{H}} 
\newcommand{\dec}{\mathbf{I}} 

\usepackage[hidelinks]{hyperref}

\usepackage[accepted]{icml2023}

\usepackage{amsmath}
\usepackage{amssymb}
\usepackage{mathtools}
\usepackage{amsthm}
\usepackage{tikz} 
\usetikzlibrary{arrows.meta, calc, decorations.markings, math, arrows}

\usepackage[capitalize,noabbrev]{cleveref}

\theoremstyle{plain}

\theoremstyle{definition}

\theoremstyle{remark}

\usepackage[textsize=tiny]{todonotes}

\icmltitlerunning{Rotation and Translation Invariant Representation Learning with Implicit Neural Representations}

\begin{document}

\twocolumn[
\icmltitle{Rotation and Translation Invariant Representation Learning \\ with Implicit Neural Representations}




\begin{icmlauthorlist}
\icmlauthor{Sehyun Kwon}{ai}
\icmlauthor{Joo Young Choi}{math}
\icmlauthor{Ernest K. Ryu}{ai,math}
\end{icmlauthorlist}
% \icmlauthor{}

\icmlaffiliation{ai}{Interdisciplinary Program in Articifial Intelligence, Seoul National University}
\icmlaffiliation{math}{Department of Mathematical Sciences, Seoul National University}

\icmlcorrespondingauthor{Ernest K. Ryu}{ernestryu@snu.ac.kr}

\icmlkeywords{implicit neural representations,INR,representation learning,rotation invariance,translation invariance,disentanglement,disentangled representation}

\vskip 0.3in
]



\printAffiliationsAndNotice{}  % leave blank if no need to mention equal contribution
\begin{abstract}
In many computer vision applications, images are acquired with arbitrary or random rotations and translations, and in such setups, it is desirable to obtain semantic representations disentangled from the image orientation. Examples of such applications include semiconductor wafer defect inspection, plankton microscope images, and inference on single-particle cryo-electron microscopy (cryo-EM) micro-graphs. In this work, we propose Invariant Representation Learning with Implicit Neural Representation (IRL-INR), which uses an implicit neural representation (INR) with a hypernetwork to obtain semantic representations disentangled from the orientation of the image. We show that IRL-INR can effectively learn disentangled semantic representations on more complex images compared to those considered in prior works and show that these semantic representations synergize well with SCAN to produce state-of-the-art unsupervised clustering results. Code: \url{https://github.com/sehyunkwon/IRL-INR}.
\end{abstract}

\section{Introduction}
\label{introduction}

In many computer vision applications, images are acquired with arbitrary or random rotations and translations. Examples of such applications include semiconductor wafer defect inspection \cite{WANG20081914, wang_2019_wafer,wafer_dataaugmentation},  plankton microscope images \cite{5414357}, and inference on single-particle cryo-electron microscopy (cryo-EM) micrographs \cite{zhongCryoDRGNReconstructionHeterogeneous2021}. In such applications, the rotation and translation of images serve as \emph{nuisance parameters} \citep[\S7.3]{cox1979} that may interfere with the inference of the semantic meaning of the image. Therefore, it is desirable to obtain semantic representations that are not dependent on such nuisance parameters.

Obtaining low-dimensional ``disentangled'' representations is an active area of research in the area of representation learning. Prior works such as $\beta$-VAE \citep{betavae} and Info-GAN \citep{chen2016} propose general methods for disentangling latent representations so that components correspond to semantically independent factors. However, such fully general approaches are limited in the extent of disentanglement that they can accomplish. Alternatively, Spatial-VAE \citep{spatialvae} and TARGET-VAE \cite{targetvae} explicitly, and therefore much more effectively, disentangle nuisance parameters from the semantic representation using an encoder with a so-called spatial generator. However, we find that these prior methods are difficult to train on more complex datasets such as semiconductor wafer maps or plankton microscope images, as we demonstrate in Section~\ref{subsec:symm-break}. We also find that the learned representations do not synergize well with modern deep-learning-based unsupervised clustering methods, as we demonstrate in Section~\ref{subsec:clustering}.

In this work, we propose \emph{Invariant Representation Learning with Implicit Neural Representation} (IRL-INR), which uses an implicit neural representation (INR) with a hypernetwork to obtain semantic representations disentangled from the orientation of the image. Through our experiments, we show that IRL-INR can learn disentangled semantic representations on more complex images. We also show that these semantic representations synergize well with SCAN \cite{vangansbeke2020scan} to produce state-of-the-art clustering results. Finally, we show a scaling phenomenon in which the clustering performance improves as the dimension of the semantic representation increases. 

% \newpage
\section{Related Works} 
\label{sec:formatting}
%\sh{Terminology about Disentangled rotation vs Predicted rotation}
\paragraph{Disentangled representation learning.}
Finding disentangled latent representations corresponding to semantically independent factors is a classical problem in machine learning \cite{COMON1994287, HYVARINEN2000411, Shakunaga_2001_CVPR}. Recently, generative models have been used extensively for this task. DR-GAN \cite{Tran_2017_CVPR}, TC-$\beta$-VAE \cite{Chen_2018_NEURIPS}, DIP-VAE \cite{kumar_2018_ICLR}, Deformation Autoencoder \cite{Shu_2018_ECCV}, $\beta$-VAE \cite{betavae}, StyleGAN \cite{stylegan}, and \citet{Francesco_2020_ICLR} are prominent prior work finding disentangled representations of images. However these methods are post-hoc approaches that do not explicitly structure the latent space to separate the semantic representations from the known factors to be disentangled. In contrast, Spatial-VAE \cite{spatialvae} attempts to explicitly separate latent space into semantic representation of a image and its rotation and translation information, but only the generative part of spatial-VAE ends up being equivariant to rotation and translation. TARGET-VAE \cite{targetvae} is the first method to successfully disentangle rotation and translation information from the semantic representation in an explicit manner. However, we find that TARGET-VAE fails to obtain meaningful semantic representation of complex data such as semiconductor wafer maps and plankton image considered in Figure~\ref{fig:breaking sym}.

% \citet{Feng_2019_CVPR} outputs a decoupled semantic feature containing rotation related and unrelated parts. \citet{Uddin_2022_CVPR} simultaneously and explicitly disentangles semantic representation from multiple transformations like rotation, scaling and contrast. In gait recognition, \citet{Zhang2019GaitRV} and \cite{ Li_2020_CVPR} used reconstruction loss to disentangle semantic representation and covariate features such as carrying status, clothing, walking speed, and viewing angle.

% Also, generative models can be utilized. \citep{Tran_2017_CVPR, Francesco_2020_ICLR},TC-$\beta$-VAE \cite{Chen_2018_NEURIPS}, DIP-VAE \cite{kumar_2018_ICLR}, Deformation Autoencoder \cite{Shu_2018_ECCV}, $\beta$-VAE \cite{betavae} and StyleGAN \cite{stylegan} accomplished to disentangle some factors from the image, but these methods are post-hoc approaches, i.e, they can not choose specific factor, such as rotation or translation, to be disentangled. In contrast, Spatial-VAE \cite{spatialvae} attempted to explicitly disentangle semantic representation of a image from the rotation and translation information. However, in Spatial-VAE, disentangled values corresponding to rotation and translation were meaningless. TARGET-VAE \cite{targetvae} is the first method to obtain meaningful predicted rotation and translation and to achieve disentangling from semantic representation. However, TARGET-VAE fails to obtain meaningful semantic representation of complex data such as wafer map of semiconductor and plankton image.

\paragraph{Invariant representation learning.}
Recently, contrastive learning methods have been widely used to learn invariant representations \cite{Wang_2015_ICCV, Sermanet2017TCN,  wu2018unsupervised,Dwibedi_2019_CVPR,hjelm2018learning, he2019moco, Misra_2020_CVPR, chen_2020_simclr, Yeh_2022_ECCV}. Contrastive learning maximizes the similarity of positive samples generated by data augmentation and maximizes dissimilarity to negative samples. Since positive samples are defined by data augmentation such as rotation, translation, crop, color jitter and etc., contrastive learning forces data representations to be invariant under the designated data augmentation.

Siamese networks is another approach for learning invariant representation \cite{NIPS1993_288cc0ff}. The approach is to maximize the similarity between an image and its augmented image. Since only maximizing similarity may lead to a bad trivial solution, having an additional constraint is essential. For example, momentum encoder \cite{NEURIPS2020_f3ada80d}, stop gradient method \cite{Chen_2021_CVPR}, and reconstruction loss \cite{NIPS2011_45fbc6d3, Giancola_2019_CVPR, Zhou_siameserecon_2020_ECCV, Liu2020RelightingII} were used to avoid the trivial solution. Our IRL-INR methodology can be interpreted as an instance of the Siamese network that uses reconstruction loss as a constraint.

% XXXTargetVAE uses
% G-CNNs\cite{pmlr-v48-cohenc16}.
% XXX

\paragraph{Implicit neural representations.}
It is natural to view an image as a discrete and finite set of measurements of an underlying continuous signal or image. To model this view, \citet{cppn} proposed using a neural network to represent a function $f$ that can be evaluated at any input position $(x,y)$ as a substitute for the more conventional approach having a neural network to output a 2D array representing an image. The modern literature now refers to this approach as an \emph{Implicit Neural Representation} (INR).
For example, \citet{Dupont_2022_AISTATS, sitzmann2019srns, NEURIPS2020_53c04118} uses deep neural networks to parameterize images and uses hypernetworks to obtain the parameters of such neural networks representing a continuous image \cite{ha2017hypernetworks}.

Taking the coordinate as an input makes INR, by definition, symmetric or equivariant under rotation and translation. Leveraging the equivariant structure of INR, \citet{spatialvae, mildenhall2020nerf,  anokhin2020image, zhongCryoDRGNReconstructionHeterogeneous2021, Karras2021, Deng_2021_ICCV, chen2021learning, targetvae} proposed the generative networks that are equivariant under rotation or translation, and our method uses the equivariance property to learn invariant representations.

% Even though underlying signal of data is continuous, data has been traditionally represented by a discrete signal in practice. For example, in a 2D image is typically represented by a pixel array. This is because the underlying signal has ``infinite'' resolution, but it is viewed as a pixel array with ``finite'' resolution through measurement like camera.

\begin{figure*}

\tikzset{every picture/.style={line width=0.75pt}} %set default line width to 0.75pt        

\begin{tikzpicture}[x=0.75pt,y=0.75pt,yscale=-1,xscale=1]
%uncomment if require: \path (0,638); %set diagram left start at 0, and has height of 638

%Image [id:dp36094664519349107] 
\draw (110.5,174) node  {\includegraphics[width=36.75pt,height=37.5pt]{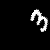}};
%Straight Lines [id:da48351678587254077] 
\draw  [dash pattern={on 0.84pt off 2.51pt}]  (52,173) -- (80,173) ;
\draw [shift={(83,173)}, rotate = 180] [fill={rgb, 255:red, 0; green, 0; blue, 0 }  ][line width=0.08]  [draw opacity=0] (5.36,-2.57) -- (0,0) -- (5.36,2.57) -- cycle    ;
%Straight Lines [id:da5381190519578207] 
\draw    (137,173) -- (157,173) ;
\draw [shift={(160,173)}, rotate = 180] [fill={rgb, 255:red, 0; green, 0; blue, 0 }  ][line width=0.08]  [draw opacity=0] (5.36,-2.57) -- (0,0) -- (5.36,2.57) -- cycle    ;
%Shape: Trapezoid [id:dp6933398854347409] 
\draw   (163,123) -- (223,141) -- (223,205) -- (163,223) -- cycle ;
%Straight Lines [id:da36174808681524095] 
\draw    (229,154) -- (257,154) ;
\draw [shift={(260,154)}, rotate = 180] [fill={rgb, 255:red, 0; green, 0; blue, 0 }  ][line width=0.08]  [draw opacity=0] (5.36,-2.57) -- (0,0) -- (5.36,2.57) -- cycle    ;
%Straight Lines [id:da13710175173779848] 
\draw    (229,193) -- (257,193) ;
\draw [shift={(260,193)}, rotate = 180] [fill={rgb, 255:red, 0; green, 0; blue, 0 }  ][line width=0.08]  [draw opacity=0] (5.36,-2.57) -- (0,0) -- (5.36,2.57) -- cycle    ;
%Straight Lines [id:da6763990160591609] 
\draw    (229,170) -- (251,170) -- (257,170) ;
\draw [shift={(260,170)}, rotate = 180] [fill={rgb, 255:red, 0; green, 0; blue, 0 }  ][line width=0.08]  [draw opacity=0] (5.36,-2.57) -- (0,0) -- (5.36,2.57) -- cycle    ;
%Straight Lines [id:da30538313471709533] 
\draw    (281,153) -- (281,54) ;
\draw [shift={(281,51)}, rotate = 90] [fill={rgb, 255:red, 0; green, 0; blue, 0 }  ][line width=0.08]  [draw opacity=0] (5.36,-2.57) -- (0,0) -- (5.36,2.57) -- cycle    ;
%Straight Lines [id:da7861366817587461] 
\draw    (275,153) -- (281,153) ;
%Straight Lines [id:da7657912611771184] 
\draw    (289,168) -- (289,54) ;
\draw [shift={(289,51)}, rotate = 90] [fill={rgb, 255:red, 0; green, 0; blue, 0 }  ][line width=0.08]  [draw opacity=0] (5.36,-2.57) -- (0,0) -- (5.36,2.57) -- cycle    ;
%Straight Lines [id:da6082345678471073] 
\draw    (275,168) -- (289,168) ;
%Straight Lines [id:da9508852027273481] 
\draw    (234,38) -- (263,38) ;
\draw [shift={(266,38)}, rotate = 180] [fill={rgb, 255:red, 0; green, 0; blue, 0 }  ][line width=0.08]  [draw opacity=0] (5.36,-2.57) -- (0,0) -- (5.36,2.57) -- cycle    ;
%Shape: Trapezoid [id:dp6125354363883906] 
\draw   (458,165) -- (439.85,225.5) -- (361.65,225.5) -- (343.5,165) -- cycle ;
%Straight Lines [id:da5038184043481221] 
\draw    (276,193) -- (347,193) ;
\draw [shift={(350,193)}, rotate = 180] [fill={rgb, 255:red, 0; green, 0; blue, 0 }  ][line width=0.08]  [draw opacity=0] (5.36,-2.57) -- (0,0) -- (5.36,2.57) -- cycle    ;
%Shape: Rectangle [id:dp6484774994965055] 
\draw   (362,0) -- (437,0) -- (437,90) -- (362,90) -- cycle ;
%Straight Lines [id:da7676104852060583] 
\draw    (335,38) -- (355,38) ;
\draw [shift={(358,38)}, rotate = 180] [fill={rgb, 255:red, 0; green, 0; blue, 0 }  ][line width=0.08]  [draw opacity=0] (5.36,-2.57) -- (0,0) -- (5.36,2.57) -- cycle    ;
%Image [id:dp6649595476896026] 
\draw (500,40) node  {\includegraphics[width=37.5pt,height=37.5pt]{image/J.png}};
%Shape: Brace [id:dp6098417311618172] 
\draw   (349.2,114.6) .. controls (349.2,119.27) and (351.53,121.6) .. (356.2,121.6) -- (389.6,121.6) .. controls (396.27,121.6) and (399.6,123.93) .. (399.6,128.6) .. controls (399.6,123.93) and (402.93,121.6) .. (409.6,121.6)(406.6,121.6) -- (443,121.6) .. controls (447.67,121.6) and (450,119.27) .. (450,114.6) ;
%Straight Lines [id:da34964773432151763] 
\draw    (400,161) -- (400,148) ;
\draw [shift={(400,145)}, rotate = 90] [fill={rgb, 255:red, 0; green, 0; blue, 0 }  ][line width=0.08]  [draw opacity=0] (5.36,-2.57) -- (0,0) -- (5.36,2.57) -- cycle    ;
%Shape: Circle [id:dp765703798326334] 
\draw  [color={rgb, 255:red, 208; green, 2; blue, 27 }  ,draw opacity=1 ] (344,106) .. controls (344,103.24) and (346.24,101) .. (349,101) .. controls (351.76,101) and (354,103.24) .. (354,106) .. controls (354,108.76) and (351.76,111) .. (349,111) .. controls (346.24,111) and (344,108.76) .. (344,106) -- cycle ;
%Shape: Circle [id:dp9683692324156519] 
\draw  [color={rgb, 255:red, 208; green, 2; blue, 27 }  ,draw opacity=1 ] (357,106) .. controls (357,103.24) and (359.24,101) .. (362,101) .. controls (364.76,101) and (367,103.24) .. (367,106) .. controls (367,108.76) and (364.76,111) .. (362,111) .. controls (359.24,111) and (357,108.76) .. (357,106) -- cycle ;
%Shape: Circle [id:dp3423680464348078] 
\draw  [color={rgb, 255:red, 208; green, 2; blue, 27 }  ,draw opacity=1 ] (371,106) .. controls (371,103.24) and (373.24,101) .. (376,101) .. controls (378.76,101) and (381,103.24) .. (381,106) .. controls (381,108.76) and (378.76,111) .. (376,111) .. controls (373.24,111) and (371,108.76) .. (371,106) -- cycle ;
%Shape: Circle [id:dp8639408354878916] 
\draw  [color={rgb, 255:red, 74; green, 144; blue, 226 }  ,draw opacity=1 ] (386,106) .. controls (386,103.24) and (388.24,101) .. (391,101) .. controls (393.76,101) and (396,103.24) .. (396,106) .. controls (396,108.76) and (393.76,111) .. (391,111) .. controls (388.24,111) and (386,108.76) .. (386,106) -- cycle ;
%Shape: Circle [id:dp06273770720341043] 
\draw  [color={rgb, 255:red, 74; green, 144; blue, 226 }  ,draw opacity=1 ] (401,106) .. controls (401,103.24) and (403.24,101) .. (406,101) .. controls (408.76,101) and (411,103.24) .. (411,106) .. controls (411,108.76) and (408.76,111) .. (406,111) .. controls (403.24,111) and (401,108.76) .. (401,106) -- cycle ;
%Shape: Circle [id:dp28945740594827984] 
\draw  [color={rgb, 255:red, 208; green, 2; blue, 27 }  ,draw opacity=1 ] (376,41) .. controls (376,38.24) and (378.24,36) .. (381,36) .. controls (383.76,36) and (386,38.24) .. (386,41) .. controls (386,43.76) and (383.76,46) .. (381,46) .. controls (378.24,46) and (376,43.76) .. (376,41) -- cycle ;
%Shape: Circle [id:dp8954968028605682] 
\draw  [color={rgb, 255:red, 208; green, 2; blue, 27 }  ,draw opacity=1 ] (376,57) .. controls (376,54.24) and (378.24,52) .. (381,52) .. controls (383.76,52) and (386,54.24) .. (386,57) .. controls (386,59.76) and (383.76,62) .. (381,62) .. controls (378.24,62) and (376,59.76) .. (376,57) -- cycle ;
%Shape: Circle [id:dp5484771141058965] 
\draw  [color={rgb, 255:red, 208; green, 2; blue, 27 }  ,draw opacity=1 ] (376,72) .. controls (376,69.24) and (378.24,67) .. (381,67) .. controls (383.76,67) and (386,69.24) .. (386,72) .. controls (386,74.76) and (383.76,77) .. (381,77) .. controls (378.24,77) and (376,74.76) .. (376,72) -- cycle ;
%Shape: Circle [id:dp8962157523481737] 
\draw  [color={rgb, 255:red, 74; green, 144; blue, 226 }  ,draw opacity=1 ] (416,106) .. controls (416,103.24) and (418.24,101) .. (421,101) .. controls (423.76,101) and (426,103.24) .. (426,106) .. controls (426,108.76) and (423.76,111) .. (421,111) .. controls (418.24,111) and (416,108.76) .. (416,106) -- cycle ;
%Shape: Circle [id:dp22388666326728834] 
\draw  [color={rgb, 255:red, 65; green, 117; blue, 5 }  ,draw opacity=1 ] (431,106) .. controls (431,103.24) and (433.24,101) .. (436,101) .. controls (438.76,101) and (441,103.24) .. (441,106) .. controls (441,108.76) and (438.76,111) .. (436,111) .. controls (433.24,111) and (431,108.76) .. (431,106) -- cycle ;
%Shape: Circle [id:dp10565607404415023] 
\draw  [color={rgb, 255:red, 65; green, 117; blue, 5 }  ,draw opacity=1 ] (446,106) .. controls (446,103.24) and (448.24,101) .. (451,101) .. controls (453.76,101) and (456,103.24) .. (456,106) .. controls (456,108.76) and (453.76,111) .. (451,111) .. controls (448.24,111) and (446,108.76) .. (446,106) -- cycle ;
%Shape: Circle [id:dp3380843088771198] 
\draw  [color={rgb, 255:red, 74; green, 144; blue, 226 }  ,draw opacity=1 ] (396,41) .. controls (396,38.24) and (398.24,36) .. (401,36) .. controls (403.76,36) and (406,38.24) .. (406,41) .. controls (406,43.76) and (403.76,46) .. (401,46) .. controls (398.24,46) and (396,43.76) .. (396,41) -- cycle ;
%Shape: Circle [id:dp22287424214790597] 
\draw  [color={rgb, 255:red, 74; green, 144; blue, 226 }  ,draw opacity=1 ] (396,57) .. controls (396,54.24) and (398.24,52) .. (401,52) .. controls (403.76,52) and (406,54.24) .. (406,57) .. controls (406,59.76) and (403.76,62) .. (401,62) .. controls (398.24,62) and (396,59.76) .. (396,57) -- cycle ;
%Shape: Circle [id:dp21151024201483437] 
\draw  [color={rgb, 255:red, 74; green, 144; blue, 226 }  ,draw opacity=1 ] (396,72) .. controls (396,69.24) and (398.24,67) .. (401,67) .. controls (403.76,67) and (406,69.24) .. (406,72) .. controls (406,74.76) and (403.76,77) .. (401,77) .. controls (398.24,77) and (396,74.76) .. (396,72) -- cycle ;
%Shape: Circle [id:dp28333452248089797] 
\draw  [color={rgb, 255:red, 65; green, 117; blue, 5 }  ,draw opacity=1 ] (416,46) .. controls (416,43.24) and (418.24,41) .. (421,41) .. controls (423.76,41) and (426,43.24) .. (426,46) .. controls (426,48.76) and (423.76,51) .. (421,51) .. controls (418.24,51) and (416,48.76) .. (416,46) -- cycle ;
%Shape: Circle [id:dp2110134099533797] 
\draw  [color={rgb, 255:red, 65; green, 117; blue, 5 }  ,draw opacity=1 ] (416,66) .. controls (416,63.24) and (418.24,61) .. (421,61) .. controls (423.76,61) and (426,63.24) .. (426,66) .. controls (426,68.76) and (423.76,71) .. (421,71) .. controls (418.24,71) and (416,68.76) .. (416,66) -- cycle ;
%Straight Lines [id:da8628859873131804] 
\draw    (386,41) -- (396,71) ;
%Straight Lines [id:da07738481760942673] 
\draw    (386,71) -- (396,41) ;
%Straight Lines [id:da15674543076471048] 
\draw    (386,57) -- (396,41) ;
%Straight Lines [id:da557691718873829] 
\draw    (386,71) -- (396,57) ;
%Straight Lines [id:da15689034470719831] 
\draw    (386,71) -- (396,71) ;
%Straight Lines [id:da9152793495623923] 
\draw    (386,57) -- (396,57) ;
%Straight Lines [id:da8895419102507541] 
\draw    (386,41) -- (396,41) ;
%Straight Lines [id:da5603304253404606] 
\draw    (396,71) -- (386,57) ;
%Straight Lines [id:da8201576551785246] 
\draw    (396,56) -- (386,42) ;
%Straight Lines [id:da8200627129169666] 
\draw    (416,66) -- (406,41) ;
%Straight Lines [id:da4700737595190887] 
\draw    (416,66) -- (406,56) ;
%Straight Lines [id:da15402528076133504] 
\draw    (416,66) -- (406,71) ;
%Straight Lines [id:da04016989600829057] 
\draw    (416,46) -- (406,41) ;
%Straight Lines [id:da41403161958488877] 
\draw    (406,56) -- (416,46) ;
%Straight Lines [id:da12848380695749184] 
\draw    (416,46) -- (406,71) ;
%Curve Lines [id:da39366718975494897] 
\draw [color={rgb, 255:red, 208; green, 2; blue, 27 }  ,draw opacity=1 ]   (360,94) .. controls (345.6,93.52) and (345.48,58.02) .. (367.17,55.18) ;
\draw [shift={(370,55)}, rotate = 180] [fill={rgb, 255:red, 208; green, 2; blue, 27 }  ,fill opacity=1 ][line width=0.08]  [draw opacity=0] (5.36,-2.57) -- (0,0) -- (5.36,2.57) -- cycle    ;
%Straight Lines [id:da9962199227105886] 
\draw [color={rgb, 255:red, 74; green, 144; blue, 226 }  ,draw opacity=1 ]   (401,98) -- (401,82) ;
\draw [shift={(401,79)}, rotate = 90] [fill={rgb, 255:red, 74; green, 144; blue, 226 }  ,fill opacity=1 ][line width=0.08]  [draw opacity=0] (5.36,-2.57) -- (0,0) -- (5.36,2.57) -- cycle    ;
%Curve Lines [id:da014988115275912373] 
%Curve Lines [id:da014988115275912373] 
\draw [color={rgb, 255:red, 65; green, 117; blue, 5 }  ,draw opacity=1 ]   (440,94) .. controls (453.92,93.04) and (454.94,57.52) .. (432.88,55.12) ;
\draw [shift={(430,55)}, rotate = 358.85] [fill={rgb, 255:red, 65; green, 117; blue, 5 }  ,fill opacity=1 ][line width=0.08]  [draw opacity=0] (5.36,-2.57) -- (0,0) -- (5.36,2.57) -- cycle    ;
%Image [id:dp06715216262310708] 
\draw (25,40) node  {\includegraphics[width=37.5pt,height=37.5pt]{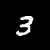}};
%Straight Lines [id:da20752232802203552] 
\draw  [dash pattern={on 0.84pt off 2.51pt}]  (25,90) -- (25,142) ;
\draw [shift={(25,145)}, rotate = 270] [fill={rgb, 255:red, 0; green, 0; blue, 0 }  ][line width=0.08]  [draw opacity=0] (5.36,-2.57) -- (0,0) -- (5.36,2.57) -- cycle    ;
%Shape: Grid [id:dp5926366725817469] 
\draw  [draw opacity=0] (86,149) -- (136,149) -- (136,198.33) -- (86,198.33) -- cycle ; \draw  [color={rgb, 255:red, 255; green, 255; blue, 255 }  ,draw opacity=1 ] (86,149) -- (86,198.33)(96,149) -- (96,198.33)(106,149) -- (106,198.33)(116,149) -- (116,198.33)(126,149) -- (126,198.33) ; \draw  [color={rgb, 255:red, 255; green, 255; blue, 255 }  ,draw opacity=1 ] (86,149) -- (136,149)(86,159) -- (136,159)(86,169) -- (136,169)(86,179) -- (136,179)(86,189) -- (136,189) ; \draw  [color={rgb, 255:red, 255; green, 255; blue, 255 }  ,draw opacity=1 ]  ;
%Shape: Grid [id:dp16171381974120214] 
\draw  [draw opacity=0] (475,15) -- (525,15) -- (525,64.33) -- (475,64.33) -- cycle ; \draw  [color={rgb, 255:red, 255; green, 255; blue, 255 }  ,draw opacity=1 ] (475,15) -- (475,64.33)(485,15) -- (485,64.33)(495,15) -- (495,64.33)(505,15) -- (505,64.33)(515,15) -- (515,64.33) ; \draw  [color={rgb, 255:red, 255; green, 255; blue, 255 }  ,draw opacity=1 ] (475,15) -- (525,15)(475,25) -- (525,25)(475,35) -- (525,35)(475,45) -- (525,45)(475,55) -- (525,55) ; \draw  [color={rgb, 255:red, 255; green, 255; blue, 255 }  ,draw opacity=1 ]  ;
%Image [id:dp9213887313206829] 
\draw (25.5,175) node  {\includegraphics[width=36.75pt,height=37.5pt]{image/J.png}};
%Shape: Ellipse [id:dp9386365533780958] 
\draw  [fill={rgb, 255:red, 0; green, 0; blue, 0 }  ,fill opacity=1 ] (135.1,38.73) .. controls (135.1,37.59) and (135.99,36.67) .. (137.08,36.67) .. controls (138.17,36.67) and (139.05,37.59) .. (139.05,38.73) .. controls (139.05,39.86) and (138.17,40.78) .. (137.08,40.78) .. controls (135.99,40.78) and (135.1,39.86) .. (135.1,38.73) -- cycle ;
%Curve Lines [id:da8106363615038371] 
\draw    (141.05,34.67) .. controls (167.66,16.6) and (192.61,16.85) .. (203.16,27.78) ;
\draw [shift={(205,30)}, rotate = 234.75] [fill={rgb, 255:red, 0; green, 0; blue, 0 }  ][line width=0.08]  [draw opacity=0] (5.36,-2.57) -- (0,0) -- (5.36,2.57) -- cycle    ;
%Shape: Grid [id:dp5800930116279384] 
\draw  [draw opacity=1] (117.08,18.73) -- (167.08,18.73) -- (167.08,68.06) -- (117.08,68.06) -- cycle ; \draw  [color={rgb, 255:red, 0; green, 0; blue, 0 }  ,draw opacity=1 ] (127.08,18.73) -- (127.08,68.06)(137.08,18.73) -- (137.08,68.06)(147.08,18.73) -- (147.08,68.06)(157.08,18.73) -- (157.08,68.06) ; \draw  [color={rgb, 255:red, 0; green, 0; blue, 0 }  ,draw opacity=1 ] (117.08,28.73) -- (167.08,28.73)(117.08,38.73) -- (167.08,38.73)(117.08,48.73) -- (167.08,48.73)(117.08,58.73) -- (167.08,58.73) ; \draw  [color={rgb, 255:red, 0; green, 0; blue, 0 }  ,draw opacity=1 ]  ;

% Text Node
\draw (21,72.4) node [anchor=north west][inner sep=0.75pt]  [font=\footnotesize]  {$\mathcal{I}$};
% Text Node
\draw (107,203.4) node [anchor=north west][inner sep=0.75pt]  [font=\footnotesize]  {$J$};
% Text Node
\draw (170,157) node [anchor=north west][inner sep=0.75pt]  [font=\footnotesize] [align=left] {Encoder};
% Text Node
\draw (184,175.4) node [anchor=north west][inner sep=0.75pt]  [font=\footnotesize]  {$\mathbf{E}_{\phi }$};
% Text Node
\draw (263,146.4) node [anchor=north west][inner sep=0.75pt]  [font=\scriptsize]  {$\hat{\theta }$};
% Text Node
\draw (263,161.4) node [anchor=north west][inner sep=0.75pt]  [font=\scriptsize]  {$\hat{\tau }$};
% Text Node
\draw (263,187.4) node [anchor=north west][inner sep=0.75pt]  [font=\footnotesize]  {$z$};
% Text Node
\draw (269,30.4) node [anchor=north west][inner sep=0.75pt]  [font=\footnotesize]  {$S_{\hat{\theta },\hat{\tau }}( x_{p} ,\ y_{p})$};
% Text Node
\draw (189,30.4) node [anchor=north west][inner sep=0.75pt]  [font=\footnotesize]  {$( x_{p} ,\ y_{p})$};
% Text Node
\draw (366,176) node [anchor=north west][inner sep=0.75pt]  [font=\footnotesize] [align=left] {Hypernetwork};
% Text Node
\draw (394,198.4) node [anchor=north west][inner sep=0.75pt]  [font=\footnotesize]  {$\mathbf{H}_{\psi }$};
% Text Node
\draw (366,3) node [anchor=north west][inner sep=0.75pt]  [font=\footnotesize] [align=left] {INR network};
% Text Node
\draw (379,17.4) node [anchor=north west][inner sep=0.75pt]  [font=\footnotesize]  {$\mathbf{I}( \cdot ,\cdot ;\eta )$};
% Text Node
\draw (449,32.4) node [anchor=north west][inner sep=0.75pt]  [font=\footnotesize]  {$\simeq $};
% Text Node
\draw (395,131) node [anchor=north west][inner sep=0.75pt]  [font=\footnotesize]  {$\eta $};
% Text Node
\draw (61,157.4) node [anchor=north west][inner sep=0.75pt]  [font=\scriptsize]  {$M$};
% Text Node
\draw (31,107.4) node [anchor=north west][inner sep=0.75pt]  [font=\scriptsize]  {$T \circ R$};

\draw (490,70.4)  ;

\end{tikzpicture}

  \caption{The IRL-INR framework. Encoder $\enc_\phi$ takes an image $J$ as input and outputs rotation representation $\hat{\theta}$, translation representation $\hat{\tau}$ and semantic representation $z$. Hypernetwork $\hyp_\psi$ takes $z$ as an input and then outputs the weights and biases of INR network. INR network $\dec$ outputs the pixel (image) value corresponding to the input $(x,y)$ coordinate.}
  \label{fig:main-architecture}

\end{figure*}

\paragraph{Deep clustering.}
Representation learning plays an essential role in modern deep clustering. Many deep-learning-based clustering methods utilize a \emph{pretext task} to extract a clustering-friendly representation. Early methods such as \citet{Tian_2014_AAAI, xie2016} used the auto-encoder to learn low-dimensional representation space and directly clustered on this obtained representation space. Later, \citet{Ji_2017_deepsubspace, Zhou_2018_CVPR, Zhang_2021_learningaselfexpressive} proposed a subspace representation learning as a pretext task, where images are well separated by mapping into a suitable low-dimensional subspace. More recently, \citet{vangansbeke2020scan, Dang_2021_CVPR, li2021contrastive, shen2021you} established state-of-the-art performance on many clustering benchmarks by utilizing contrastive learning-based pretext tasks such as SimCLR \cite{chen_2020_simclr} or MOCO \cite{he2019moco}. However, none of the pretext tasks considered in prior work explicitly take into account rotation and translation invariant clustering.

% \begin{figure*}[htb!]
% % \raisebox{-4.5cm}
%   \includegraphics[width=\linewidth,height=8cm]{./figure/figure1_temp.png}
%   \caption{Overview of IRL-INR}
%   \label{fig:main-architecture}
% \end{figure*}

% In our methodology, we also experimentally found that auto-eocnder based method fails to reconstruct sparse image such as semiconductor's wafer map, but INR reconstructs well. This property is crucial for real world image dataset. Second, since INR is a function takes coordinate as an input, it will automatically be equivariant, i.e, if we rotate the input coordinates then obtained pixel array is also rotated. So if we know the rotation degree $\theta$ of the image, then we can explicitly apply this $\theta$ in our training dynamics by rotating input coordintates by $-\theta$ to rotationally invaraint learning. 

\section{Method}
%-------------------------------------------------------------------------

Our method \emph{Invariant Representation Learning with Implicit Neural Representation} (IRL-INR) obtains a representation that disentangles the semantic representation from the rotation and translation of the image, using an implicit neural representation (INR) with a hypernetwork. Our main framework is illustrated in Figure~\ref{fig:main-architecture}, and we describe the details below.

% \paragraph{Domain Problems}
% In many real world applications such as wafer map of semiconductor clustering, plankton images classification and single-particle cryo-electron microscopy (cryo-EM) micro-graphs, rotation and translation are nuisance factors which are irrelevant to their class labels. \jy{However, naviely obtained representations are entangeld with such nuisance factors.} To solve this problem, \citet{Rodriguez_plankton_architecture} used neural networks with stronger inductive bias to be rotationally invariant. \citet{} reduced the effect of rotation by using data-augmenation. 

% \jy{Above references seem weak, many be remove the last two sentences and merge the next paragraph?} 

% In recent, representation learning method has been arised. There are three main approaches: (\lowercase\expandafter{\romannumeral1}) \textit{disentangled representation learning} and (\lowercase\expandafter{\romannumeral2}) \textit{invariant representation learning}.

\subsection{Data and its measurement model}

Our data $J^{(1)},\dots,J^{(N)}$ are images with resolution $P$ (number of pixels) and $C$ color channels. In the applications we consider, $C=1$ or $C=3$.
We index the images with the spatial indices reshaped into a single dimension, so that $J^{(i)}\in \mathbb{R}^{C\times P}$ and
\[
J^{(i)}_p\in\mathbb{R}^C,\qquad p=1,\dots,P
\]
for $i=1,\dots,N$. We assume $J^{(i)}$ represents measurements of a true underlying continuous image $\mathcal{I}^{(i)}$ that has been randomly rotated and translated for $i=1,\dots,N$. We further detail our measurement model below.

We assume there exist continuous 2-dimensional images $\mathcal{I}^{(1)},\dots,\mathcal{I}^{(N)}$ 
(so $\mathcal{I}^{(i)}(x,y)\in\mathbb{R}^C$ for any $x,y\in\mathbb{R}$) .
% that are rotated and translated to a canonical orientation\sh{(canonical orientation?)}. 
We observe/measure a randomly rotated and translated version of $\mathcal{I}^{(1)},\dots,\mathcal{I}^{(N)}$ on a discretized finite grid, to obtain $J^{(1)},\dots, J^{(N)}$. Mathematically, we write
\[
J^{(i)}=M[T_{\tau^{(i)}}[R_{\theta^{(i)}}[\mathcal{I}^{(i)}]]],\qquad i=1,\dots,N,
\]
where $R_{\theta^{(i)}}$ denotes rotation by angle $\theta^{(i)}\in [0,2\pi)$, 
$T_{\tau^{(i)}}$ denotes translation by direction $\tau^{(i)}\in \mathbb{R}^2$,
and $M$ is a measurement operator that measures a continuous image on a finite grid. 
More specifically, given a continuous image $\tilde{\mathcal{I}}$, the measurement $M[\tilde{\mathcal{I}}]$ is a finite image
\[
(M[\tilde{\mathcal{I}}])_{p}=\tilde{\mathcal{I}}(x_p,y_p)\in \mathbb{R}^C,\qquad p=1,\dots,P
\]
with a pre-specified set of gridpoints $\{(x_p,y_p)\}_{p=1}^P$, which we take to be a uniform grid on $[-1,1]^2$.
Throughout this work, we assume that $\theta^{(1)},\dots,\theta^{(N)}\stackrel{\text{IID}}{\sim}\mathrm{Uniform}([0,2\pi])$, i.e., that the rotations sampled  uniformly at random, and that translations $\tau^{(1)},\dots,\tau^{(N)}$ are sampled IID from some distribution.

To clarify, we do not have access to the true underlying continuous images $\mathcal{I}^{(1)},\dots,\mathcal{I}^{(N)}$, so we do not use them on our  framework. Also, the rotation $\theta^{(i)}$ and translation $\tau^{(i)}$ of $\mathcal{I}^{(i)}$ that produced the observed image $J^{(i)}$ for $i=1,\dots,N$ are impossible to learn without additional supervision, so we do not attempt to learn it.

% $\sim\mathrm{Uniform}([0,2\pi])$,
% where $\theta^{(1)},\dots,\theta^{(N)}$ and $\tau^{(1)},\dots,\tau^{(N)}$ are random rotation and translations, 

\subsection{Implicit neural representation with a hypernetwork}

Our framework takes in, as input, a discrete image $J$, which we assume originates from a true underlying continuous image $\mathcal{I}$. The framework, as illustrated in Figure~\ref{fig:main-architecture}, uses 
the rotation and translation operators $R_\theta$ and $T_\tau $ and three neural networks $\enc_\phi$, $\hyp_\psi$, and $\dec$. 

Define the rotation operation $R_\theta$ and translation operation $T_\tau$ on points and images as follows.
For notational convenience, define $S_{\theta,\tau}=R_\theta\circ T_\tau $.
When translating and rotating a point in $\mathbb{R}^2$, define $S_{\theta,\tau}$ as
\[
S_{\theta,\tau}(x,y)=
\begin{bmatrix}
\cos{\theta}&-\sin{\theta}\\
\sin{\theta}&\cos{\theta}
\end{bmatrix}
\left(\begin{bmatrix}
x\\y
\end{bmatrix}+\tau\right)\in \mathbb{R}^2.
\]
For rotating and translating a continuous image $\mathcal{I}$, define
\[
S_{\theta,\tau}^{-1}[\mathcal{I}](x,y)=
\mathcal{I}\left(
S_{\theta,\tau}(x,y)
\right),
% =
% \mathcal{I}\left(
% XXXR_\theta^{-1}(T_\tau^{-1}(x,y))
% \right).
\]
where $S_{\theta,\tau}^{-1}=T_\tau^{-1}\circ R_\theta^{-1}=T_{-\tau}\circ R_{-\theta}$.
For rotating and translating a discrete image $J$, we use an analogous formula with nearest neighbor interpolation.

% we use the \verb|rotate| and \verb|XXX| function of \verb|torchvision| with nearest neighbor interpolation.

The encoder network
% $\enc_\phi \colon \mathbb{R}^{I}\rightarrow \mathbb{R}^d\times\mathbb{R}\times \mathbb{R}$
% defined as
\[
\enc_\phi(J)=(z,{\hat{\theta}},{\hat{\tau}})
\in\mathbb{R}^{d}\times\mathbb{R}\times\mathbb{R}^2,
\]
where $J$ is an input image and $\phi$ is a trainable parameter, is trained such that the \emph{semantic representation} $z\in\mathbb{R}^d$ captures a representation of $\mathcal{I}$ disentangled from the arbitrary orientation $J$ is presented in.

The rotation representation  $\hat{\theta}\in[0,2\pi)$ and translation representation $\hat{\tau}\in\mathbb{R}^2$ are trained to be estimates of the rotation and translation with respect to a certain \emph{canonical orientation}. Specifically, given an image $J$ and its canonical orientation $J^\text{(can)}$, we define $(\hat{\theta},\hat{\tau})$ such that
% such that $\enc_\phi(J)=(z,\hat{\tau},\hat{\theta})$ 
\[
J^\text{(can)}=S_{\hat{\theta},\hat{\tau}}[J],
\]
and the equivariance property \eqref{eq:equivariance} that we soon discuss implies that
\[
\enc_\phi(J^\text{(can)})=(z,0,0).
\]
This canonical orientation $J^\text{(can)}$ is not (cannot be) the orientation of  $\mathcal{I}$. Rather, it is an orientation that we designate through the symmetry braking technique that we soon describe in Section~\ref{subsec:symm-break}.

% We call this the \emph{canonical orientation} and we define it precisely as follows.
% Given an image $J$ such that $\enc_\phi(J)=(z,\hat{\tau},\hat{\theta})$, define
% \[
% J^\text{(can)}=S_{\hat{\theta},\hat{\tau}}^{-1}[J]
% \]
% to be $J$ in its canonical orientation.

% is learned arbitrarily as in \citep{targetvae} or it is an orientation that we designate through the symmetry braking technique that we soon describe in Section~\ref{subsec:symm-break}.

% Later in Section~\ref{subsec:scale-z}, we discuss how $d$, the dimension of the semantic representation $z$, is an important hyperparameter.

The hypernetwork has the form
\[
\hyp_\psi(z)=\eta,
\]
where the semantic representation $z\in\mathbb{R}^d$ is the input and $\psi$ is a trainable parameter.
(Notably, $\hat{\theta}$ and $\hat{\tau}$ are not inputs.)
The output $\hyp_\psi(z)=\eta=(w_1,b_1,w_2,b_2,\dots,w_k,b_k)$ will be used as the weights and biases of the $k$ layers of the INR network, to be defined soon. We train the hypernetwork so that the INR network produces a continuous image representation approximating $\mathcal{I}$.
% In a sense, both $z$ and $\eta$ are representations of the underyling continuous-image $\mathcal{I}$.

% We train $\hyp_\psi$ so that is a representation of the image $J$ (and by extension the true underyling continuous image $\mathcal{I}$) disentangled from its particular orientation.

The implicit neural representation (INR) network has the form
\[
\dec(x,y;\eta)\in \mathbb{R}^C,
\]
where $x,y\in\mathbb{R}$ and $\eta$ is the output of the hypernetwork.
The IRL-INR framework is trained so that 
\[
\dec(\cdot,\cdot;\eta^{(i)})\approx \mathcal{I}^{(i)}(\cdot,\cdot)
\]
in some sense, where $\eta^{(i)}$ is produced by $\hyp_\psi$ and $\enc_\phi$ with $J^{(i)}$ provided as input.
% for $i=1,\dots,N$ in some sense
More specifically, we view $\dec(x,y;\eta)$ as a continuous 2-dimensional image with inputs $(x,y)$ and fixed parameter $\eta$, and we want $\dec(x,y;\eta^{(i)})$ and $\mathcal{I}^{(i)}(x,y)$ to be the same image in a different orientation.
The INR network is a deep neural network (specifically, we use an MLP), but it has no trainable parameters as its weights and biases $\eta$ are generated by the hypernetwork $\hyp_\psi(z)$.

% Throughout training, the INR network will be used as
% \[
% \dec(S_{\hat{\theta},\hat{\tau}}^{-1}(x_p,y_p)
% ;
% \eta)
% \]
% with a pre-specified set of gridpoints $\{(x_p,y_p)\}_{p=1}^P$,

% this structure is , by construction, equivariant in the sense that 

% XXX designed to be equivariant XXX

% More specifically, we 
% $ J_p\approx \dec(\tilde{x}_p,\tilde{y}_p;\eta)$

\subsection{Reconstruction and consistency losses}
\label{subsec:recon-consist}
We train IRL-INR with the loss
\[
\mathcal{L}(\phi,\psi)=  \lambda_\text{recon}\, \mathcal{L}_{\text{recon}}
+
\lambda_{\text{consis}} \, \mathcal{L}_{\text{consis}}
+
\lambda_{\text{symm}} \, \mathcal{L}_{\text{symm}},
\]
where $\lambda_\text{recon}>0$, $ \lambda_{\text{consis}}>0$, and $\lambda_{\text{symm}}>0$.
We define $\mathcal{L}_{\text{recon}}$ and $\mathcal{L}_{\text{consis}} $ in this section and define $\mathcal{L}_{\text{symm}}$ in Section~\ref{subsec:symm-break}.

\subsubsection{Reconstruction Loss}
We use the reconstruction loss
\[
\mathcal{L}_{\text{recon}}(\phi,\psi) = \mathbb{E}_{J}[\hat{\mathcal{L}}_{\text{recon}}(J;\phi,\psi)],
\]
with the per-image loss  $\hat{\mathcal{L}}_{\text{recon}}(J;\phi,\psi)$
defined as
\begin{align*}
(z,\hat{\theta},\hat{\tau})&=\enc_\phi(J)\\
\eta&= \hyp_\psi(z)\\
(\tilde{x}_p,\tilde{y}_p)&=
S_{\hat{\theta},\hat{\tau}}(x_p,y_p),\qquad p=1,\dots,P\\
\hat{\mathcal{L}}_{\text{recon}}(J;\phi,\psi) &=
\frac{1}{P}\sum\limits_{p=1}^P\left[ \left\| J_p- \dec(\tilde{x}_p,\tilde{y}_p;\eta) \right\|^2 \right].
\end{align*}
Given an image $J$ and its canonical orientation $J^\text{(can)}$, minimizing the reconstruction loss induces $J_p\approx \dec(\tilde{x}_p,\tilde{y}_p;\eta) $, which is roughly equivalent to $J^\text{(can)}_p\approx \dec(x_p,y_p;\eta) $ for $p=1,\dots,P$.
This requires the latent representation $(z,\hat{\theta},\hat{\tau})=\enc_\phi(J)$ to contain sufficient information about $J$ so that $\hyp_\psi$ and $\dec$ are capable of reconstructing $J$.
This is a similar role as those served by the reconstruction losses of autoencoders and VAEs.

% If the reconstruction loss is minimized, the joint representation $(z,\hat{\theta},\hat{\tau})$ should encode sufficient information to describe the image $J$. 

We believe that the INR structure already carries a significant inductive bias that promotes disentanglement between the semantic representation and the orientation information $(\hat{\theta},\hat{\tau})$. However, it is still possible that the same image in different orientations produces different semantic representations $z$ while still producing the same reconstruction. (Two different latent vectors can produce the same reconstructed image in autoencoders and INRs.) Therefore, we use an additional consistency loss to further enforce disentanglement between the semantic representation and the orientation of the image.

% We try to make it always the same latent vector regardless of the rotation of the input image. However, if only consistency loss is used, a phenomenon in which the latent vector, which is inferred rotation invariant for each image, collapses into one point may occur. To prevent this phenomenon, that is, to make the latent vector discriminative according to the input image, we made the rotationally invariant vector to generate the reference image (\jy{What is reference image? Perhaps explain the reference degree prior to this section?}). We put the rotationally invariant vector into HyperNetwork, and HyperNetwork generates the parameters of the INR. Let the image created by reshaping the values created by putting coordinates in the generated INR as input is called I hat. We will make the I hat the reference image at this time. At this time, we will use a simple trick to apply the predicted rotation information.

% INR is equivariant by definition because it receives coordinates as input and outputs pixel values at each coordinate. If an image is generated by rotating the input coordinate by the predicted angle theta hat using this point, the I hat should be the same as the input image. Therefore, if the input coordinates that are not rotated are inserted into the INR after learning is completed, a reference image is created. (\jy{math symbols missing in several places})

\subsubsection{Consistency Loss}
We use the consistency loss
\[
\mathcal{L}_{\text{consis}}(\phi) = \mathbb{E}_{J}[\hat{\mathcal{L}}_{\text{consis}}(J;\phi)]
\]
with the per-image loss $\hat{\mathcal{L}}_{\text{consis}}(J;\phi)$ is defined as
\begin{align*}
\tau_1,\tau_2&\sim\mathcal{N}(0,\sigma^2I_2)\\
\theta_1,\theta_2&\sim\mathrm{Uniform}([0,2\pi])\\
(z_i,\hat{\theta}_i,\hat{\tau}_i)&=\enc_\phi(
S_{\theta_i,\tau_i}[J]),\qquad i=1,2\\
% (z_2,\hat{\theta}_2,\hat{\tau}_2)&=\enc_\phi(I_{\theta_2})\\
% \eta&= \hyp_\psi(z)\\
% &=
% \\
    \hat{\mathcal{L}}_{\text{consis}}(J;\phi) &= 
1- \frac{z_1 \cdot z_2}{\|z_1\| \, \|z_2\|}.
\end{align*}
Note that this is the cosine similarity between $z_1$ and $z_2$.
% To clarify, $z_1 \cdot z_2$ denotes the inner product between $z_1$ and $z_2$.
Since $S_{\theta_1,\tau_1}[J]$ and $S_{\theta_2,\tau_2}[J]$ are also measurements of the same underlying continuous image $\mathcal{I}$, minimizing this consistency loss enforces $\enc_\phi$ to produce the same semantic representation $z$ regardless of the orientation in which $J$ is provided.
(Of course, $\enc_\phi$ produces different $\hat{\theta}$ and $\hat{\tau}$ depending on the orientation of $J$.)

It is possible to use other distance measures, such as the MSE loss, instead of the cosine similarity in measuring the discrepancy between $z_1$ and $z_2$. However, we found that the cosine similarity distance synergized well with the SCAN-based clustering of Section~\ref{subsec:clustering}.

\paragraph{Equivariance of encoder.}
Minimizing the reconstruction and consistency losses induces the following equivariance property.
If $\enc_\phi(J)=(z,\hat{\tau},\hat{\theta})$, then
\begin{equation}
\enc_\phi(S_{\theta,\tau}[J])\approx (z,R_{\hat{\theta}-\theta}[\hat{\tau}]-\tau,\hat{\theta}-\theta)
\label{eq:equivariance}
\end{equation}
for all $\tau\in \mathbb{R}^2$ and $\theta\in [0,2\pi)$,
where $\hat{\theta}-\theta\in [0,2\pi)$ should be understood in the sense of modulo $2\pi$.
In other words, rotating $J$ by $\theta$ will subtract $\theta$ to the rotation predicted by $\enc_\phi$.
For translation, the rotation effect must be taken into account.
To see why, note that 
minimizing the consistency loss enforces $\enc_\phi(J)$ and $\enc_\phi(S_{\theta,\tau}[J])$ to produce an (approximately) equal semantic representation $z$, and therefore, the corresponding $\eta=\hyp_\psi(z)$ will be (approximately) equal.
Minimizing the reconstruction loss implies
\begin{align*}
0&\stackrel{\text{(a)}}{\approx}
\hat{\mathcal{L}}_{\text{recon}}(J;\phi,\psi) \\
% &=
% \frac{1}{P}\sum\limits_{p=1}^P\left[ \left\| J_p- \dec(\tilde{x}_p,\tilde{y}_p
% ;
% \eta) \right\|^2 \right]\\
&=
\frac{1}{P}\sum\limits_{p=1}^P\left[ \left\| J_p- \dec(
S_{\hat{\theta},\hat{\tau}}[(x_p,y_p)];\eta) \right\|^2 \right]\\
&\stackrel{\text{(b)}}{\approx}
\frac{1}{P}\sum\limits_{p=1}^P\left[ \left\| 
(S_{\theta,\tau}[J])_p- 
\dec(S_{\hat{\theta}-\theta,R_{\hat{\theta}-\theta}[\hat{\tau}]-\tau}[(x_p,y_p)];\eta) \right\|^2 \right],\\
&\stackrel{\text{(c)}}{\approx}
\hat{\mathcal{L}}_{\text{recon}}(S_{\theta,\tau}[J];\phi,\psi)\stackrel{\text{(a)}}{\approx} 0.
\end{align*}
Steps (a) holds since the recontruction loss is minimized.
Step (b) holds since if two images are similar, then their rotated and translated versions are also similar.
More precisely, let $J'$ be the discrete image defined as $J'_p=\dec(S_{\hat{\theta},\hat{\tau}}[(x_p,y_p)];\eta)$ for $p=1,\dots P$. If $J\approx J'$, then $S_{\theta,\tau}[J]\approx S_{\theta,\tau}[J']$.
Furthermore, 
\begin{align*}
(S_{\theta,\tau}[J'])_p
&\stackrel{\text{(d)}}{\approx}
\dec(S_{\theta,\tau}^{-1}S_{\hat{\theta},\hat{\tau}}[(x_p,y_p)];\eta)\\
&\stackrel{\text{(d)}}{\approx}
\dec(S_{\hat{\theta}-\theta,R_{\hat{\theta}-\theta}[\hat{\tau}]-\tau}[(x_p,y_p)];\eta),
\end{align*}
where we the approximation of (d) captures interpolation artifacts.
Step (c) holds since the left-hand-side and the right-hand-side are both approximately $0$.
Finally, the fact that (c) holds implies that the equivariance property \eqref{eq:equivariance} holds.

\subsection{Symmetry-breaking loss}
\label{subsec:symm-break}
Our assumed data measurement model is symmetric/invariant with respect to the group of rotations and translations.
More specifically, let $\mathcal{G}$ be the group generated by rotations and translations, then for any image $J$ and $g\in \mathcal{G}$,
the images 
\[
J,\qquad\tilde{J}=g[J]
\]
are equally likely observations and carry exactly the same information about the true underlying continuous image $\mathcal{I}$.%
\footnote{We point out two technicalities in this statement. First, strictly speaking, $J$ and $\tilde{J}$ do not have \emph{exactly} the same pixel values due to interpolation artifacts, except when the translation and rotation exactly aligns with the image grid. Second, the invariance with respect to translation holds only if $\tau$ has a uniform prior on $\mathbb{R}^2$, which is an improper prior. On the other hand, the rotation group is compact and we do assume the rotation is uniformly distributed on $[0,2\pi)$, which is a proper prior.}
However, our framework inevitably decides on a canonical orientation as it disentangles the semantic representation $z$ from the orientation information $(\hat{\theta},\hat{\tau})$, such that input image $J$ and its canonical orientation $J^\text{(can)}$ satisfy
\[
J^\text{(can)}=S_{\hat{\theta},\hat{\tau}}[J],\qquad \enc_\phi(J^\text{(can)})\approx (z,0,0).
\]

% \begin{figure}[]
%   \centering
%     \includegraphics[width=.9\linewidth]{./figure/figure2.png}
%     \caption{Methods without symmetry fail to reconstruct WM811k and WHOI-Plankton images.}
% \end{figure}

\begin{figure}
\centering
        \begin{subfigure}[b]{0.23\linewidth}
                \centering
                \includegraphics[width=.6\linewidth]{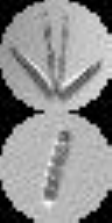}\\ [5mm]
                \includegraphics[width=.6\linewidth]{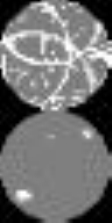}
                \captionsetup{font=scriptsize,labelformat=empty}
                \caption{Ground Truth}

        \end{subfigure}%
        \begin{subfigure}[b]{0.23\linewidth}
              \centering
                \includegraphics[width=.6\linewidth]{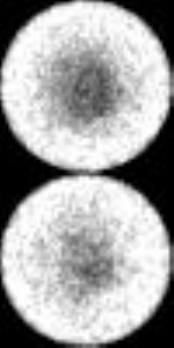}\\ [5mm]
                \includegraphics[width=.6\linewidth]{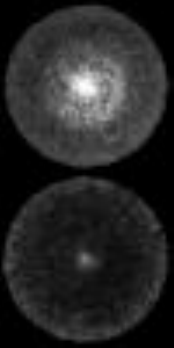}
                \captionsetup{font=scriptsize, labelformat=empty}
                \caption{TARGET-VAE}                
        \end{subfigure}%
        \begin{subfigure}[b]{0.23\linewidth}
              \centering
                \includegraphics[width=.6\linewidth]{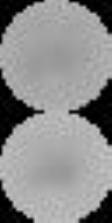}\\ [5mm]
                \includegraphics[width=.6\linewidth]{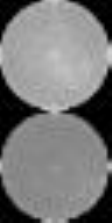}
                \captionsetup{font=scriptsize,labelformat=empty}
                \caption{IRL-INR w/o s.b.}
        \end{subfigure}%
        % \cdashline
        % \hspace{0.025\linewidth}
        % \vline
        % \hspace{0.025\linewidth}
        \begin{subfigure}[b]{0.23\linewidth}
                \centering
                \includegraphics[width=.6\linewidth]{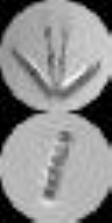}\\ [5mm]
                \includegraphics[width=.6\linewidth]{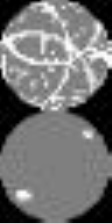}
                \captionsetup{font=scriptsize,labelformat=empty}
                \caption{IRL-INR w/ s.b.}

        \end{subfigure}%
        % \hspace{0.06\linewidth}
        % \hspace{0.06\linewidth}
        % \begin{subfigure}[b]{0.25\textwidth}
        %         \centering
        %         \includegraphics[width=.85\linewidth]{example-image-a} \\ [-0.3mm]
        %         \includegraphics[width=.85\linewidth]{example-image-b} \\ [2mm]
        %         \includegraphics[width=.85\linewidth]{example-image-a} \\ [-0.3mm]
        %         \includegraphics[width=.85\linewidth]{example-image-b}
        %         \caption{}
        % \end{subfigure}%
        % \begin{subfigure}[b]{0.25\textwidth}
        %         \centering
        %         \includegraphics[width=.85\linewidth]{example-image-a} \\ [-0.3mm]
        %         \includegraphics[width=.85\linewidth]{example-image-b} \\ [2mm]
        %         \includegraphics[width=.85\linewidth]{example-image-a} \\ [-0.3mm]
        %         \includegraphics[width=.85\linewidth]{example-image-b}
        %         \caption{}
        % \end{subfigure}%
        % \begin{subfigure}[b]{0.25\textwidth}
        %         \centering
        %         \includegraphics[width=.85\linewidth]{example-image-a} \\ [-0.3mm]
        %         \includegraphics[width=.85\linewidth]{example-image-b} \\ [2mm]
        %         \includegraphics[width=.85\linewidth]{example-image-a} \\ [-0.3mm]
        %         \includegraphics[width=.85\linewidth]{example-image-b}
        %         \caption{}
        % \end{subfigure}
        \caption{Methods without symmetry fail to reconstruct WM811k and WHOI-Plankton images.}\label{fig:breaking sym}
\end{figure}

\begin{figure*}[h!]
\centering
        \begin{subfigure}[b]{0.5\linewidth}
                \centering
                \includegraphics[width=.95\linewidth]{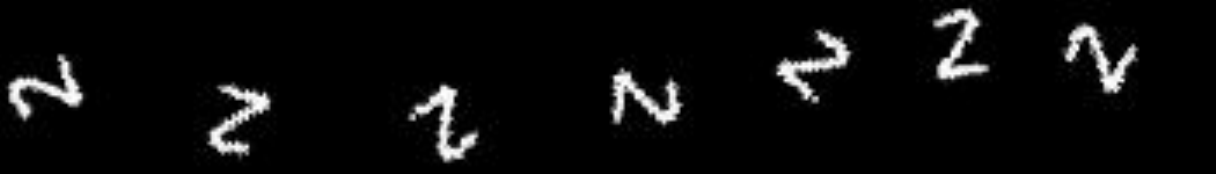}\\ [.5mm]
                \includegraphics[width=.95\linewidth]{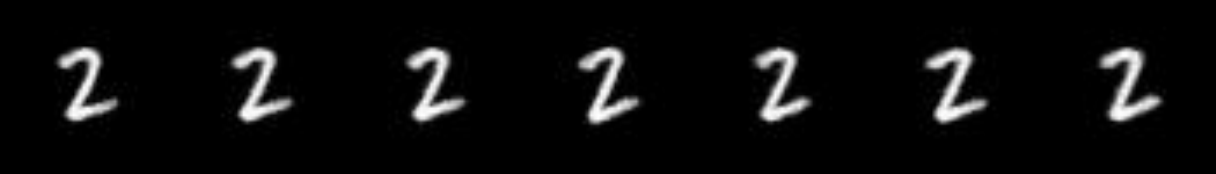}
                % \captionsetup{font=scriptsize}
                \caption{MNIST (U)}
        \end{subfigure}%
        \begin{subfigure}[b]{0.5\linewidth}
                \centering
                \includegraphics[width=.95\linewidth]{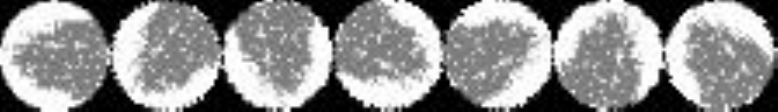}\\ [.5mm]
                \includegraphics[width=.95\linewidth]{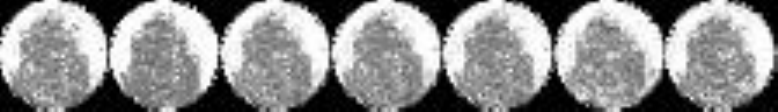}
                % \captionsetup{font=scriptsize}
                \caption{WM811K}
        \end{subfigure}%
        \\ [1mm]
        \begin{subfigure}[b]{0.5\linewidth}
                \centering
                \includegraphics[width=.95\linewidth]{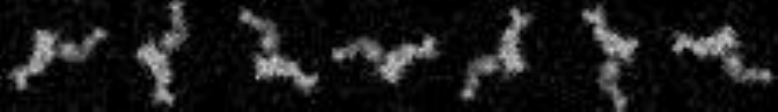}\\ [.5mm]
                \includegraphics[width=.95\linewidth]{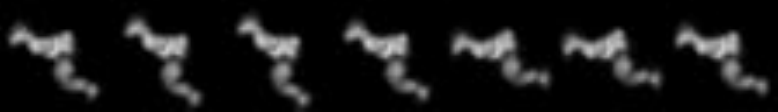}
                % \captionsetup{font=scriptsize}
                \caption{5HDB}
        \end{subfigure}%
        \begin{subfigure}[b]{0.5\linewidth}
                \centering
                \includegraphics[width=.95\linewidth]{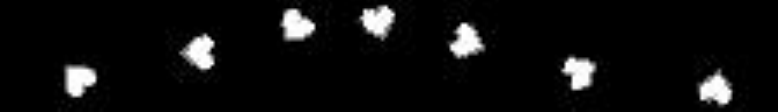}\\ [.5mm]
                \includegraphics[width=.95\linewidth]{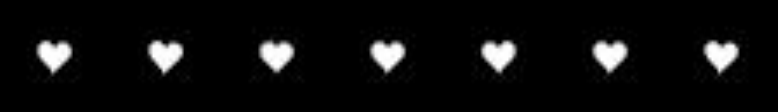}
                % \captionsetup{font=scriptsize}
                \caption{dSprites}
        \end{subfigure}%
        \\ [1mm]
        \begin{subfigure}[b]{0.5\linewidth}
                \centering
                \includegraphics[width=.95\linewidth]{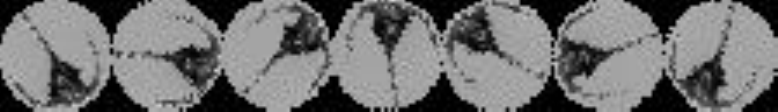}\\ [.5mm]
                \includegraphics[width=.95\linewidth]{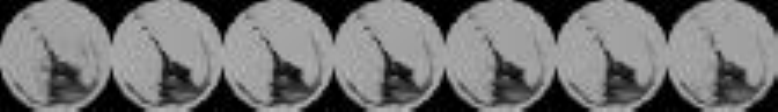}
                % \captionsetup{font=scriptsize}
                \caption{WHOI-Plankton}
        \end{subfigure}%
        \begin{subfigure}[b]{0.5\linewidth}
                \centering
                \includegraphics[width=.95\linewidth]{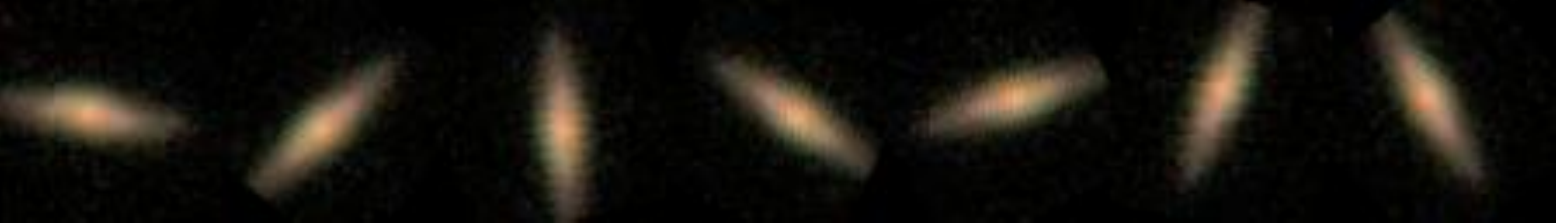}\\ [.5mm]
                \includegraphics[width=.95\linewidth]{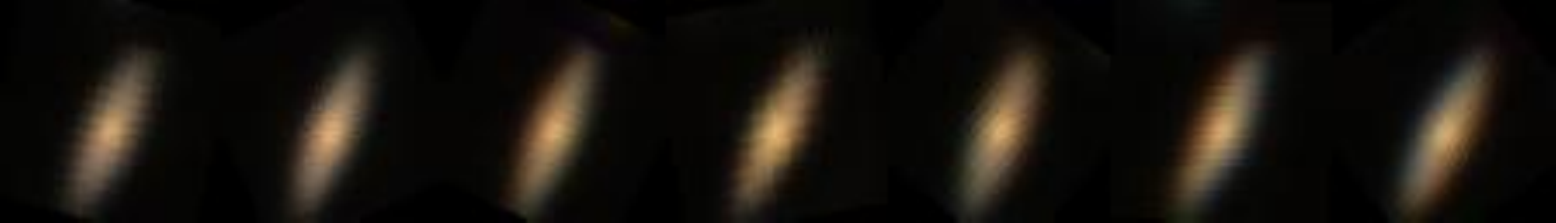}
                % \captionsetup{font=scriptsize}
                \caption{Galaxy Zoo}
        \end{subfigure}%
        \caption{To validate the disentanglement of semantic representations, we verify that the reconstructions are indeed invariant under rotation and translation. The first row of (a)--(f) are rotated by $\frac{2\pi}{7}$ degrees. The second row of (a)--(f) are reconstructions using only the semantic representation $z$, without any rotation or translation. We see that the reconstructions are invariant with respect to the rotations and translations. The setup is further detailed in \cref{Appendix: Reconstruct J can} and more images are provided in \cref{Appendix:Reconstruct J can image}.}
        \label{fig:recon images}
\end{figure*}

This canonical orientation is not orientation of the true underlying continuous image $\mathcal{I}$. The prior work of \citet{spatialvae,targetvae} allows the canonical orientation to be determined by the neural networks and their training process. Some of the datasets used in \citet{targetvae} are fully rotationally symmetric (such as ``MNIST(U)'') and for those setups, the symmetry makes the determination of the  canonical orientation an arbitrary choice.
We find that if we \emph{break the symmetry} by manually prescribing a rule for the canonical orientation, the trainability of the framework significantly improves as we soon demonstrate.

We propose a symmetry breaking based on the center of mass of the image. Given a continuous image $\mathcal{I}$, we define its center of mass as
\[
(m_x,m_y)=\frac{1}{\|\mathcal{I}\|_1}\int^\infty_{-\infty}\int^\infty_{-\infty} (x,y) \|\mathcal{I}(x,y)\|_1\;dxdy\in \mathbb{R}^2
\]
where $\|\mathcal{I}\|_1=\int^\infty_{-\infty}\int^\infty_{-\infty} |\mathcal{I}(x,y)|\;dxdy$ is $L_1$ norm. 
For a discrete image $J$, we use an analogous discretized formula.
Given an image $J$ with center of mass $m=(m_x,m_y)$, let $\tau=-m$ and let
$\theta\in [0,2\pi)$ such that
\[
m=\|\tau\|(\cos\theta,-\sin\theta).
\]
Then $J^\text{(can)}=S_{\hat{\theta},\hat{\tau}}[J]$ and $J^\text{(can)}$ has its center of mass at $(0,0)$.

%  \begin{align*}
%      \theta := \frac{m_y}{\sqrt{m_x^2+m_y^2}}
%  \end{align*}
% where $(m_x,m_y)$ is a center of mass of $I$. 

We use the symmetry-breaking loss
\[
   \mathcal{L}_\text{symm} (\phi)=
   \mathbb{E}_{J}[    \hat{\mathcal{L}}_\text{symm} (J;\phi)]
\]
with the per-image loss $\hat{\mathcal{L}}_\text{symm} (J;\phi)$ is defined as
\begin{align*}
(z,\hat{\tau},\hat{\theta})&=\enc_\phi(J)\\
m&=\text{CoM}(J)\\
   \hat{\mathcal{L}}_\text{symm} (J;\phi)&=
    \left\|
    m-\|\hat{\tau}\|(\cos\hat{\theta},-\sin\hat{\theta}) \right\|^2,
\end{align*}
where $\text{CoM}(J)$ denotes the center of mass of $J$.

The use of an INR with a hypernetwork is essential in directly enforcing the representation to disentangled while allowing the network to be sufficiently expressive to be able to learn sufficiently complex tasks. Specifically, we show in Figure~\ref{fig:breaking sym} that we could not train TARGET-VAE and IRL-INR to reconstruct the WM811k dataset without using the symmetry breaking technique.

% In this section, we conducted an ablation study to confirm the importance of symmetry breaking, as shown . IRL-INR w/o s.b was trained with the rotation prediction loss removed. That is, the encoder no longer disentangles rotation and semantics representation. The encoder only outputs a semantics representation. This scheme is thought to be more difficult to learn semantics representation because the encoder must be forced to automatically remove rotation information. This experiment demonstrates that breaking the symmetry in our method is essential to guide what information needs to be specifically disentangled. TARGET-VAE, which does not perform symmetry breaking, also failed to reconstruct in datasets such as semiconductors and plankton.
% symmetry breaking이 필요한 이유는 어떤 정보가 remove되어야 하는지 disentangling을 통해 encoder에게 guide역할을 하기 때문이다. 만약 symmetry breaking이 없으면,(즉, rotation-translation prediction loss가 없으면) encoder가 '스스로' rotation-translation information을 latent vector로부터 remove해야 하는데, 이것은 더 어려울 수 있기 때문이다

%------------------------------------------------------------------------

\section{Experiments}

\subsection{Experimental setup}
The encoder network $\enc_\phi(J)$ uses the ResNet18 architecture \cite{he2016} with an MLP as the head. The hypernetwork $\hyp_\psi(z)$ is an  MLP with input dimension $d$. The INR network $\dec(x,y;\eta)$ uses a random Fourier feature (RFF) encoding in the style of \citep{Rahimi_randomfeature,tancik2020} followed by an MLP with output dimension 1 (for grayscale images) or 3 (for rgb images). The architectures for $\hyp_\psi(z)$ and $\dec(x,y;\eta)$ are inspired by \citet{Dupont_2022_AISTATS}. Further details of the architecture can be found in Appendix~\ref{sec:architecture_detail} or the code provided as supplementary materials.

We use the Adam optimizer with learning rate $1\times 10^{-4}$, weight decay $5\times 10^{-4}$, and batch size $128$.
For the loss function scaling coefficients, we use $\lambda_\text{recon}=\lambda_\text{consis}=1$ and $\lambda_\text{symm}=15$. We use the MSE and cosine similarity distances for the consistency loss for our results of Section~\ref{sec:reconstruction} and Section~\ref{subsec:clustering}, respectively.

We evaluate the performance of IRL-INR against the recent prior work TARGET-VAE \citep{targetvae}. For TARGET-VAE experiments, we mostly use the code and settings provided by the authors. 
We use the TARGET-VAE with $P_{16}$ and $d=32$, which  \citet{targetvae} report to perform the best for the clustering. For more complex datasets, such as WM811K or WHOI-PLANKTON, we increase the number of layers from $2$ to $6$ in their ``spatial generator'' network, as the authors did for the cryo-EM dataset. 
For the clustering experiments of Sections~\cref{subsec:clustering} and \cref{subsec:scale-z}, we separate the training set and test set and evaluate the accuracy on the test set.

% We do not train and evaluate on the full datasets and evaluate clustering performance on only test set. Also, we provide mean and standard deviation measured over 5 runs. 

\subsubsection{Datasets}

% \paragraph{MNIST(U).}
MNIST(U) is derived from MNIST with random rotations and translations respectively sampled from $\mathrm{Uniform}([0, 2\pi))$ and $\mathcal{N}(0, 5^2)$. To accommodate the translations, we embed the images into $50\times 50$ pixels as was done in \citep{targetvae}.

% However maybe we should clarify the fact that we use 50x50 following the convention of Target-VAE

% train and test set of MNIST(U) have 60,000, and 10,000 images of dimensions 50x50 pixels, respectively. (\jy{Number of samples might not be necessary to explain. However maybe we should clarify the fact that we use 50x50 following the convention of Target-VAE})

% \paragraph{WM811K}
% WM811K XXXWafer inspection is very important for increasing the yield of a micro/nano-fabrication process in the semiconductor industry because it is possible to figure out the root causes of various process issues based on different kinds of detected wafer map failure patterns. It is important to obtain a reduced representation by obtaining features that are rotationally invariant.

WM811k is a dataset of silicon wafer maps classified into 9 defect patterns \cite{Wu2015WaferMF}. The wafer maps are circular, and the semiconductor fabrication process makes the data rotationally invariant. Because the original full dataset has a severe class imbalance, we distill the dataset into 7350 training set and 3557 test set images with reasonably balanced 9 defect classes and resize the images to $32\times 32$ pixels.

% However, the imbalance between classes is severe, with the smallest class being 134. Therefore, in order to minimize the imbalance between classes, we chose to use a small number of images with a large number of classes and to use all images with a small number of classes. (\jy{we randomly choose subset of classes with large number of samples to compensate the class imbalance?}) 

% image dataset containing 811,457 

% we chose to use a small number of images with a large number of classes and to use all images with a small number of classes. (\jy{as in the WM811K, we randomly choose subset of classes with large number of samples to compensate the imbalance.}) 
% (\jy{3.4 million with 70classes is reduced to 1200 with 10 classes? The explanation does not justify this difference.})

% \paragraph{5HDB.}
5HDB consists of 20,000 simulated projections of integrin $\alpha_{\text{IIb}}$ in complex with integrin $\beta_3$ \cite{Lin2015SubunitBI, spatialvae} with varying orientations.
There are 16,000 training set and 4000 test set images of $40\times 40$ pixels.

% \paragraph{dSprites.}
dSprites consists of 2D shapes procedurally generated from 6 ground truth independent latent factors \cite{betavae}.
% These factors are color(white), shape, scale, rotation, x and y positions of a sprite.
All possible combinations of these latents are present exactly once, resulting in 737,280 total images.

% is the largest public dataset comprising 9-labeled  . 
% \paragraph{WHOI-Plankton.}
WHOI-Plankton is an expert-labeled image dataset for plankton \cite{Orenstein2015WHOIPlanktonAL}. The orientation of the plankton with respect to the microscope is random, so the dataset exhibits rotation and translation invariance. However, the original dataset has a severe class imbalance, so we distill the dataset into 10 balanced classes with 1000 training set and 200 test set images. We also perform a circular crop and resize the images to $32\times32$ pixels.

% \paragraph{Galaxy Zoo.}
Galaxy Zoo consists of 61,578 RGB color images of galaxies from the Sloan Digital Sky Survey \cite{lintott2008}. Each image is cropped and downsampled to $64\times 64$ pixels following common practice \cite{10.1093/mnras/stv632}. We divide into 50,000 training set and 11,578 test set images.

\subsection{Validating disentanglement}
\label{sec:reconstruction}
In this section, we validate whether the encoder network $\enc_\phi(J)=(z,\hat{\theta},\hat{\tau})$ is indeed successfully trained to produce a semantic representation $z$ disentangled from the orientation of the input image $J$.

Figure~\ref{fig:recon images} shows images and their reconstructions with MNIST(U), WM811k, 5HDB, dSprites, WHOI-Plankton, and Galaxy Zoo datasets.
The first row of each subfigure shows images that have been rotated or translated from a given image, and we compute $\enc_\phi(J)=(z,\hat{\theta},\hat{\tau})$.
% Encoder receives input images as the first row and outputs invariant latent vectors. 
The second row of each figure is the reconstruction of these images by the disentangled semantic representation $z$. More specifically, the reconstructions correspond to $\dec(x,y;\hyp_\psi(z))$ with $(x,y)$ \emph{not} rotated or translated. (So the $(\hat{\theta},\hat{\tau})$ output by $\enc_\phi(J)$ is not used.) We can see that the reconstruction is indeed (approximately) invariant regardless of the orientation of the input image $J$. For comparison, TARGET-VAE was unable to learn representations from the WM811k and WHOI-Plankton datasets, as discussed in Section~\ref{subsec:symm-break}.

\begin{figure}[h]
\centering
\begin{minipage}{.48\linewidth}
    \includegraphics[width=40mm]{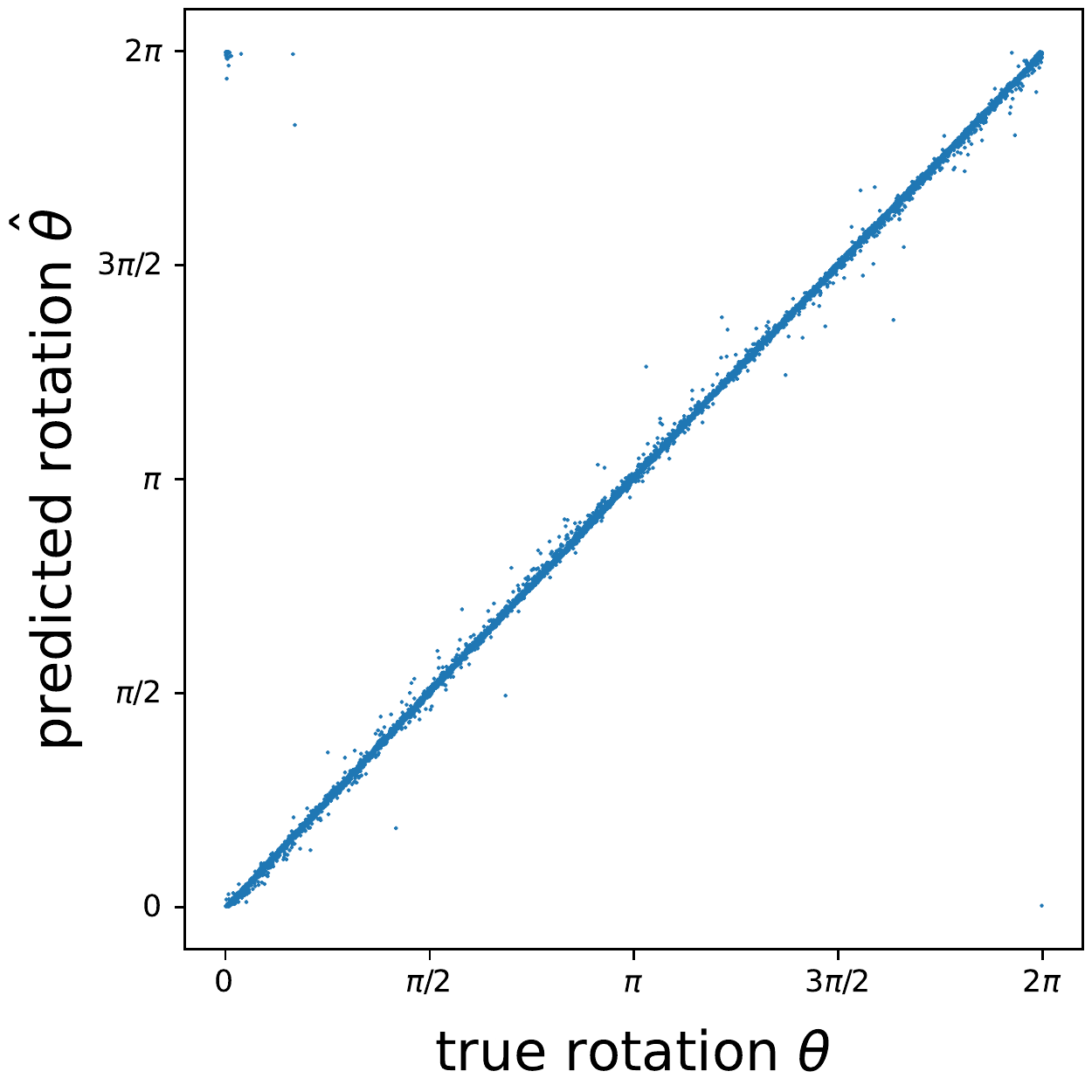}
    \subcaption*{MNIST(U)}
    \label{img1}
\end{minipage}
% \hfill
\begin{minipage}{.48\linewidth}
    \includegraphics[width=40mm]{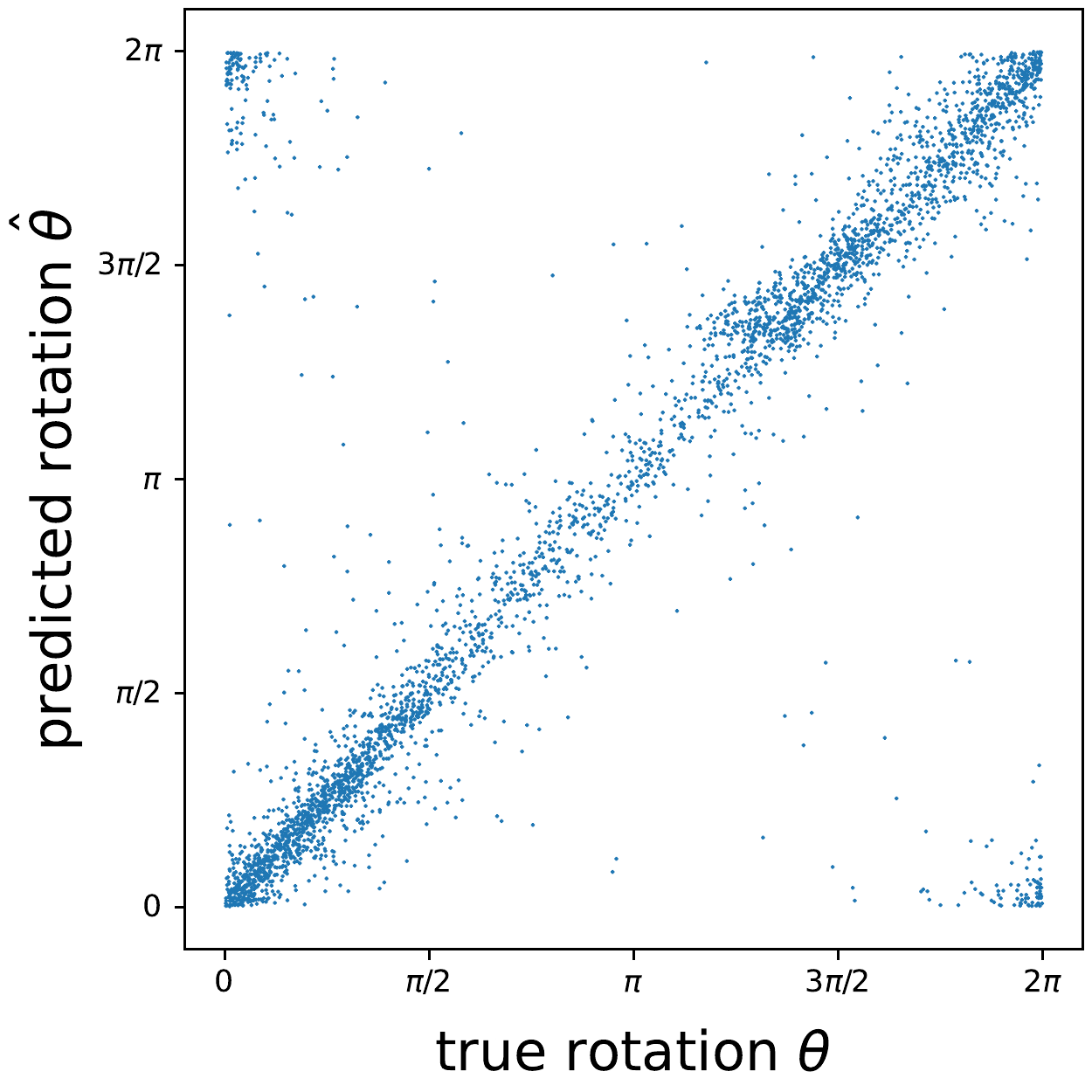}
    \subcaption*{WM811k}
    \label{img2}
\end{minipage}
\caption{Difference between predicted rotation values and true rotation values on MNIST(U) and WM811.}
\label{fig:predicted theta plotting}
\end{figure}

Table~\ref{table:correlation} and Figure~\ref{fig:predicted theta plotting} shows how well the predicted rotation $\hat{\theta}$ and predicted translation $\hat{\tau}$ matched the true rotation $\theta$ and true translation $\tau$.
Table~\ref{table:correlation} shows the Pearson correlation between the predicted rotation $\hat{\theta}$ and the true rotation $\theta$, predicted translation $\hat{\tau}$ and true translation $\tau$. We confirmed that our method has the highest correlation value. Also, in \cref{fig:predicted theta plotting} we plotted values of $\theta$ and $\hat{\theta}$. We can observe that most of predicted rotation degree are exactly same with true rotation. Interestingly, in the case of the WM811k, there were many cases where predicted degree and true degree are differed by $2\pi$, which is acceptable because the rotation degree is equivalent under mod $ 
2\pi$.

% Also, we provide a graph that actually plots the predicted rotation value and the true rotation value  (\jy{Figure 3 and 5 had the same reference naming, so I changed the ref for Figure 5 as z dim scaling}) Since there are cases where the difference between the predicted rotation value and the true rotation value is as much as $2\pi$, rotation can be interpreted as having a lower correlation than translation. However, since the rotation value is equivalent under $2\pi$, there is no problem with the difference between the predicted rotation value and the true rotation value by $2\pi$. This phenomenon can be equally applied to WM811k.

\begin{table}[]
\centering
\begin{tabular}[t]{lccc}
\toprule
&Translation & Rotation\\
\midrule
Spatial-VAE & 0.982, 0.983 & 0.005  \\
$\text{TARGET-VAE P}_{4}$ & 0.975, 0.976 & 0.80  \\
$\text{TARGET-VAE P}_{8}$ & 0.972, 0.971 & 0.859 \\
$\text{TARGET-VAE P}_{16}$ & 0.974, 0.971 & 0.93 \\
IRL-INR & \textbf{0.999, 0.999} & \textbf{0.9891} \\
\bottomrule
\end{tabular}
\caption{Correlation between true rotation and predicted rotation and true translation and predcited translation from MNIST(U).}
\label{table:correlation}
\end{table}%

\subsection{Clustering with semantic representations and SCAN}
\label{subsec:clustering}
As the semantic representations are disentangled from the orientation of the image, they should be more amenable to be used for clustering, provided that the semantic meaning of the underlying dataset is invariant under different orientations of the image. In this section, we use the semantic representations to perform clustering based on two approaches: directly with $z$ and using SCAN.

Table~\ref{tab:clustering on semantics representation space} shows the results of applying $k$-means and agglomerative clustering on the semantics representation $z$. For TARGET-VAE, we used the original authors' code and hyperparameters to best reproduce their results. We see that the semantic representation produced by our framework IRL-INR has better clustering accuracies and has significantly less variability.

% We did our best from the given code by the authors of TARGET-VAE. We repeated the experiment 5 times in MNIST(U) and WM811k. We measured the mean and standard deviation. Our method improves the clustering performance on the feature space than existing representation learning methods.
\begin{table}[!htb]
  \centering
  \begin{tabular}{lccc}
    \toprule
     &MNIST(U)& WM811k\\
    \midrule
    Spatial-VAE (K-means) & 31.87 $\pm$ 3.72& 27.4 $\pm$ 1.16\\
    Spatial-VAE (Agg) & 35.62 $\pm$ 2.08 & 28.73 $\pm$ 1.39\\
    Target-VAE (K-means) & 64.63 $\pm$ 4.4 & 39.6 $\pm$ 1.29\\
    Target-VAE (Agg) & 68.8 $\pm$ 4.39 & 40.11 $\pm$ 2.7\\
    IRL-INR (K-means) & 59.6 $\pm$ 1.12 & 55.06 $\pm$ 1.83\\
    IRL-INR (Agg) & \textbf{71.53} $\bf{\pm}$ \textbf{1.01} & \textbf{56.74} $\pm$ 
    \textbf{1.15}\\
    \bottomrule
  \end{tabular}
  \caption{Clustering on semantics representation $z$.
  The confidence interval is a single standard deviation measured over 5 runs.
  }
  \label{tab:clustering on semantics representation space}
\end{table}

% \subsection{Clustering with SCAN}
% \label{subsec:SCAN}

To further improve the clustering accuracy, we combine our framework with one of the state-of-the-art deep-learning-based method SCAN \cite{vangansbeke2020scan}. The original SCAN framework adopted SimCLR \cite{chen_2020_simclr} as its pretext task. % minimizing the InfoNCE loss \cite{DBLP:journals/corr/abs-1807-03748} as its pretext task.
We instead use the training of IRL-INR as a pretext task and then use the trained encoder network $\enc_\phi$ with only the $z$ output for the SCAN framework. (The $(\hat{\theta},\hat{\tau})$ are not used with SCAN.) Since SimCLR is based on InfoNCE loss \cite{DBLP:journals/corr/abs-1807-03748}, which uses cosine similarity, we also use cosine similarity distance in training IRL-INR.

Table~\ref{tab:example} shows that IRL-INL synergizes well with SCAN to produce state-of-the-art performance. Specifically, the clustering accuracy significantly outperforms vanilla SCAN (with InfoNCE loss) and combining Target-VAE with SCAN yields little or no improvement compared to directly  clustering the semantic representations of Target-VAE.

\begin{figure}[htb!]
\centering
\begin{minipage}{.48\linewidth}
    \includegraphics[width=40mm]{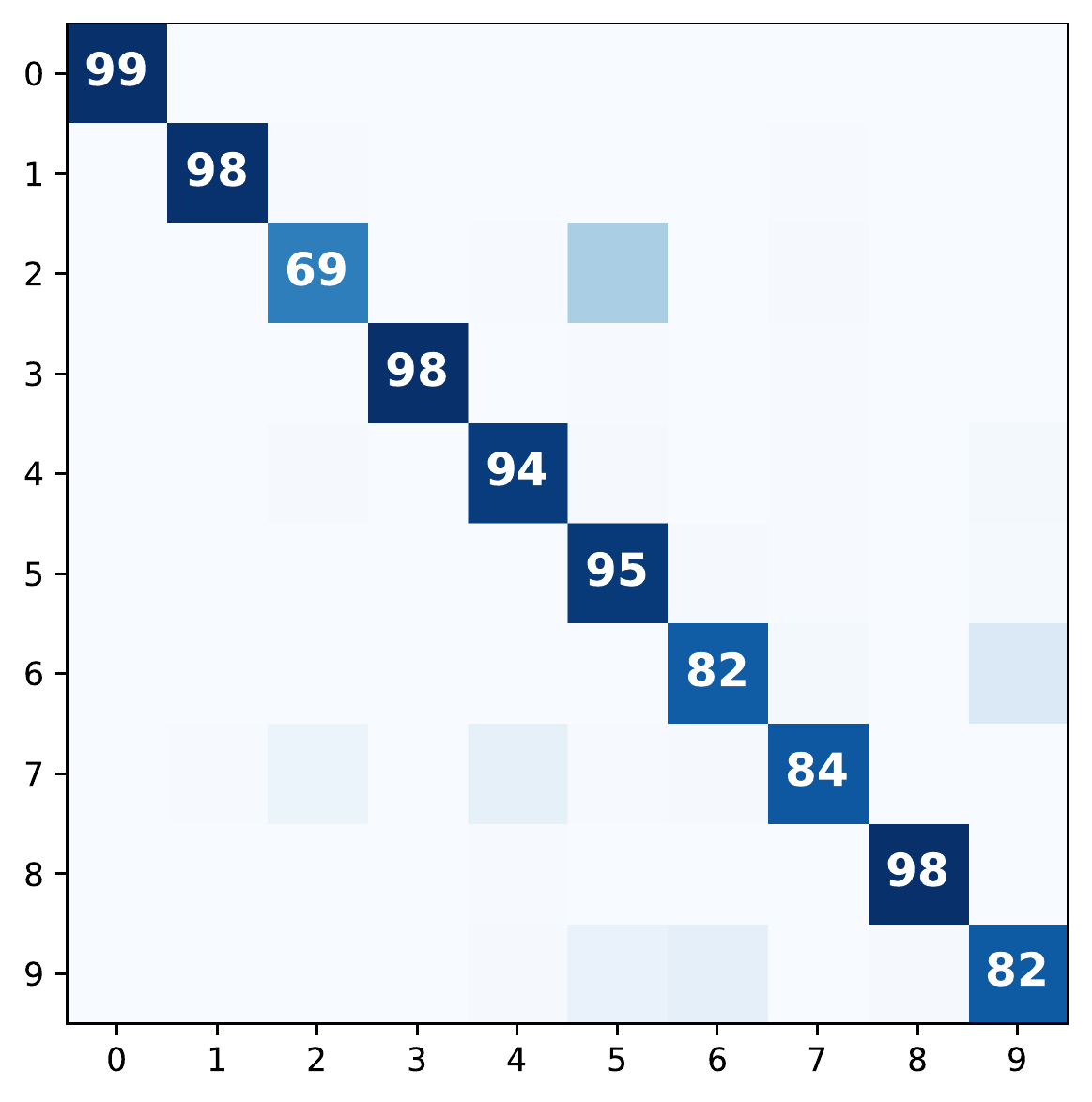}
    \subcaption*{MNIST(U)}
    \label{img1}
\end{minipage}
% \hfill
\begin{minipage}{.48\linewidth}
    \includegraphics[width=40mm]{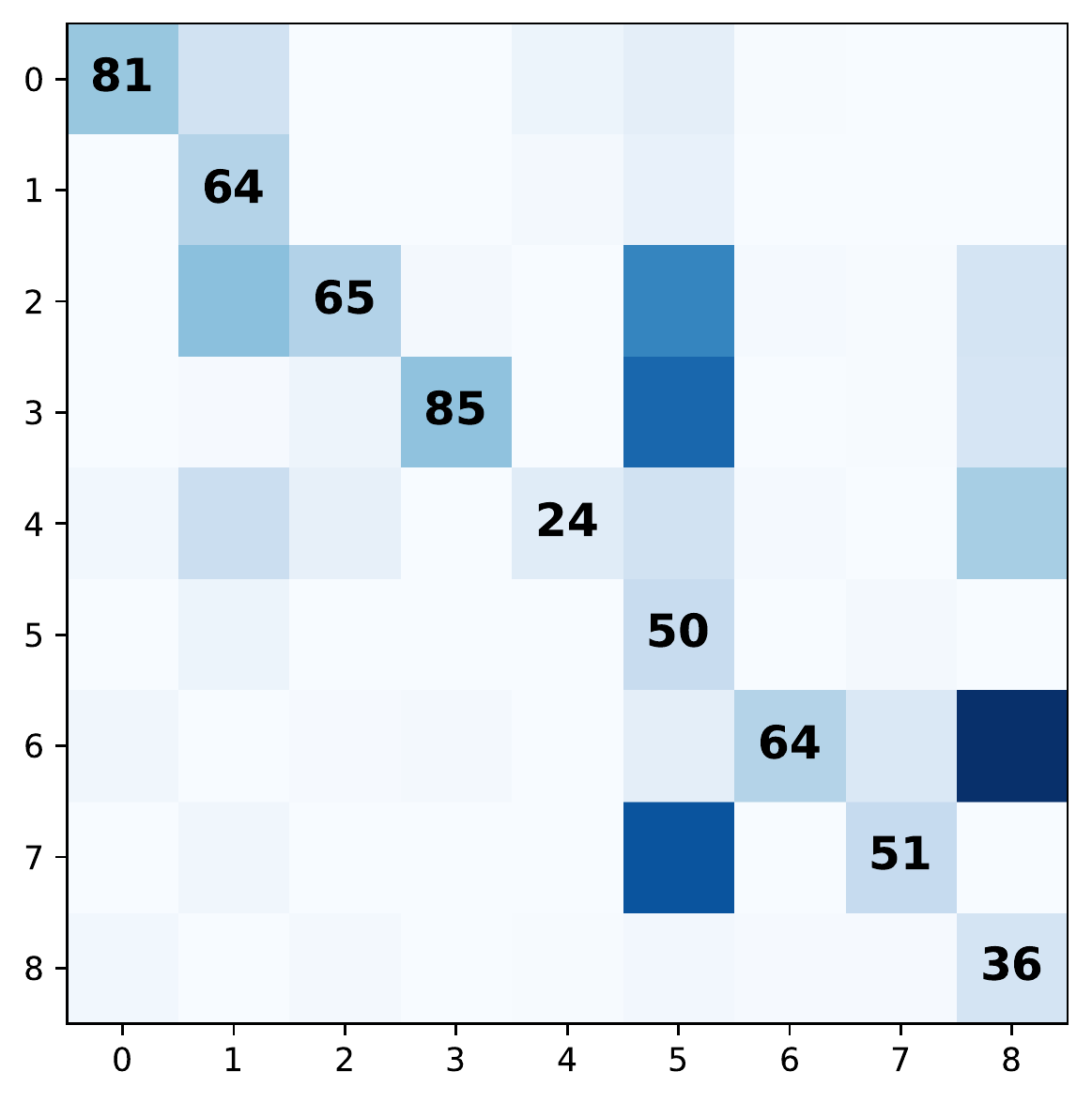}
    \subcaption*{WM811k}
    \label{img2}
\end{minipage}
\vspace{-0.1in}
\caption{\mbox{Confusion matrices of clustering of IRL-INR + SCAN.}
% According to the confusion matrix, accuracy is low in 6 and 9, which is thought to be because 6 and 9 are in a equal relationship with each other.
}
\label{fig:confusion matrix}
\end{figure}

The confusion matrix \cref{fig:confusion matrix} shows that there is significant misclassification between 6 and 9.
In \cref{sec:clustering on removed mnist}, we present our clustering results with the class 9 removed, and  IRL-INR + SCAN achieves a 98\% accuracy.
% This can be thought as IRL-INR is trained to extract rotationally invariant features.

%  This rotation and translation invariant backbone improves clustering performance through ignoring random rotation and translation which are nuisance information when do the clustering. XXX We need more justification for ``why does INR work well with SCAN? ''XXX There was no performance gain using TARGET-VAE as a pretext task

% Inspired by this, we use cosine distance as the consistency loss when combining our method with SCAN. To be trained on complex data such as wm811k using cosine distance as consistency loss, we only succeeded in our scheme. 

% InfoNCE minimizes the cosine distance between latent vectors with the same semantics, while forcing it to maximize the cosine distance between latent vectors with different semantics. 

\begin{table}[!htb]
  \centering
  \begin{tabular}{lccc}
    \toprule
     \!\!\!  &\!\!\!\scalebox{1.0}{MNIST(U)}& \!\!\!\scalebox{1.0}{WM811k}\\
    \midrule
    \!\!\!\scalebox{1.0}{%
TARGET-VAE + SCAN} &\!\!\! \scalebox{1.0}{63.09 $\pm$ 1.7 } & \scalebox{1.0}{43.39  $\pm$ 4.55} \\
    % \!\!\!\scalebox{0.88}{%
% SimCLR + SCAN ($d=32$})&\!\!\! \scalebox{0.88}{84.8 $\pm$ 0.74 } & \scalebox{0.88}{55.2  $\pm$ 2.13}\\
    \!\!\!\scalebox{1.0}{%
    SimCLR + SCAN }&\!\!\! \scalebox{1.0}{85.4 $\pm$ 1.46 } & \scalebox{1.0}{57.1  $\pm$ 2.81}\\
%     \!\!\!\scalebox{0.88}{%
% IRL-INR + SCAN ($d=32$}) &\!\!\! \scalebox{0.88}{86.18 $\pm$ 2.11 } & \scalebox{0.88}{59.78  $\pm$ 3.41}\\
     \!\!\!\scalebox{1.0}{%
IRL-INR + SCAN} &\!\!\! \scalebox{1.0}{\textbf{90.4} $\pm$ \textbf{1.74}} &\!\!\! \scalebox{1.0}{\textbf{64.6} $\pm$ \textbf{1.01}} \\
    \bottomrule
  \end{tabular}
  \caption{Using IRL-INR as pretext task for SCAN outperformed other combinations using TARGET-VAE and SimCLR. Here, $d$ is the dimension of the semantic representation $z$.}
  \label{tab:example}
\end{table}

%   \vspace{0.1in}

\subsection{Scaling latent dimension $d$}
\label{subsec:scale-z}

\cref{table:scaling} shows that clustering accuracy of IRL-INR scales (improves) as the latent dimension $d$, the size of the semantic representation $z$, becomes larger.
This phenomenon may seem counterintuitive as one might think that a smaller semantic representation is a more compressed and, therefore, better representation.

We also tried scaling the output dimension of the SimCLR + SCAN's backbone model, but we did not find any noticeable performance gain or trend. To clarify, SimCLR + SCAN and IRL-INR + SCAN used the same backbone model, ResNet18, but only IRL-INR + SCAN exhibited the scaling improvement. We also conducted a similar scaling experiment with TARGET-VAE, but we did not find any performance gain or trend with or without SCAN.

% We emphasize once again that our encoder and the backbone model of \sh{SimCLR + }SCAN are exactly same structure.

% We emphasize once again that our encoder and the backbone model of \sh{SimCLR + }SCAN are exactly same structure.

\begin{table}[!htb]
  \centering
  \begin{tabular}{lccc}
    \toprule
     \!\!\! IRL-INR + SCAN &\!\!\!\ MNIST(U)& \!\!\! WM811k\\
    \midrule
    $d = 32$ &\!\!\! 84.18 $\pm$ 2.11 & 53.78  $\pm$ 3.41 \\
    $d = 64$ &\!\!\! 86 $\pm$ 1.78 &\!\!\! 55.4 $\pm$ 1.35\\
    $d = 128$ &\!\!\! 85.8 $\pm$ 1.46 &\!\!\! 56.2 $\pm$ 1.16\\
    $d = 256$ &\!\!\! 87 $\pm$ 0.89 &\!\!\! 58.6 $\pm$ 1.62\\
    $d = 512$ &\!\!\! \textbf{90.4} $\pm$ \textbf{1.74} &\!\!\! \textbf{64.6} $\pm$ \textbf{1.01} \\
    \bottomrule
  \end{tabular}
\vspace{-0.1in}
  \caption{Increasing latent dimension $d$ of IRL-INR leads to better clustering performance.}
  \label{table:scaling}
\end{table}

\subsection{Ablation studies}
\label{ablation}
IRL-INR uses a hypernetwork-INR architecture, but one can alternatively consider: (1) using a simple MLP generator or (2) using a standard autoencoder while directly rotating the generated output image through pixel interpolation.
We find that both alternatives fail in the following sense. For the more complex images, such as the plankton microscope or silicon wafer maps, if we use the losses for requiring the semantic representation to be invariant, then the models fail the image reconstruction pretext task in the sense that the reconstruction comes out to be a blur with no discernable content. When we nevertheless proceeded to cluster the latents, the accuracy was poor. Using the hypernetwork was the only option that allowed us to do clustering successfully for the silicon wafer maps.

Also, the loss function of IRL-INR consists of three components: (1) reconstruction loss (2) consistency loss (3) symmetry breaking loss. We conducted ablation study on the different loss terms and found that all components are essential to reconstruction and clustering. For example, removing the consistency loss does not affect the reconstruction quality but does significantly reduce the clustering accuracy. Also, as we see in \cref{fig:breaking sym}, removing the symmetry-breaking loss significantly degrades the reconstruction quality, thereby worsening the clustering results.

\section{Conclusion}
% In this work, we propose Invariant Representation Learning with Implicit Neural Representation (IRL-INR), which uses an INR with a hypernetwork to obtain semantic representations disentangled from the orientation of the image. We show that IRL-INR can learn disentangled semantic representations complex images and achieve state-of-the-art clustering accuracy when combine with the deep-learning-based clustering method SCAN.
We proposed IRL-INR, which uses an INR with a hypernetwork to obtain semantic representations disentangled from the orientation of the image and used the semantic representations to achieve state-of-the-art clustering accuracy. Using explicit invariances in representation learning is a relatively underexplored approach. We find such representations to be especially effective in unsupervised clustering as there is a stronger reliance on inductive biases in the setup. Further exploring how to exploit various invariances exhibited in different datasets is an interesting future direction.

% \newpage

\section*{Acknowledgements}
This work was supported by Samsung Electronics Co., Ltd (IO221012-02844-01) and the Creative-Pioneering Researchers Program through Seoul National University. We thank Jongmin Lee and Donghwan Rho for providing careful reviews and valuable feedback. We thank Haechang Harry Lee for the discussion on the use of unsupervised clustering for semiconductor wafer defect inspection. Finally, we thank the anonymous reviewers for their thoughtful comments.

\bibliography{example_paper}
\bibliographystyle{icml2023}

%%%%%%%%%%%%%%%%%%%%%%%%%%%%%%%%%%%%%%%%%%%%%%%%%%%%%%%%%%%%%%%%%%%%%%%%%%%%%%%
%%%%%%%%%%%%%%%%%%%%%%%%%%%%%%%%%%%%%%%%%%%%%%%%%%%%%%%%%%%%%%%%%%%%%%%%%%%%%%%
% APPENDIX
%%%%%%%%%%%%%%%%%%%%%%%%%%%%%%%%%%%%%%%%%%%%%%%%%%%%%%%%%%%%%%%%%%%%%%%%%%%%%%%
%%%%%%%%%%%%%%%%%%%%%%%%%%%%%%%%%%%%%%%%%%%%%%%%%%%%%%%%%%%%%%%%%%%%%%%%%%%%%%%
\newpage
\appendix
\onecolumn
\section{Architectural details}
\label{sec:architecture_detail}
% \subsection*{Encoder}
\paragraph{Encoder.} For the encoder network $\enc_\phi$, we use the ResNet18 architecture \cite{he2016}, a 3-layered MLP head with dimensions [512, 512, $d$+2] where $d$ is dimension of semantic representation, and the ReLU activation. 

\paragraph{Hypernetwork.}
For the Hypernetwork $\hyp_\psi$, we use a 4-layered MLP with dimensions [$d$, 256, 256, 256, $k$] where $k$ is the number of parameters (weights and biases) of the INR network, and the LeakyReLU(0.1) activation.

\paragraph{INR Network.}
We parameterize a continuous image $\mathcal{I}$ by the INR-Network $\dec$. Basically, $\dec$ takes coordinate $(x,y)$ as an input and outputs pixel value of that coordinate. More specifically, $\dec$ consists of two parts. For the first, $\dec$ transforms input coordinate to fourier features following \citet{Rahimi_randomfeature, tancik2020}, where fourier features of $(x,y)$ is defined as
\begin{equation*}
    \text{FF}(\mathbf{x})=\left(\begin{array}{c}
\cos (2 \pi B \mathbf{x}) \\
\sin (2 \pi B \mathbf{x})
\end{array}\right)
\end{equation*}
with $B$ being an $f$ by $2$ random matrix whose entries are sampled from $\mathcal{N}(0, \sigma^2)$. The number of frequencies $f$ and the variance $\sigma^2$ are hyperparameters. In this paper, we use $f=256$ and $\sigma=2.0$. For the second part, fourier features are fed to the 4-layered MLP with dimensions [$2f$-256-256-256-$C$], which then outputs the pixel value, where $C$ is the number of color channls ($C=1$ for greyscale images, and $C=3$ for RGB images). 

\section{Experimental details}
We report the experimental details in Table \ref{table:hyperparameters}. We use the Adam optimizer with learning rate = 0.0001, batch size = 128, and weight decay = 0.0005, for all datasets. We train the model for 200, 500, 2000, 100 epochs for MNIST(U), WM811k, WHOI-Plankton, and \{5HDB, dSprites, Galaxy zoo\} respectively. We run all our experiments on a single NVIDIA RTX 3090 Ti GPU with 24 GB memory.

\begin{table}[!htb]
  \centering
  \begin{tabular}{ lccccc }
    \toprule
    Dataset & LR & Batch size & WD & Epochs \\
    \midrule
    MNIST(U) & 0.0001 & 128 & 0.0005 & 200 \\
    WM811k & 0.0001 & 128 & 0.0005 & 500 \\
    WHOI-Plankton & 0.0001 & 512 & 0.001 & 2000 \\
    5HDB & 0.0001 & 128 & 0.0005 & 100 \\
    dSprites & 0.0001 & 128 & 0.0005 & 100 \\
    Galaxy Zoo & 0.0001 & 128 & 0.0005 & 100 \\
    \bottomrule
  \end{tabular}
  \vspace{-5mm}
  \caption{Hyperparameters for the experiments}
  \label{table:hyperparameters}
\end{table}

\section{Data-augmentation for SCAN training}
The SCAN framework \cite{vangansbeke2020scan} consists of two stages: the first stage is \textit{pretext task stage} that learns a meaningful representation and the second stage is \textit{minimizing clustering loss} stage. For \textit{pretext task} stage, the original authors of SCAN experimented with various pretext tasks and observed that contrastive learning methods, such as MoCo \cite{he2019moco} and SimCLR \cite{chen_2020_simclr}, were most effective. In this paper, we use SimCLR as the pretext task for SCAN, which the authors used for the dataset with small resolution such as CIFAR10. We denote this combination as `SimCLR +  SCAN'.

SimCLR uses random crop (with flip and resize), color distortion and Gaussian blur data-augmentation strategies, and we denote this strategy by $\textbf{S}_1$. For \textit{minimizing clustering loss} stage, the authors of SCAN reported that adding strong augmentations including RandAugment \cite{cubuk_randaugment} and Cutout \cite{devries2017cutout}, which we denote by $\textbf{S}_2$, showed better performance. So, in the original SCAN framework, $\textbf{S}_1$ is applied in the first stage, and $\textbf{S}_1+\textbf{S}_2$ in the second stage.

However, if we naively follow the data-augmentation strategy used by the original SCAN, clustering performance of MNIST(U) and WM811k were suboptimal, as reported in Table \ref{scan-data-augmentation}. We suspect that this is due to the nature of contrastive learning methods. Recall that contrastive learning methods, such as SimCLR, only force the representation to be invariant under specific data augmentation strategy. Hence to extract invariant representation from dataset with strong rotation and translation variations, such as MNIST(U), WM811k, more powerful rotation and translation augmentations should be applied, especially in the pretext task stage. So we add \textit{random rotation}, \textbf{R}, where rotation angle is sampled from $\mathrm{Uniform}([0,2\pi))$ and \textit{random translation}, \textbf{T}, where translation is sampled from $\mathrm{Uniform}([-T,T])$. For the implementation details, we use functionals provided by \verb|Torchvision|: \verb|RandomRotation(180)| and \verb|RandomAffine(translate=(-T/P, T/P))| for \textbf{T} for \textbf{R} and \textbf{T} respectively. In our experiment, we set \verb|T| = 0.07 $\times$ \verb|P| where \verb|P| is spatial dimension of the image. 

\begin{table}[h!]
\begin{tabular}{cccc}
    \toprule
     \!\!\! Data Augmentation (Pretext) &\!\!\! Data Augmentation (Clustering)& \!\!\!Accuracy\\
    \midrule
\textbf{R}\!\!\! & \textbf{R} &\!\!\! 41.38 $\pm$ 2.07\\
  $\textbf{R}+\textbf{T}$\!\!\!& $\textbf{R}+\textbf{T}$ &\!\!\! 45.19 $\pm$ 3.62\\
 $\textbf{S}_1$\!\!\! & $\textbf{S}_1+\textbf{S}_2$ &\!\!\! 52.8 $\pm$ 3.86\\
   $\textbf{S}_1+\textbf{R}$\!\!\!&  $\textbf{S}_1+\textbf{S}_2+\textbf{R}$  &\!\!\! 83.66 $\pm$  1.71 \\
   $\textbf{S}_1+\textbf{R}+\textbf{T}$\!\!\! & $\textbf{S}_1+\textbf{S}_2+\textbf{R}+\textbf{T}$ &\!\!\! 85.4 $\pm$ 1.46 \\
    \bottomrule
  \end{tabular}
  \vspace{-5mm}
  \caption{Data augmetation strategies for SimCLR +  SCAN}
  \label{scan-data-augmentation}
  \end{table}

For IRL-INR, we apply $\textbf{R} + \textbf{T}$ for the pretext task stage. As in the SimCLR +  SCAN, IRL-INR + SCAN does benefit from stronger augmentation strategies as reported in Table \ref{IRL-INR-data-augmentation}. However, it can still outperform SimCLR +  SCAN with very simple data augmentation strategies such as \textbf{R}, or $\textbf{R}+\textbf{T}$.

\begin{table}[h!]
\begin{tabular}{cccc}
    \toprule
     \!\!\! Data Augmentation (Pretext) &\!\!\! Data Augmentation (Clusteirng)& \!\!\!Accuracy\\
    \midrule
$\textbf{R} + \textbf{T}$ & \textbf{R} &\!\!\! 86.42 $\pm$ 1.06\\
$\textbf{R} + \textbf{T}$ &  $\textbf{R}+\textbf{T}$&\!\!\! 87.11 $\pm$ 1.24\\
$\textbf{R} + \textbf{T}$ &  $\textbf{S}_1+\textbf{S}_2$ &\!\!\! 85.3 $\pm$ 1.88\\
$\textbf{R} + \textbf{T}$ & $\textbf{S}_1+\textbf{S}_2+\textbf{R}$ &\!\!\! 89.71 $\pm$ 2.93 \\
$\textbf{R} + \textbf{T}$ &  $\textbf{S}_1+\textbf{S}_2+\textbf{R}+\textbf{T}$ &\!\!\! \textbf{90.4} $\pm$ \textbf{1.74} \\
    \bottomrule

  \end{tabular}
\caption{Data Augmentation Strategies for IRL-INR + SCAN}
\vspace{-5mm}
\label{IRL-INR-data-augmentation}
\end{table}

% \begin{table}[!htb]
%   \centering
%   \begin{tabular}{lcccc}
%     \toprule
%      Pretext task / D.A &\!\!\! Clustering / D.A& \!\!\!Accuracy\\
%     \midrule
%     \!\!\!\textbf{S}&\!\!\! 52.8 $\pm$ 3.86 &\!\!\! 85.3 $\pm$ 1.88\\
%     \!\!\!\textbf{R}&\!\!\! 41.38 $\pm$ 2.07 &\!\!\! 86.42 $\pm$ 1.06\\
%     \!\!\!\textbf{R+T}&\!\!\! 45.19 $\pm$ 3.62 &\!\!\! 87.11 $\pm$ 1.24\\
%     \!\!\!\textbf{S+R}&\!\!\! 83.66 $\pm$  1.71 &\!\!\! 89.71 $\pm$ 2.93 \\
%     \!\!\!\textbf{S+R+T}&\!\!\! 85.4 $\pm$ 1.46 &\!\!\! \textbf{90.4} $\pm$ \textbf{1.74} \\
%     \bottomrule
%   \end{tabular}

%   \caption{Various data-augmentation strategies. \textbf{S} denotes ... \textbf{R} denotes ... \\
%   \textbf{R+T} denotes ... \textbf{S+R} denotes ... \textbf{S+R+T} denotes ...}
%   \label{data-augmentation}
% \end{table}
\section{MNIST(U)\textbackslash\{9\}}
\label{sec:clustering on removed mnist}
As shown in \cref{fig:confusion matrix}, clustering accuracy of MNIST(U) was low for \{6\} and \{9\}.
This seems natural, because once a network has learned the rotation invariant representations of images, it could identify \{6\} and \{9\} as having similar (if not same) semantic representation. 
% \begin{figure*}[!htb]
% \centering
%         \begin{subfigure}[b]{0.5\linewidth}
%                 \centering
%                 \includegraphics[width=.95\linewidth]{image/Appendix/IRL-INR + SCAN on R-MNIST/rotated_6.pdf}\\ [2mm]
%                 \includegraphics[width=.95\linewidth]{image/Appendix/IRL-INR + SCAN on R-MNIST/output_6.pdf}
%                 % \captionsetup{font=scriptsize}
%                 \caption{reconstruct `6' to `9'}
%         \end{subfigure}%
%         \begin{subfigure}[b]{0.5\linewidth}
%                 \centering
%                 \includegraphics[width=.95\linewidth]{image/Appendix/IRL-INR + SCAN on R-MNIST/rotated_9.pdf}\\ [2mm]
%                 \includegraphics[width=.95\linewidth]{image/Appendix/IRL-INR + SCAN on R-MNIST/output_9.pdf}
%                 % \captionsetup{font=scriptsize}
%                 \caption{reconstruct `9' to `6'}
%         \end{subfigure}
%         \caption{.}
%         \label{fig:misrecon}
% \end{figure*}

\begin{figure}[!htb]
\begin{floatrow}
\ffigbox{%
\includegraphics[width=.5\linewidth]{./figure/confusion_matrix_MNIST_U_.pdf}%
\includegraphics[width=.5\linewidth]{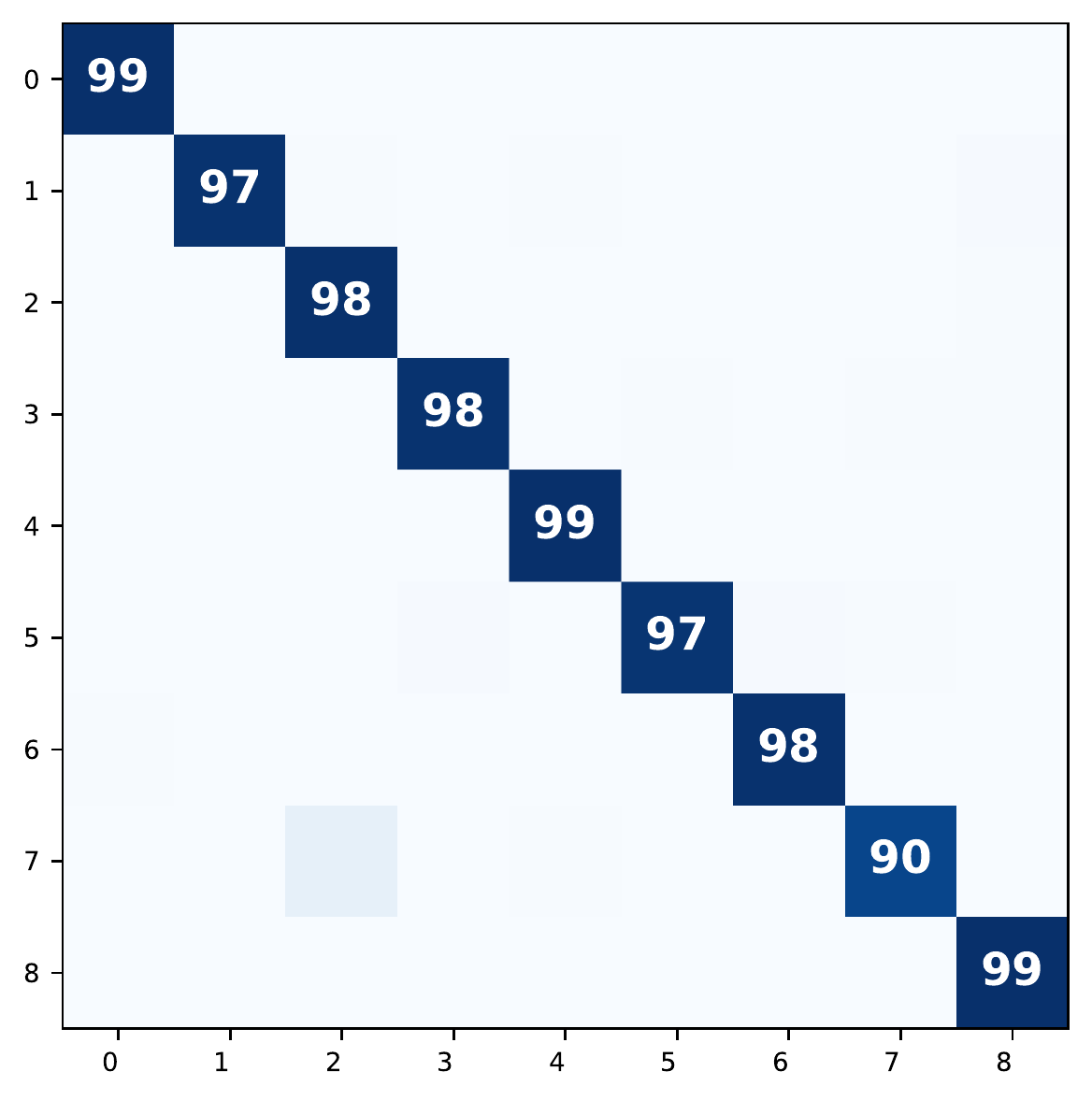}%
}{\caption{Confusion matrices for MNIST (U) dataset (left), and MNIST (U)\textbackslash\{9\} dataset (right)}}
\capbtabbox{%
  \begin{tabular}{lcc}
    \toprule
     \!\!\!&\!\!\!MNIST(U)\\
    \midrule
    \!\!\!SimCLR + SCAN &\!\!\! 85.4 $\pm$ 1.46 \\
     \!\!\!IRL-INR + SCAN&\!\!\! \textbf{90.4} $\pm$ \textbf{1.74} \\
    \bottomrule
    \toprule
     \!\!\!&\!\!\!MNIST(U)\textbackslash\{9\}\\
    \midrule
    \!\!\!SimCLR + SCAN &\!\!\! 93.8 $\pm$ 0.74 \\
     \!\!\!IRL-INR + SCAN  &\!\!\! \textbf{97.6} $\pm$ \textbf{0.48} \\
    \bottomrule
  \end{tabular}
  \vspace{5mm}
}{%
  \caption{Clustering accuracy for MNIST(U) dataset and MNIST(U)\textbackslash\{9\} dataset}%
  \label{mnist u-9 accuracy}
}
\end{floatrow}
\end{figure}

To verify this conjecture, we experiment IRL-INR + SCAN and SimCLR +  SCAN with a new dataset 'MNIST(U)\textbackslash\{9\}' created by removing \{9\} from original MNIST(U) dataset. As reported in Table \ref{mnist u-9 accuracy}, removing \{9\} significantly increased the clustering accuracy for both methods. Interestingly, by removing \{9\} we observed that the accuracy for \{2\}, which was lower than the average accuracy, significantly improved as well.

%%%%%%%%%%%%%%%%%%%%%%%%%%%%%%%%%%%%%%%%%%%%%%%%%%%%%%%%%%%%%%%%%%%%%%%%%%%%%%%
%%%%%%%%%%%%%%%%%%%%%%%%%%%%%%%%%%%%%%%%%%%%%%%%%%%%%%%%%%%%%%%%%%%%%%%%%%%%%%%
\section{Reconstructing $J$}
\label{Appendix: Reconstruct J}
In this section, we show image samples demonstrating that the IRL-INR does faithfully reconstruct the input image $J$.

\subsection{Image generation process}
The encoder $\enc_\phi$ outputs the rotation representation $\hat{\theta}$, translation representation $\hat{\tau}$, and semantic representation $z$. Hypernetwork $\hyp_\psi$ takes $z$ as an input and then outputs $\eta$, where $\eta$ is the set of weights and biases of INR network. The rotation representation $\hat{\theta}\in[0,2\pi)$ and translation representation $\hat{\tau}\in\mathbb{R}^2$ are trained to be estimates of the rotation and translation with respect to a certain canonical orientation. Hence, $\dec(\tilde{x}_p,\tilde{y}_p;\eta) \approx J_p$, where $(\tilde{x}_p,\tilde{y}_p)=S_{\hat{\theta},\hat{\tau}}(x_p,y_p)$. By accurately predicting the rotation degree and translation values, IRL-INR reconstructs identical images to the input images (\cref{Appendix:Reconstruct J image}).

\begin{figure}[h]
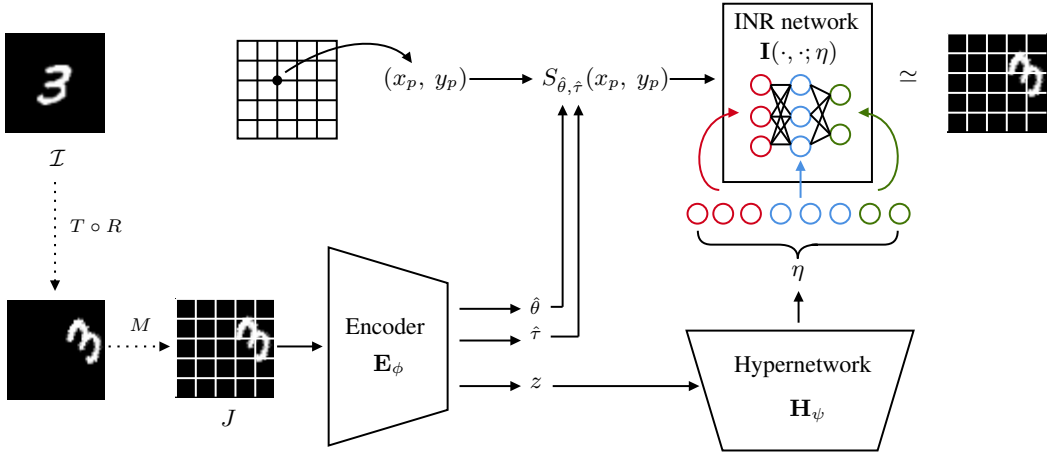

\centering

\tikzset{every picture/.style={line width=0.75pt}} %set default line width to 0.75pt        

\begin{tikzpicture}[x=0.75pt,y=0.75pt,yscale=-1,xscale=1]
%uncomment if require: \path (0,638); %set diagram left start at 0, and has height of 638

%Image [id:dp36094664519349107] 
\draw (110.5,174) node  {\includegraphics[width=36.75pt,height=37.5pt]{image/J.png}};
%Straight Lines [id:da48351678587254077] 
\draw  [dash pattern={on 0.84pt off 2.51pt}]  (52,173) -- (80,173) ;
\draw [shift={(83,173)}, rotate = 180] [fill={rgb, 255:red, 0; green, 0; blue, 0 }  ][line width=0.08]  [draw opacity=0] (5.36,-2.57) -- (0,0) -- (5.36,2.57) -- cycle    ;
%Straight Lines [id:da5381190519578207] 
\draw    (137,173) -- (157,173) ;
\draw [shift={(160,173)}, rotate = 180] [fill={rgb, 255:red, 0; green, 0; blue, 0 }  ][line width=0.08]  [draw opacity=0] (5.36,-2.57) -- (0,0) -- (5.36,2.57) -- cycle    ;
%Shape: Trapezoid [id:dp6933398854347409] 
\draw   (163,123) -- (223,141) -- (223,205) -- (163,223) -- cycle ;
%Straight Lines [id:da36174808681524095] 
\draw    (229,154) -- (257,154) ;
\draw [shift={(260,154)}, rotate = 180] [fill={rgb, 255:red, 0; green, 0; blue, 0 }  ][line width=0.08]  [draw opacity=0] (5.36,-2.57) -- (0,0) -- (5.36,2.57) -- cycle    ;
%Straight Lines [id:da13710175173779848] 
\draw    (229,193) -- (257,193) ;
\draw [shift={(260,193)}, rotate = 180] [fill={rgb, 255:red, 0; green, 0; blue, 0 }  ][line width=0.08]  [draw opacity=0] (5.36,-2.57) -- (0,0) -- (5.36,2.57) -- cycle    ;
%Straight Lines [id:da6763990160591609] 
\draw    (229,170) -- (251,170) -- (257,170) ;
\draw [shift={(260,170)}, rotate = 180] [fill={rgb, 255:red, 0; green, 0; blue, 0 }  ][line width=0.08]  [draw opacity=0] (5.36,-2.57) -- (0,0) -- (5.36,2.57) -- cycle    ;
%Straight Lines [id:da30538313471709533] 
\draw    (281,153) -- (281,54) ;
\draw [shift={(281,51)}, rotate = 90] [fill={rgb, 255:red, 0; green, 0; blue, 0 }  ][line width=0.08]  [draw opacity=0] (5.36,-2.57) -- (0,0) -- (5.36,2.57) -- cycle    ;
%Straight Lines [id:da7861366817587461] 
\draw    (275,153) -- (281,153) ;
%Straight Lines [id:da7657912611771184] 
\draw    (289,168) -- (289,54) ;
\draw [shift={(289,51)}, rotate = 90] [fill={rgb, 255:red, 0; green, 0; blue, 0 }  ][line width=0.08]  [draw opacity=0] (5.36,-2.57) -- (0,0) -- (5.36,2.57) -- cycle    ;
%Straight Lines [id:da6082345678471073] 
\draw    (275,168) -- (289,168) ;
%Straight Lines [id:da9508852027273481] 
\draw    (234,38) -- (263,38) ;
\draw [shift={(266,38)}, rotate = 180] [fill={rgb, 255:red, 0; green, 0; blue, 0 }  ][line width=0.08]  [draw opacity=0] (5.36,-2.57) -- (0,0) -- (5.36,2.57) -- cycle    ;
%Shape: Trapezoid [id:dp6125354363883906] 
\draw   (458,165) -- (439.85,225.5) -- (361.65,225.5) -- (343.5,165) -- cycle ;
%Straight Lines [id:da5038184043481221] 
\draw    (276,193) -- (347,193) ;
\draw [shift={(350,193)}, rotate = 180] [fill={rgb, 255:red, 0; green, 0; blue, 0 }  ][line width=0.08]  [draw opacity=0] (5.36,-2.57) -- (0,0) -- (5.36,2.57) -- cycle    ;
%Shape: Rectangle [id:dp6484774994965055] 
\draw   (362,0) -- (437,0) -- (437,90) -- (362,90) -- cycle ;
%Straight Lines [id:da7676104852060583] 
\draw    (335,38) -- (355,38) ;
\draw [shift={(358,38)}, rotate = 180] [fill={rgb, 255:red, 0; green, 0; blue, 0 }  ][line width=0.08]  [draw opacity=0] (5.36,-2.57) -- (0,0) -- (5.36,2.57) -- cycle    ;
%Image [id:dp6649595476896026] 
\draw (500,40) node  {\includegraphics[width=37.5pt,height=37.5pt]{image/J.png}};
%Shape: Brace [id:dp6098417311618172] 
\draw   (349.2,114.6) .. controls (349.2,119.27) and (351.53,121.6) .. (356.2,121.6) -- (389.6,121.6) .. controls (396.27,121.6) and (399.6,123.93) .. (399.6,128.6) .. controls (399.6,123.93) and (402.93,121.6) .. (409.6,121.6)(406.6,121.6) -- (443,121.6) .. controls (447.67,121.6) and (450,119.27) .. (450,114.6) ;
%Straight Lines [id:da34964773432151763] 
\draw    (400,161) -- (400,148) ;
\draw [shift={(400,145)}, rotate = 90] [fill={rgb, 255:red, 0; green, 0; blue, 0 }  ][line width=0.08]  [draw opacity=0] (5.36,-2.57) -- (0,0) -- (5.36,2.57) -- cycle    ;
%Shape: Circle [id:dp765703798326334] 
\draw  [color={rgb, 255:red, 208; green, 2; blue, 27 }  ,draw opacity=1 ] (344,106) .. controls (344,103.24) and (346.24,101) .. (349,101) .. controls (351.76,101) and (354,103.24) .. (354,106) .. controls (354,108.76) and (351.76,111) .. (349,111) .. controls (346.24,111) and (344,108.76) .. (344,106) -- cycle ;
%Shape: Circle [id:dp9683692324156519] 
\draw  [color={rgb, 255:red, 208; green, 2; blue, 27 }  ,draw opacity=1 ] (357,106) .. controls (357,103.24) and (359.24,101) .. (362,101) .. controls (364.76,101) and (367,103.24) .. (367,106) .. controls (367,108.76) and (364.76,111) .. (362,111) .. controls (359.24,111) and (357,108.76) .. (357,106) -- cycle ;
%Shape: Circle [id:dp3423680464348078] 
\draw  [color={rgb, 255:red, 208; green, 2; blue, 27 }  ,draw opacity=1 ] (371,106) .. controls (371,103.24) and (373.24,101) .. (376,101) .. controls (378.76,101) and (381,103.24) .. (381,106) .. controls (381,108.76) and (378.76,111) .. (376,111) .. controls (373.24,111) and (371,108.76) .. (371,106) -- cycle ;
%Shape: Circle [id:dp8639408354878916] 
\draw  [color={rgb, 255:red, 74; green, 144; blue, 226 }  ,draw opacity=1 ] (386,106) .. controls (386,103.24) and (388.24,101) .. (391,101) .. controls (393.76,101) and (396,103.24) .. (396,106) .. controls (396,108.76) and (393.76,111) .. (391,111) .. controls (388.24,111) and (386,108.76) .. (386,106) -- cycle ;
%Shape: Circle [id:dp06273770720341043] 
\draw  [color={rgb, 255:red, 74; green, 144; blue, 226 }  ,draw opacity=1 ] (401,106) .. controls (401,103.24) and (403.24,101) .. (406,101) .. controls (408.76,101) and (411,103.24) .. (411,106) .. controls (411,108.76) and (408.76,111) .. (406,111) .. controls (403.24,111) and (401,108.76) .. (401,106) -- cycle ;
%Shape: Circle [id:dp28945740594827984] 
\draw  [color={rgb, 255:red, 208; green, 2; blue, 27 }  ,draw opacity=1 ] (376,41) .. controls (376,38.24) and (378.24,36) .. (381,36) .. controls (383.76,36) and (386,38.24) .. (386,41) .. controls (386,43.76) and (383.76,46) .. (381,46) .. controls (378.24,46) and (376,43.76) .. (376,41) -- cycle ;
%Shape: Circle [id:dp8954968028605682] 
\draw  [color={rgb, 255:red, 208; green, 2; blue, 27 }  ,draw opacity=1 ] (376,57) .. controls (376,54.24) and (378.24,52) .. (381,52) .. controls (383.76,52) and (386,54.24) .. (386,57) .. controls (386,59.76) and (383.76,62) .. (381,62) .. controls (378.24,62) and (376,59.76) .. (376,57) -- cycle ;
%Shape: Circle [id:dp5484771141058965] 
\draw  [color={rgb, 255:red, 208; green, 2; blue, 27 }  ,draw opacity=1 ] (376,72) .. controls (376,69.24) and (378.24,67) .. (381,67) .. controls (383.76,67) and (386,69.24) .. (386,72) .. controls (386,74.76) and (383.76,77) .. (381,77) .. controls (378.24,77) and (376,74.76) .. (376,72) -- cycle ;
%Shape: Circle [id:dp8962157523481737] 
\draw  [color={rgb, 255:red, 74; green, 144; blue, 226 }  ,draw opacity=1 ] (416,106) .. controls (416,103.24) and (418.24,101) .. (421,101) .. controls (423.76,101) and (426,103.24) .. (426,106) .. controls (426,108.76) and (423.76,111) .. (421,111) .. controls (418.24,111) and (416,108.76) .. (416,106) -- cycle ;
%Shape: Circle [id:dp22388666326728834] 
\draw  [color={rgb, 255:red, 65; green, 117; blue, 5 }  ,draw opacity=1 ] (431,106) .. controls (431,103.24) and (433.24,101) .. (436,101) .. controls (438.76,101) and (441,103.24) .. (441,106) .. controls (441,108.76) and (438.76,111) .. (436,111) .. controls (433.24,111) and (431,108.76) .. (431,106) -- cycle ;
%Shape: Circle [id:dp10565607404415023] 
\draw  [color={rgb, 255:red, 65; green, 117; blue, 5 }  ,draw opacity=1 ] (446,106) .. controls (446,103.24) and (448.24,101) .. (451,101) .. controls (453.76,101) and (456,103.24) .. (456,106) .. controls (456,108.76) and (453.76,111) .. (451,111) .. controls (448.24,111) and (446,108.76) .. (446,106) -- cycle ;
%Shape: Circle [id:dp3380843088771198] 
\draw  [color={rgb, 255:red, 74; green, 144; blue, 226 }  ,draw opacity=1 ] (396,41) .. controls (396,38.24) and (398.24,36) .. (401,36) .. controls (403.76,36) and (406,38.24) .. (406,41) .. controls (406,43.76) and (403.76,46) .. (401,46) .. controls (398.24,46) and (396,43.76) .. (396,41) -- cycle ;
%Shape: Circle [id:dp22287424214790597] 
\draw  [color={rgb, 255:red, 74; green, 144; blue, 226 }  ,draw opacity=1 ] (396,57) .. controls (396,54.24) and (398.24,52) .. (401,52) .. controls (403.76,52) and (406,54.24) .. (406,57) .. controls (406,59.76) and (403.76,62) .. (401,62) .. controls (398.24,62) and (396,59.76) .. (396,57) -- cycle ;
%Shape: Circle [id:dp21151024201483437] 
\draw  [color={rgb, 255:red, 74; green, 144; blue, 226 }  ,draw opacity=1 ] (396,72) .. controls (396,69.24) and (398.24,67) .. (401,67) .. controls (403.76,67) and (406,69.24) .. (406,72) .. controls (406,74.76) and (403.76,77) .. (401,77) .. controls (398.24,77) and (396,74.76) .. (396,72) -- cycle ;
%Shape: Circle [id:dp28333452248089797] 
\draw  [color={rgb, 255:red, 65; green, 117; blue, 5 }  ,draw opacity=1 ] (416,46) .. controls (416,43.24) and (418.24,41) .. (421,41) .. controls (423.76,41) and (426,43.24) .. (426,46) .. controls (426,48.76) and (423.76,51) .. (421,51) .. controls (418.24,51) and (416,48.76) .. (416,46) -- cycle ;
%Shape: Circle [id:dp2110134099533797] 
\draw  [color={rgb, 255:red, 65; green, 117; blue, 5 }  ,draw opacity=1 ] (416,66) .. controls (416,63.24) and (418.24,61) .. (421,61) .. controls (423.76,61) and (426,63.24) .. (426,66) .. controls (426,68.76) and (423.76,71) .. (421,71) .. controls (418.24,71) and (416,68.76) .. (416,66) -- cycle ;
%Straight Lines [id:da8628859873131804] 
\draw    (386,41) -- (396,71) ;
%Straight Lines [id:da07738481760942673] 
\draw    (386,71) -- (396,41) ;
%Straight Lines [id:da15674543076471048] 
\draw    (386,57) -- (396,41) ;
%Straight Lines [id:da557691718873829] 
\draw    (386,71) -- (396,57) ;
%Straight Lines [id:da15689034470719831] 
\draw    (386,71) -- (396,71) ;
%Straight Lines [id:da9152793495623923] 
\draw    (386,57) -- (396,57) ;
%Straight Lines [id:da8895419102507541] 
\draw    (386,41) -- (396,41) ;
%Straight Lines [id:da5603304253404606] 
\draw    (396,71) -- (386,57) ;
%Straight Lines [id:da8201576551785246] 
\draw    (396,56) -- (386,42) ;
%Straight Lines [id:da8200627129169666] 
\draw    (416,66) -- (406,41) ;
%Straight Lines [id:da4700737595190887] 
\draw    (416,66) -- (406,56) ;
%Straight Lines [id:da15402528076133504] 
\draw    (416,66) -- (406,71) ;
%Straight Lines [id:da04016989600829057] 
\draw    (416,46) -- (406,41) ;
%Straight Lines [id:da41403161958488877] 
\draw    (406,56) -- (416,46) ;
%Straight Lines [id:da12848380695749184] 
\draw    (416,46) -- (406,71) ;
%Curve Lines [id:da39366718975494897] 
\draw [color={rgb, 255:red, 208; green, 2; blue, 27 }  ,draw opacity=1 ]   (360,94) .. controls (345.6,93.52) and (345.48,58.02) .. (367.17,55.18) ;
\draw [shift={(370,55)}, rotate = 180] [fill={rgb, 255:red, 208; green, 2; blue, 27 }  ,fill opacity=1 ][line width=0.08]  [draw opacity=0] (5.36,-2.57) -- (0,0) -- (5.36,2.57) -- cycle    ;
%Straight Lines [id:da9962199227105886] 
\draw [color={rgb, 255:red, 74; green, 144; blue, 226 }  ,draw opacity=1 ]   (401,98) -- (401,82) ;
\draw [shift={(401,79)}, rotate = 90] [fill={rgb, 255:red, 74; green, 144; blue, 226 }  ,fill opacity=1 ][line width=0.08]  [draw opacity=0] (5.36,-2.57) -- (0,0) -- (5.36,2.57) -- cycle    ;
%Curve Lines [id:da014988115275912373] 
%Curve Lines [id:da014988115275912373] 
\draw [color={rgb, 255:red, 65; green, 117; blue, 5 }  ,draw opacity=1 ]   (440,94) .. controls (453.92,93.04) and (454.94,57.52) .. (432.88,55.12) ;
\draw [shift={(430,55)}, rotate = 358.85] [fill={rgb, 255:red, 65; green, 117; blue, 5 }  ,fill opacity=1 ][line width=0.08]  [draw opacity=0] (5.36,-2.57) -- (0,0) -- (5.36,2.57) -- cycle    ;
%Image [id:dp06715216262310708] 
\draw (25,40) node  {\includegraphics[width=37.5pt,height=37.5pt]{image/I.png}};
%Straight Lines [id:da20752232802203552] 
\draw  [dash pattern={on 0.84pt off 2.51pt}]  (25,90) -- (25,142) ;
\draw [shift={(25,145)}, rotate = 270] [fill={rgb, 255:red, 0; green, 0; blue, 0 }  ][line width=0.08]  [draw opacity=0] (5.36,-2.57) -- (0,0) -- (5.36,2.57) -- cycle    ;
%Shape: Grid [id:dp5926366725817469] 
\draw  [draw opacity=0] (86,149) -- (136,149) -- (136,198.33) -- (86,198.33) -- cycle ; \draw  [color={rgb, 255:red, 255; green, 255; blue, 255 }  ,draw opacity=1 ] (86,149) -- (86,198.33)(96,149) -- (96,198.33)(106,149) -- (106,198.33)(116,149) -- (116,198.33)(126,149) -- (126,198.33) ; \draw  [color={rgb, 255:red, 255; green, 255; blue, 255 }  ,draw opacity=1 ] (86,149) -- (136,149)(86,159) -- (136,159)(86,169) -- (136,169)(86,179) -- (136,179)(86,189) -- (136,189) ; \draw  [color={rgb, 255:red, 255; green, 255; blue, 255 }  ,draw opacity=1 ]  ;
%Shape: Grid [id:dp16171381974120214] 
\draw  [draw opacity=0] (475,15) -- (525,15) -- (525,64.33) -- (475,64.33) -- cycle ; \draw  [color={rgb, 255:red, 255; green, 255; blue, 255 }  ,draw opacity=1 ] (475,15) -- (475,64.33)(485,15) -- (485,64.33)(495,15) -- (495,64.33)(505,15) -- (505,64.33)(515,15) -- (515,64.33) ; \draw  [color={rgb, 255:red, 255; green, 255; blue, 255 }  ,draw opacity=1 ] (475,15) -- (525,15)(475,25) -- (525,25)(475,35) -- (525,35)(475,45) -- (525,45)(475,55) -- (525,55) ; \draw  [color={rgb, 255:red, 255; green, 255; blue, 255 }  ,draw opacity=1 ]  ;
%Image [id:dp9213887313206829] 
\draw (25.5,175) node  {\includegraphics[width=36.75pt,height=37.5pt]{image/J.png}};
%Shape: Ellipse [id:dp9386365533780958] 
\draw  [fill={rgb, 255:red, 0; green, 0; blue, 0 }  ,fill opacity=1 ] (135.1,38.73) .. controls (135.1,37.59) and (135.99,36.67) .. (137.08,36.67) .. controls (138.17,36.67) and (139.05,37.59) .. (139.05,38.73) .. controls (139.05,39.86) and (138.17,40.78) .. (137.08,40.78) .. controls (135.99,40.78) and (135.1,39.86) .. (135.1,38.73) -- cycle ;
%Curve Lines [id:da8106363615038371] 
\draw    (141.05,34.67) .. controls (167.66,16.6) and (192.61,16.85) .. (203.16,27.78) ;
\draw [shift={(205,30)}, rotate = 234.75] [fill={rgb, 255:red, 0; green, 0; blue, 0 }  ][line width=0.08]  [draw opacity=0] (5.36,-2.57) -- (0,0) -- (5.36,2.57) -- cycle    ;
%Shape: Grid [id:dp5800930116279384] 
\draw  [draw opacity=1] (117.08,18.73) -- (167.08,18.73) -- (167.08,68.06) -- (117.08,68.06) -- cycle ; \draw  [color={rgb, 255:red, 0; green, 0; blue, 0 }  ,draw opacity=1 ] (127.08,18.73) -- (127.08,68.06)(137.08,18.73) -- (137.08,68.06)(147.08,18.73) -- (147.08,68.06)(157.08,18.73) -- (157.08,68.06) ; \draw  [color={rgb, 255:red, 0; green, 0; blue, 0 }  ,draw opacity=1 ] (117.08,28.73) -- (167.08,28.73)(117.08,38.73) -- (167.08,38.73)(117.08,48.73) -- (167.08,48.73)(117.08,58.73) -- (167.08,58.73) ; \draw  [color={rgb, 255:red, 0; green, 0; blue, 0 }  ,draw opacity=1 ]  ;

% Text Node
\draw (21,72.4) node [anchor=north west][inner sep=0.75pt]  [font=\footnotesize]  {$\mathcal{I}$};
% Text Node
\draw (107,203.4) node [anchor=north west][inner sep=0.75pt]  [font=\footnotesize]  {$J$};
% Text Node
\draw (170,157) node [anchor=north west][inner sep=0.75pt]  [font=\footnotesize] [align=left] {Encoder};
% Text Node
\draw (184,175.4) node [anchor=north west][inner sep=0.75pt]  [font=\footnotesize]  {$\mathbf{E}_{\phi }$};
% Text Node
\draw (263,146.4) node [anchor=north west][inner sep=0.75pt]  [font=\scriptsize]  {$\hat{\theta }$};
% Text Node
\draw (263,161.4) node [anchor=north west][inner sep=0.75pt]  [font=\scriptsize]  {$\hat{\tau }$};
% Text Node
\draw (263,187.4) node [anchor=north west][inner sep=0.75pt]  [font=\footnotesize]  {$z$};
% Text Node
\draw (269,30.4) node [anchor=north west][inner sep=0.75pt]  [font=\footnotesize]  {$S_{\hat{\theta } ,\hat{\tau }}( x_{p} ,\ y_{p})$};
% Text Node
\draw (189,30.4) node [anchor=north west][inner sep=0.75pt]  [font=\footnotesize]  {$( x_{p} ,\ y_{p})$};
% Text Node
\draw (366,176) node [anchor=north west][inner sep=0.75pt]  [font=\footnotesize] [align=left] {Hypernetwork};
% Text Node
\draw (394,198.4) node [anchor=north west][inner sep=0.75pt]  [font=\footnotesize]  {$\mathbf{H}_{\psi }$};
% Text Node
\draw (366,3) node [anchor=north west][inner sep=0.75pt]  [font=\footnotesize] [align=left] {INR network};
% Text Node
\draw (379,17.4) node [anchor=north west][inner sep=0.75pt]  [font=\footnotesize]  {$\mathbf{I}( \cdot ,\cdot ;\eta )$};
% Text Node
\draw (449,32.4) node [anchor=north west][inner sep=0.75pt]  [font=\footnotesize]  {$\simeq $};
% Text Node
\draw (395,131) node [anchor=north west][inner sep=0.75pt]  [font=\footnotesize]  {$\eta $};
% Text Node
\draw (61,157.4) node [anchor=north west][inner sep=0.75pt]  [font=\scriptsize]  {$M$};
% Text Node
\draw (31,107.4) node [anchor=north west][inner sep=0.75pt]  [font=\scriptsize]  {$T \circ R$};
\draw (490,70.4) ;
\end{tikzpicture}
\caption{Using $\hat{\theta}, \hat{\tau}$ and $z$ for reconstruction $J$. The input coordinates are rotated and translated by $\hat{\theta}$ and $\hat{\tau}$ for generating $J$.}
\label{fig:main-architectureXXX}

\end{figure}

\newpage
\subsection{Reconstruction results for $J$}
\begin{figure}[h!]
\centering
        \begin{subfigure}[b]{0.43\linewidth}
                \centering
                \includegraphics[width=.9\linewidth]{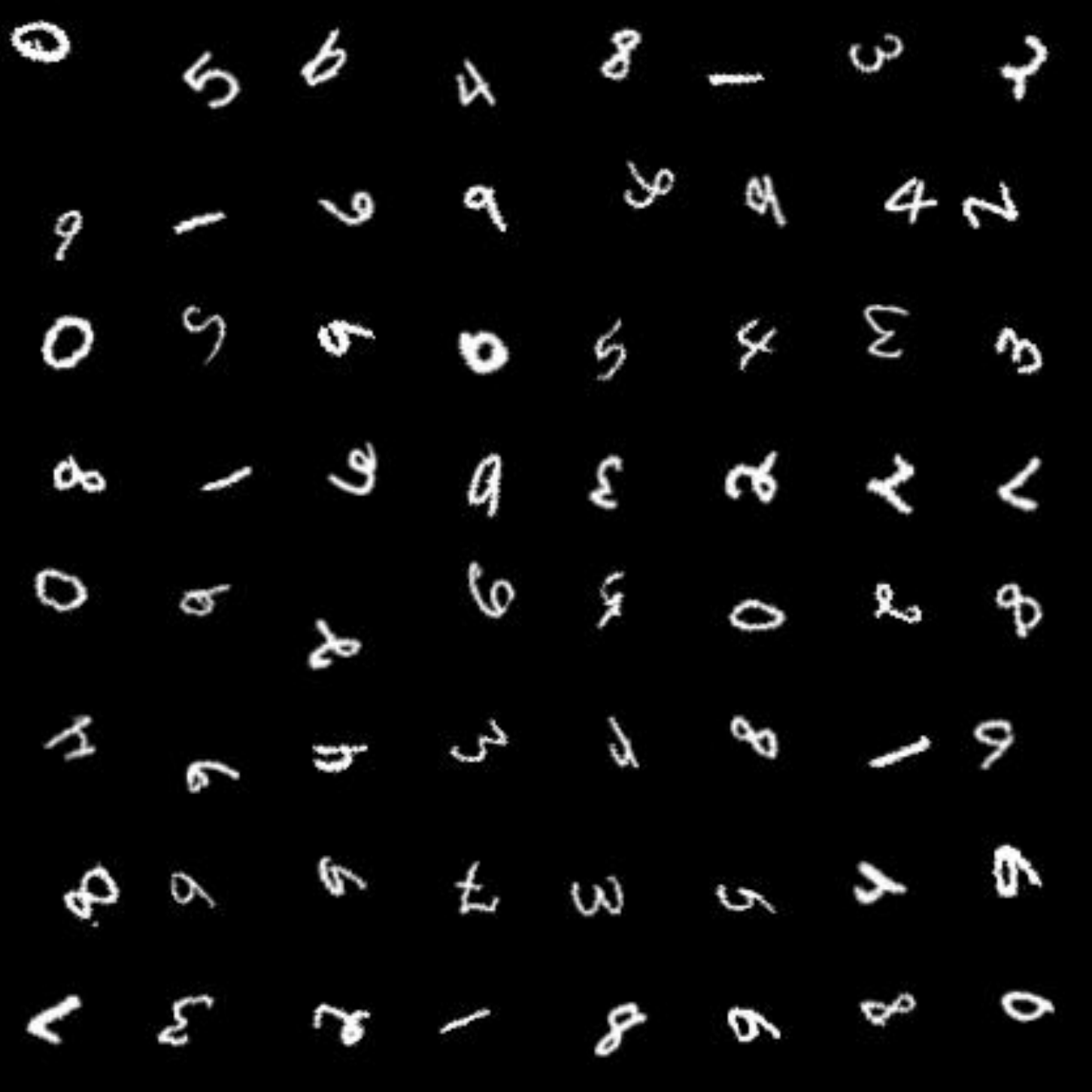}
                \captionsetup{labelformat=empty}
                \caption{Input Ground Truth}
                % \hspace{1mm}
        \end{subfigure}%
        % \\ [1mm]
        \begin{subfigure}[b]{0.43\linewidth}
                \centering
                \includegraphics[width=.9\linewidth]{image/Appendix/recon1/mnist_input.pdf}
                \captionsetup{labelformat=empty}
                \caption{Output Reconstruction}
        \end{subfigure}%
        \\ [1mm]
        \begin{subfigure}[b]{0.43\linewidth}
                \centering
                \includegraphics[width=.9\linewidth]{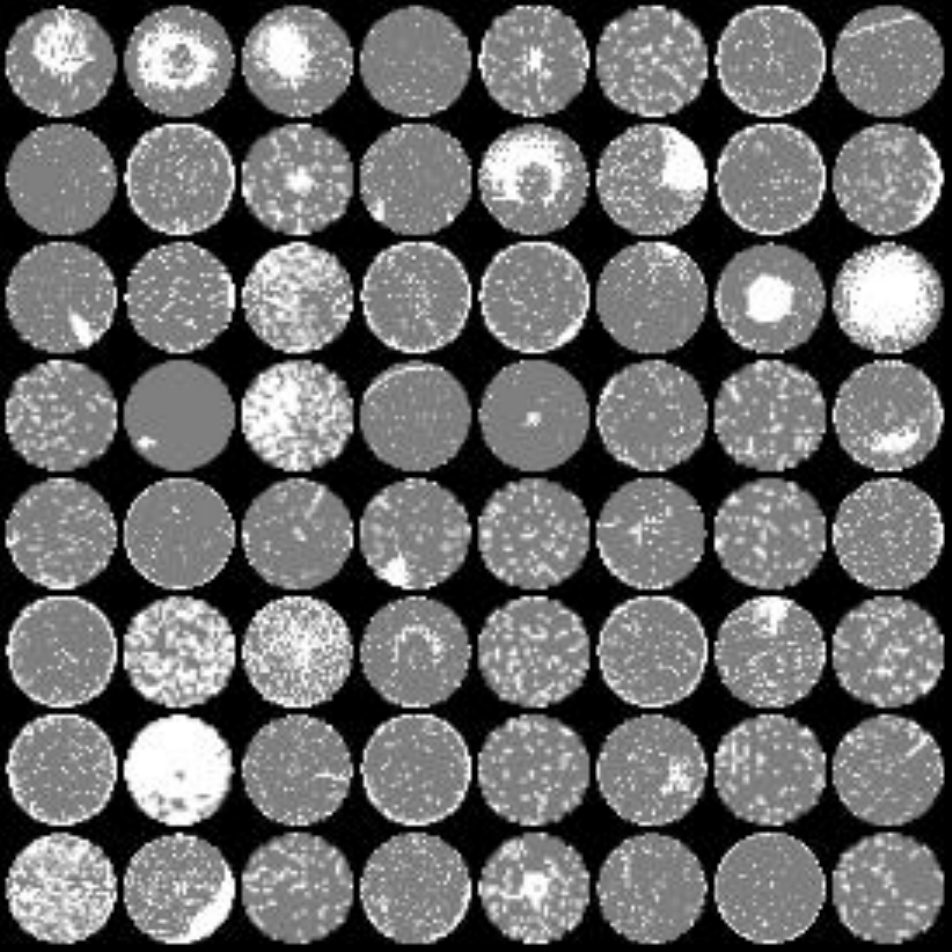}
                \captionsetup{labelformat=empty}
                \caption{Input Ground Truth}
        \end{subfigure}%
        % \\ [1mm]
        \begin{subfigure}[b]{0.43\linewidth}
                \centering
                \includegraphics[width=.9\linewidth]{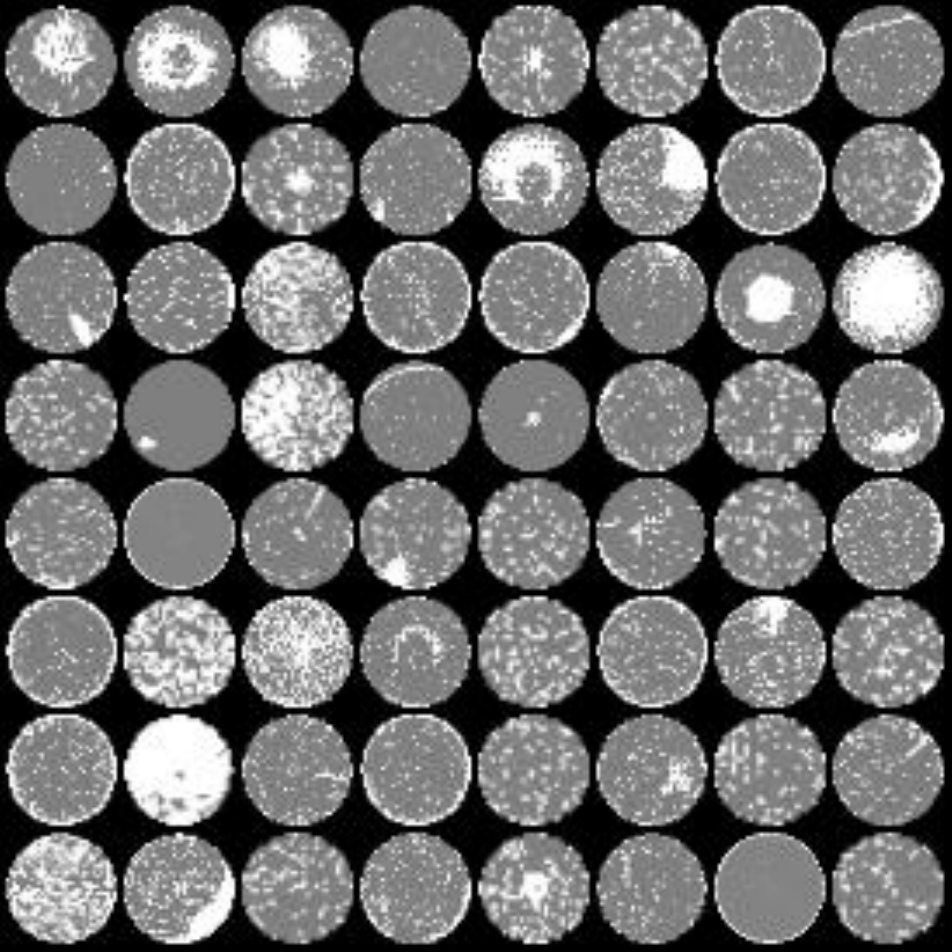}
                \captionsetup{labelformat=empty}
                \caption{Output Reconstruction}
        \end{subfigure}%
        \\ [1mm]
        \begin{subfigure}[b]{0.43\linewidth}
                \centering
                \includegraphics[width=.9\linewidth]{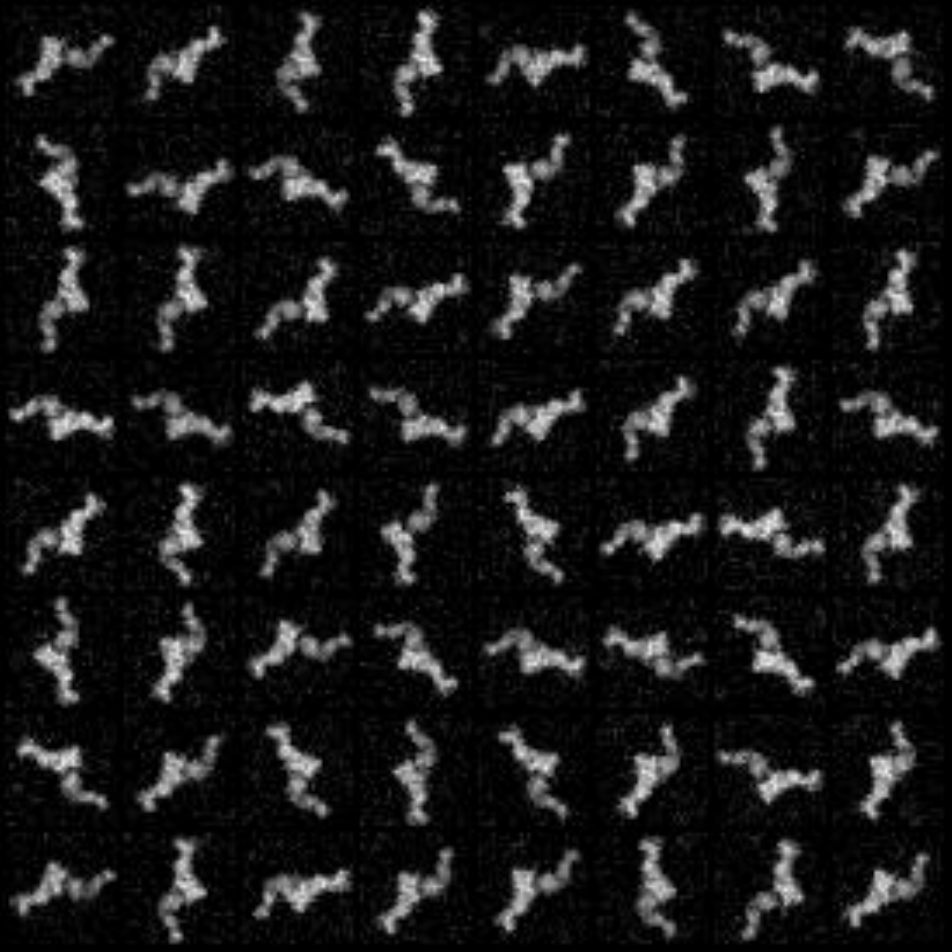}
                \captionsetup{labelformat=empty}
                \caption{Input Ground Truth}
        \end{subfigure}%
        % \\ [1mm]
        \begin{subfigure}[b]{0.43\linewidth}
                \centering
                \includegraphics[width=.9\linewidth]{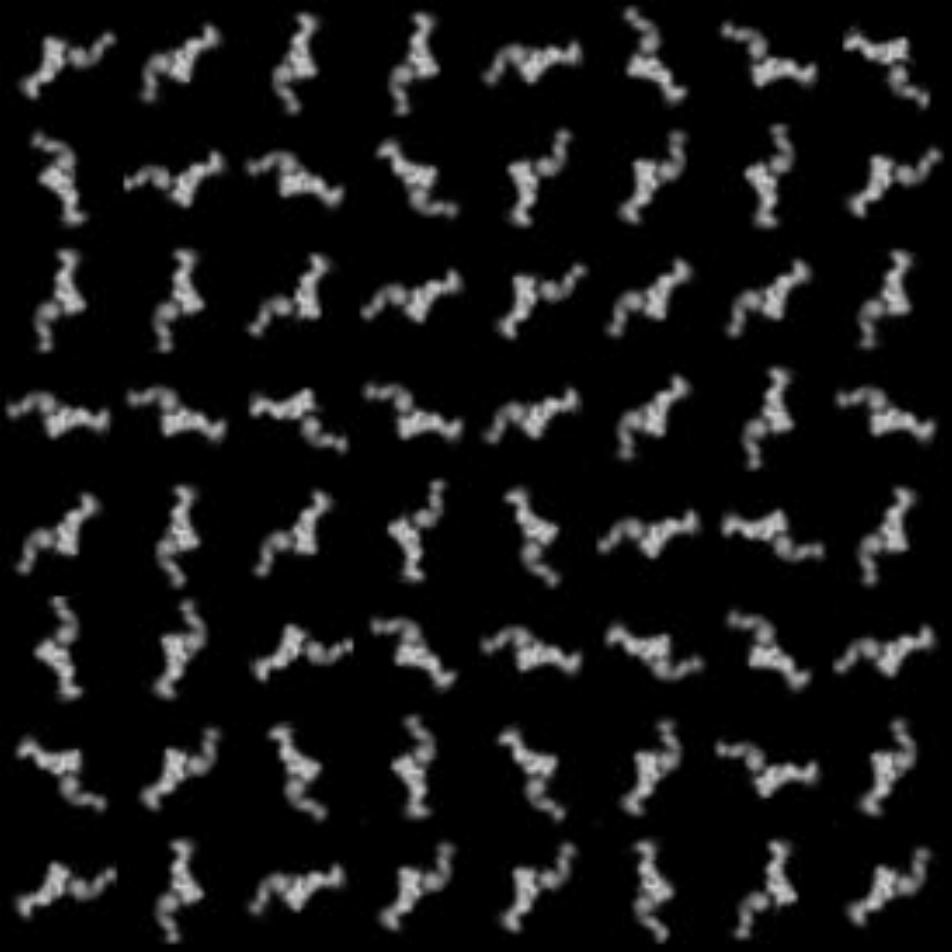}
                \captionsetup{labelformat=empty}
                \caption{Output Reconstruction}
        \end{subfigure}%
\end{figure}
\begin{figure}\ContinuedFloat
                \centering
        \begin{subfigure}[b]{0.43\linewidth}
                \centering
                \includegraphics[width=.9\linewidth]{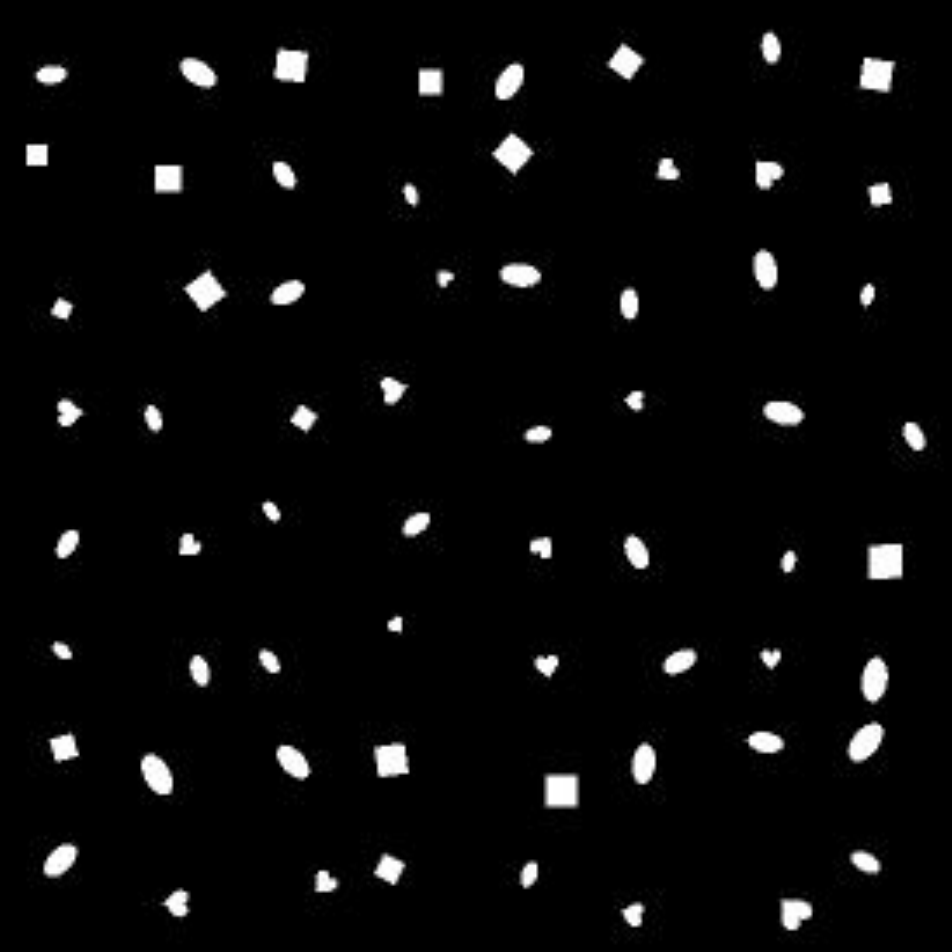}
                \captionsetup{labelformat=empty}
                \caption{Input Ground Truth}
        \end{subfigure}%
        % \\ [1mm]
        \begin{subfigure}[b]{0.43\linewidth}
                \centering
                \includegraphics[width=.9\linewidth]{image/Appendix/recon1/dsprites_input.pdf}
                \captionsetup{labelformat=empty}
                \caption{Output Reconstruction}
        \end{subfigure}%
        \\ [1mm]
        \begin{subfigure}[b]{0.43\linewidth}
                \centering
                \includegraphics[width=.9\linewidth]{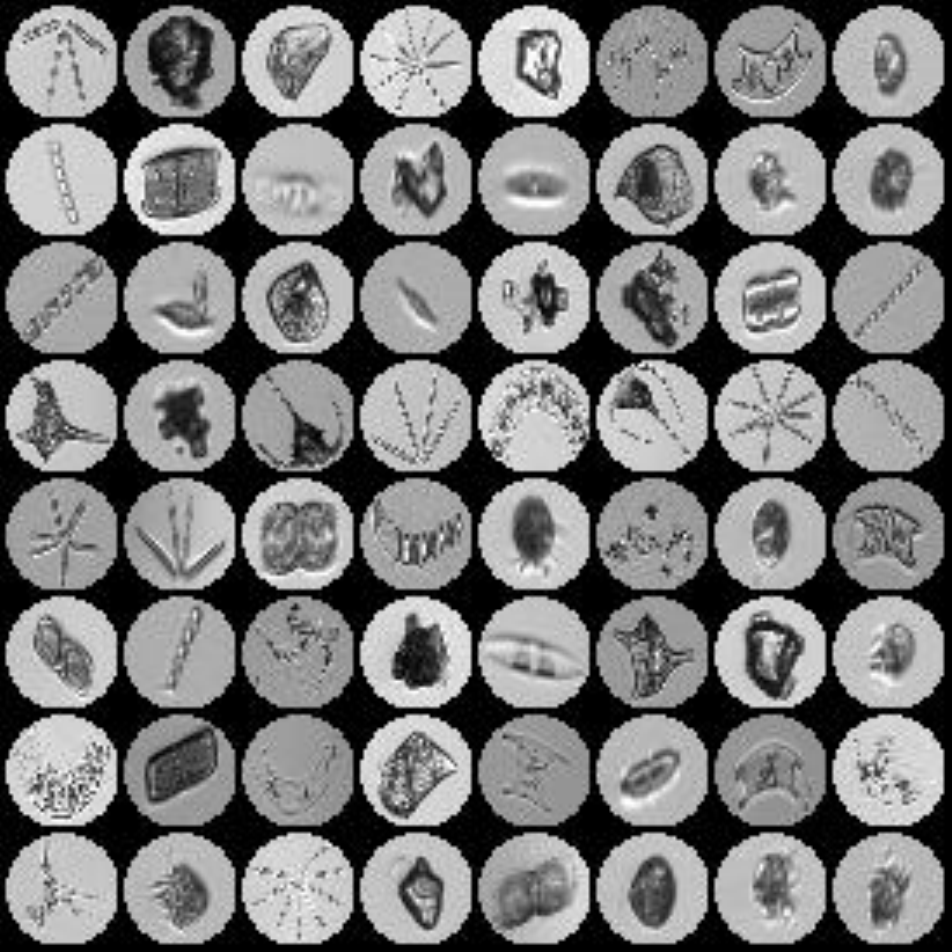}
                \captionsetup{labelformat=empty}
                \caption{Input Ground Truth}
        \end{subfigure}%
        % \\ [1mm]
        \begin{subfigure}[b]{0.43\linewidth}
                \centering
                \includegraphics[width=.9\linewidth]{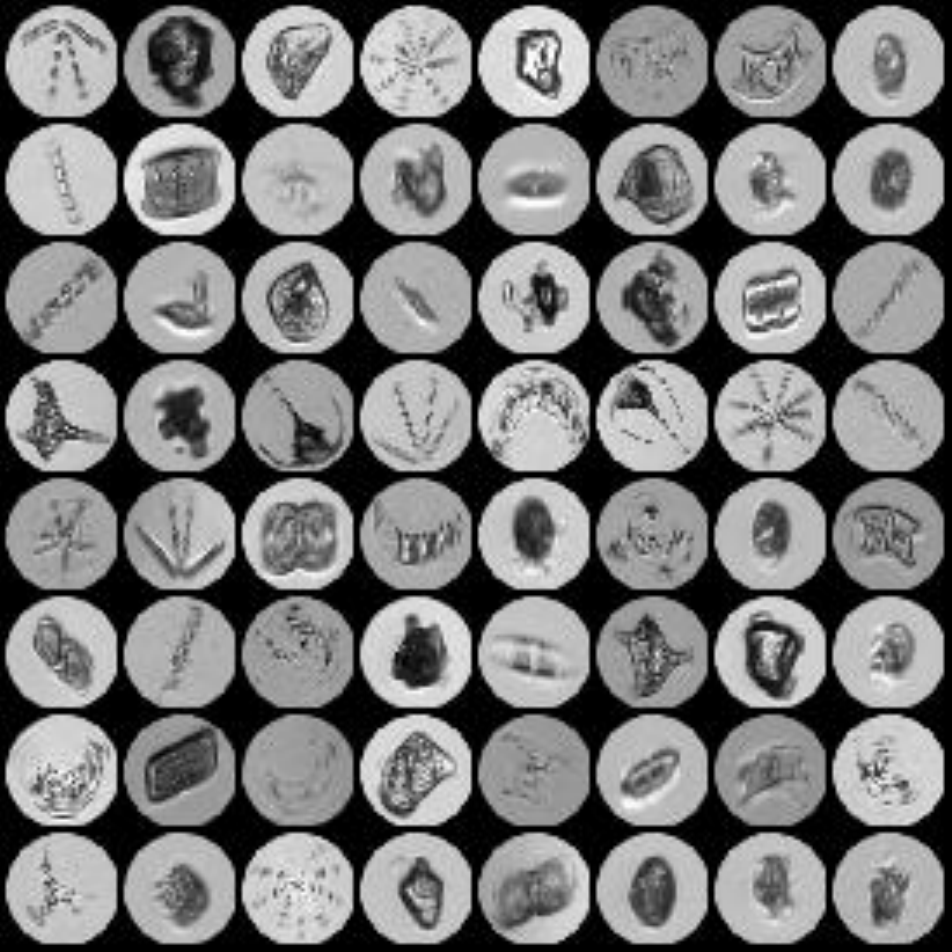}
                \captionsetup{labelformat=empty}
                \caption{Output Reconstruction}
        \end{subfigure}%
        \\ [1mm]
        \begin{subfigure}[b]{0.43\linewidth}
                \centering
                \includegraphics[width=.9\linewidth]{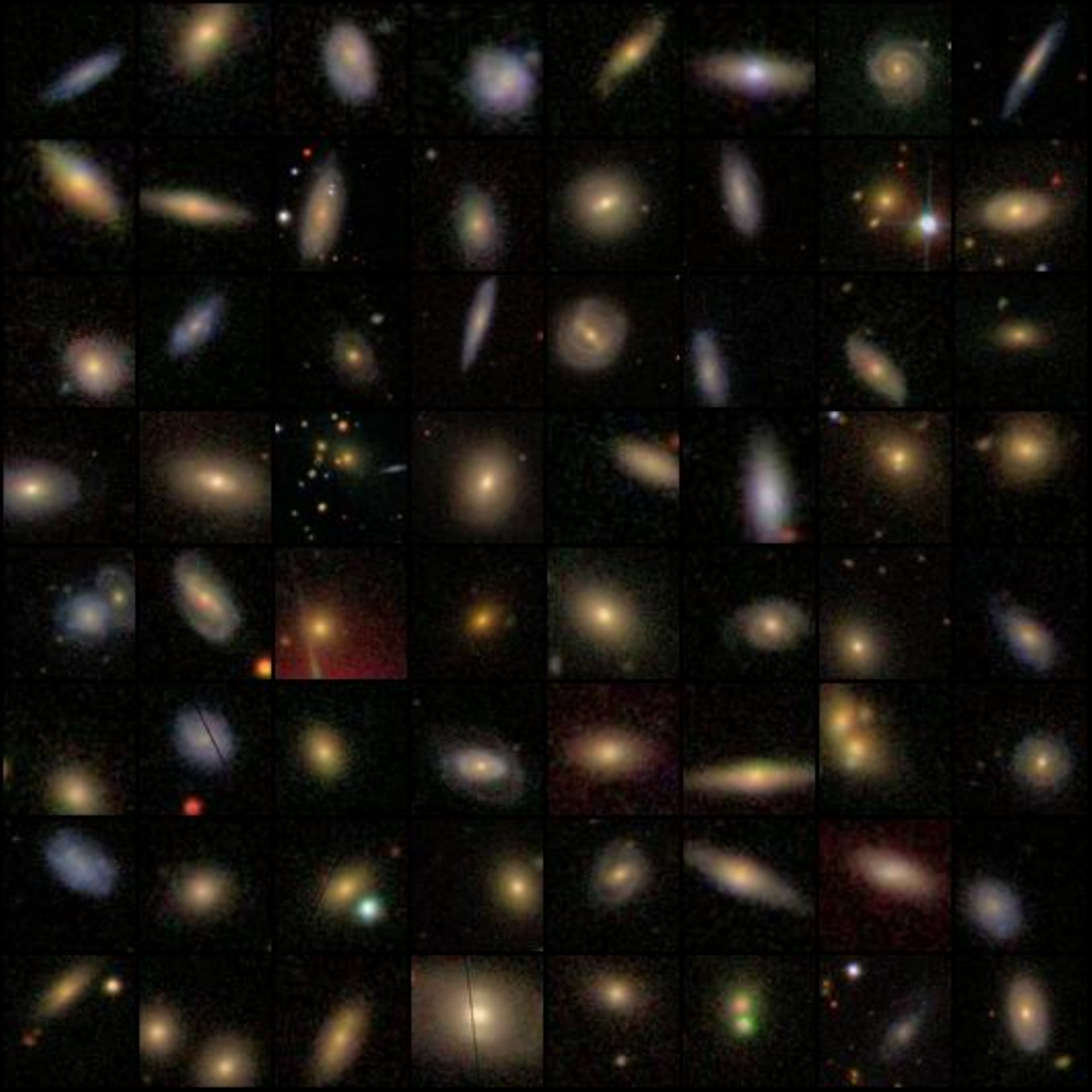}
                \captionsetup{labelformat=empty}
                \caption{Input Ground Truth}
        \end{subfigure}%
        % \\ [1mm]
        \begin{subfigure}[b]{0.43\linewidth}
                \centering
                \includegraphics[width=.9\linewidth]{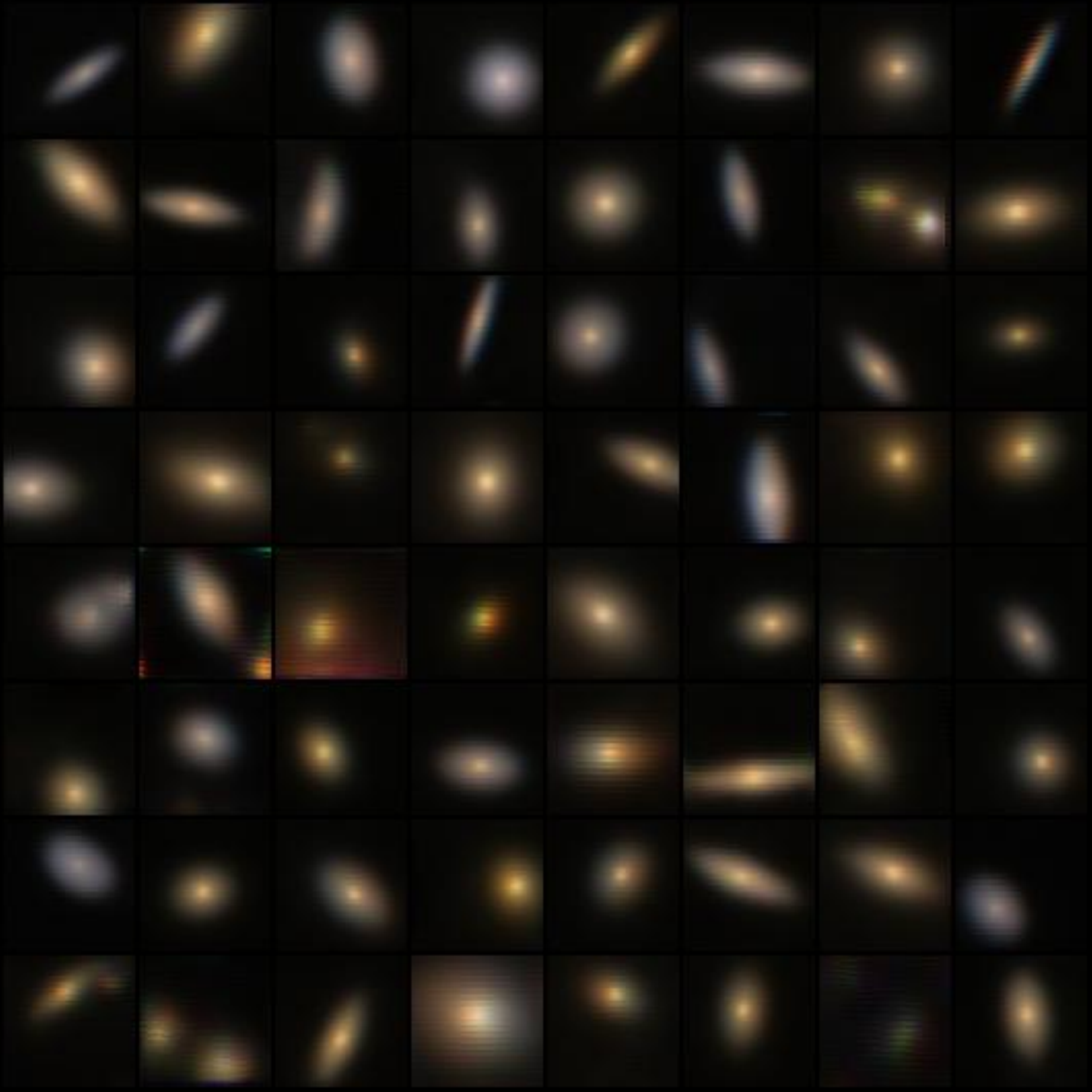}
                \captionsetup{labelformat=empty}
                \caption{Output Reconstruction}
        \end{subfigure}%
        \vspace{-5mm}
        \caption{The output images (Right) are reconstructed very similar to the input images (Left).}
        \label{Appendix:Reconstruct J image}
\end{figure}

\section{Reconstructing $J^{\text{(can)}}$}
\label{Appendix: Reconstruct J can}
In this section, we show image samples demonstrating that IRL-INR does obtain an invariant representation of the input image $J$ regardless of its orientation.
Specifically, we show that when the INR network $\dec(\cdot,\cdot;\eta)$ is provided with non-transformed coordinates (or when $\hat{\theta},\hat{\tau}$ is ignored), the input $R_\theta[J]$ with any $\theta\in[0,2\pi)$ is reconstructed into the same canonical orientation $J^{\text{(can)}}$.

\subsection{Image generation process}
The Encoder $\enc_\phi$ outputs the rotation representation $\hat{\theta}$, translation representation $\hat{\tau}$, and semantic representation $z$. Hypernetwork $\hyp_\psi$ takes $z$ as an input and then outputs $\eta$, where $\eta$ is the set of weights and biases of INR network. We ignore the rotation and translation representations $\hat{\theta}$ and $\hat{\tau}$, so $\dec(x_p,y_p;\eta) \approx J^\text{(can)}_p$.

% \sh{Following this procedure, IRL-INR can reconstruct $J^{\text{can}}$ from arbitrarily rotated and translated input images. We can see that the reconstructions are invariant with respect to the rotations and translations (\cref{Appendix: Reconstruct J can}).} 

% Since IRL-INR is trained to be $\dec(\cdot,\cdot;\eta)\approx \mathcal{I}(\cdot,\cdot)$, we get $\dec(x_p,y_p;\eta) \approx J^\text{(can)}_p$. 
\begin{figure}[h]
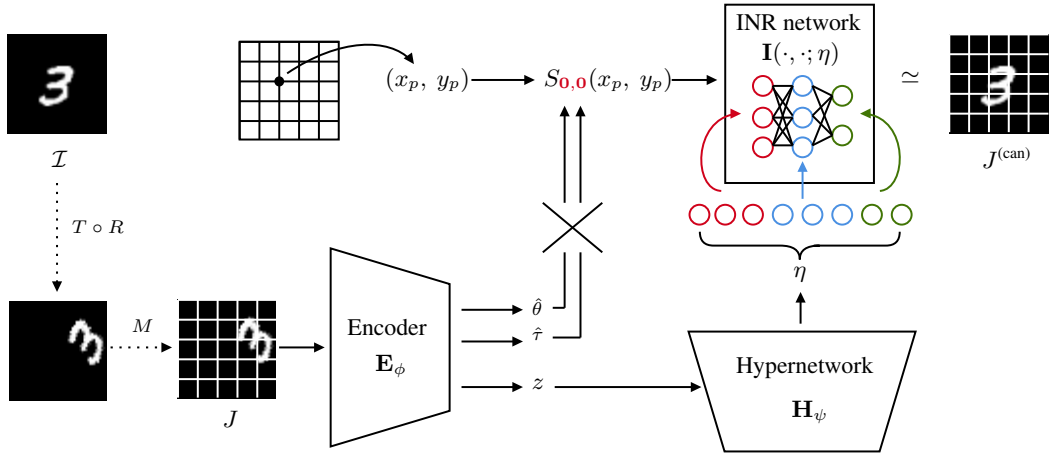


\tikzset{every picture/.style={line width=0.75pt}} %set default line width to 0.75pt        

\tikzset{every picture/.style={line width=0.75pt}} %set default line width to 0.75pt        

\begin{tikzpicture}[x=0.75pt,y=0.75pt,yscale=-1,xscale=1]
%uncomment if require: \path (0,638); %set diagram left start at 0, and has height of 638

%Image [id:dp36094664519349107] 
\draw (110.5,174) node  {\includegraphics[width=36.75pt,height=37.5pt]{image/J.png}};
%Straight Lines [id:da48351678587254077] 
\draw  [dash pattern={on 0.84pt off 2.51pt}]  (52,173) -- (80,173) ;
\draw [shift={(83,173)}, rotate = 180] [fill={rgb, 255:red, 0; green, 0; blue, 0 }  ][line width=0.08]  [draw opacity=0] (5.36,-2.57) -- (0,0) -- (5.36,2.57) -- cycle    ;
%Straight Lines [id:da5381190519578207] 
\draw    (137,173) -- (157,173) ;
\draw [shift={(160,173)}, rotate = 180] [fill={rgb, 255:red, 0; green, 0; blue, 0 }  ][line width=0.08]  [draw opacity=0] (5.36,-2.57) -- (0,0) -- (5.36,2.57) -- cycle    ;
%Shape: Trapezoid [id:dp6933398854347409] 
\draw   (163,123) -- (223,141) -- (223,205) -- (163,223) -- cycle ;
%Straight Lines [id:da36174808681524095] 
\draw    (229,154) -- (257,154) ;
\draw [shift={(260,154)}, rotate = 180] [fill={rgb, 255:red, 0; green, 0; blue, 0 }  ][line width=0.08]  [draw opacity=0] (5.36,-2.57) -- (0,0) -- (5.36,2.57) -- cycle    ;
%Straight Lines [id:da13710175173779848] 
\draw    (229,193) -- (257,193) ;
\draw [shift={(260,193)}, rotate = 180] [fill={rgb, 255:red, 0; green, 0; blue, 0 }  ][line width=0.08]  [draw opacity=0] (5.36,-2.57) -- (0,0) -- (5.36,2.57) -- cycle    ;
%Straight Lines [id:da6763990160591609] 
\draw    (229,170) -- (251,170) -- (257,170) ;
\draw [shift={(260,170)}, rotate = 180] [fill={rgb, 255:red, 0; green, 0; blue, 0 }  ][line width=0.08]  [draw opacity=0] (5.36,-2.57) -- (0,0) -- (5.36,2.57) -- cycle    ;
%Straight Lines [id:da30538313471709533] 
\draw    (281,100) -- (281,54) ;
\draw [shift={(281,51)}, rotate = 90] [fill={rgb, 255:red, 0; green, 0; blue, 0 }  ][line width=0.08]  [draw opacity=0] (5.36,-2.57) -- (0,0) -- (5.36,2.57) -- cycle    ;
%Straight Lines [id:da7861366817587461] 
\draw    (275,153) -- (281,153) ;
%Straight Lines [id:da7657912611771184] 
\draw    (289,100) -- (289,54) ;
\draw [shift={(289,51)}, rotate = 90] [fill={rgb, 255:red, 0; green, 0; blue, 0 }  ][line width=0.08]  [draw opacity=0] (5.36,-2.57) -- (0,0) -- (5.36,2.57) -- cycle    ;
%Straight Lines [id:da6082345678471073] 
\draw    (275,168) -- (289,168) ;
%Straight Lines [id:da9508852027273481] 
\draw    (234,38) -- (263,38) ;
\draw [shift={(266,38)}, rotate = 180] [fill={rgb, 255:red, 0; green, 0; blue, 0 }  ][line width=0.08]  [draw opacity=0] (5.36,-2.57) -- (0,0) -- (5.36,2.57) -- cycle    ;
%Shape: Trapezoid [id:dp6125354363883906] 
\draw   (458,165) -- (439.85,225.5) -- (361.65,225.5) -- (343.5,165) -- cycle ;
%Straight Lines [id:da5038184043481221] 
\draw    (276,193) -- (347,193) ;
\draw [shift={(350,193)}, rotate = 180] [fill={rgb, 255:red, 0; green, 0; blue, 0 }  ][line width=0.08]  [draw opacity=0] (5.36,-2.57) -- (0,0) -- (5.36,2.57) -- cycle    ;
%Shape: Rectangle [id:dp6484774994965055] 
\draw   (362,0) -- (437,0) -- (437,90) -- (362,90) -- cycle ;
%Straight Lines [id:da7676104852060583] 
\draw    (335,38) -- (355,38) ;
\draw [shift={(358,38)}, rotate = 180] [fill={rgb, 255:red, 0; green, 0; blue, 0 }  ][line width=0.08]  [draw opacity=0] (5.36,-2.57) -- (0,0) -- (5.36,2.57) -- cycle    ;
%Image [id:dp6649595476896026] 
\draw (500,40) node  {\includegraphics[width=37.5pt,height=37.5pt]{image/I.png}};
%Shape: Brace [id:dp6098417311618172] 
\draw   (349.2,114.6) .. controls (349.2,119.27) and (351.53,121.6) .. (356.2,121.6) -- (389.6,121.6) .. controls (396.27,121.6) and (399.6,123.93) .. (399.6,128.6) .. controls (399.6,123.93) and (402.93,121.6) .. (409.6,121.6)(406.6,121.6) -- (443,121.6) .. controls (447.67,121.6) and (450,119.27) .. (450,114.6) ;
%Straight Lines [id:da34964773432151763] 
\draw    (400,161) -- (400,148) ;
\draw [shift={(400,145)}, rotate = 90] [fill={rgb, 255:red, 0; green, 0; blue, 0 }  ][line width=0.08]  [draw opacity=0] (5.36,-2.57) -- (0,0) -- (5.36,2.57) -- cycle    ;
%Shape: Circle [id:dp765703798326334] 
\draw  [color={rgb, 255:red, 208; green, 2; blue, 27 }  ,draw opacity=1 ] (344,106) .. controls (344,103.24) and (346.24,101) .. (349,101) .. controls (351.76,101) and (354,103.24) .. (354,106) .. controls (354,108.76) and (351.76,111) .. (349,111) .. controls (346.24,111) and (344,108.76) .. (344,106) -- cycle ;
%Shape: Circle [id:dp9683692324156519] 
\draw  [color={rgb, 255:red, 208; green, 2; blue, 27 }  ,draw opacity=1 ] (357,106) .. controls (357,103.24) and (359.24,101) .. (362,101) .. controls (364.76,101) and (367,103.24) .. (367,106) .. controls (367,108.76) and (364.76,111) .. (362,111) .. controls (359.24,111) and (357,108.76) .. (357,106) -- cycle ;
%Shape: Circle [id:dp3423680464348078] 
\draw  [color={rgb, 255:red, 208; green, 2; blue, 27 }  ,draw opacity=1 ] (371,106) .. controls (371,103.24) and (373.24,101) .. (376,101) .. controls (378.76,101) and (381,103.24) .. (381,106) .. controls (381,108.76) and (378.76,111) .. (376,111) .. controls (373.24,111) and (371,108.76) .. (371,106) -- cycle ;
%Shape: Circle [id:dp8639408354878916] 
\draw  [color={rgb, 255:red, 74; green, 144; blue, 226 }  ,draw opacity=1 ] (386,106) .. controls (386,103.24) and (388.24,101) .. (391,101) .. controls (393.76,101) and (396,103.24) .. (396,106) .. controls (396,108.76) and (393.76,111) .. (391,111) .. controls (388.24,111) and (386,108.76) .. (386,106) -- cycle ;
%Shape: Circle [id:dp06273770720341043] 
\draw  [color={rgb, 255:red, 74; green, 144; blue, 226 }  ,draw opacity=1 ] (401,106) .. controls (401,103.24) and (403.24,101) .. (406,101) .. controls (408.76,101) and (411,103.24) .. (411,106) .. controls (411,108.76) and (408.76,111) .. (406,111) .. controls (403.24,111) and (401,108.76) .. (401,106) -- cycle ;
%Shape: Circle [id:dp28945740594827984] 
\draw  [color={rgb, 255:red, 208; green, 2; blue, 27 }  ,draw opacity=1 ] (376,41) .. controls (376,38.24) and (378.24,36) .. (381,36) .. controls (383.76,36) and (386,38.24) .. (386,41) .. controls (386,43.76) and (383.76,46) .. (381,46) .. controls (378.24,46) and (376,43.76) .. (376,41) -- cycle ;
%Shape: Circle [id:dp8954968028605682] 
\draw  [color={rgb, 255:red, 208; green, 2; blue, 27 }  ,draw opacity=1 ] (376,57) .. controls (376,54.24) and (378.24,52) .. (381,52) .. controls (383.76,52) and (386,54.24) .. (386,57) .. controls (386,59.76) and (383.76,62) .. (381,62) .. controls (378.24,62) and (376,59.76) .. (376,57) -- cycle ;
%Shape: Circle [id:dp5484771141058965] 
\draw  [color={rgb, 255:red, 208; green, 2; blue, 27 }  ,draw opacity=1 ] (376,72) .. controls (376,69.24) and (378.24,67) .. (381,67) .. controls (383.76,67) and (386,69.24) .. (386,72) .. controls (386,74.76) and (383.76,77) .. (381,77) .. controls (378.24,77) and (376,74.76) .. (376,72) -- cycle ;
%Shape: Circle [id:dp8962157523481737] 
\draw  [color={rgb, 255:red, 74; green, 144; blue, 226 }  ,draw opacity=1 ] (416,106) .. controls (416,103.24) and (418.24,101) .. (421,101) .. controls (423.76,101) and (426,103.24) .. (426,106) .. controls (426,108.76) and (423.76,111) .. (421,111) .. controls (418.24,111) and (416,108.76) .. (416,106) -- cycle ;
%Shape: Circle [id:dp22388666326728834] 
\draw  [color={rgb, 255:red, 65; green, 117; blue, 5 }  ,draw opacity=1 ] (431,106) .. controls (431,103.24) and (433.24,101) .. (436,101) .. controls (438.76,101) and (441,103.24) .. (441,106) .. controls (441,108.76) and (438.76,111) .. (436,111) .. controls (433.24,111) and (431,108.76) .. (431,106) -- cycle ;
%Shape: Circle [id:dp10565607404415023] 
\draw  [color={rgb, 255:red, 65; green, 117; blue, 5 }  ,draw opacity=1 ] (446,106) .. controls (446,103.24) and (448.24,101) .. (451,101) .. controls (453.76,101) and (456,103.24) .. (456,106) .. controls (456,108.76) and (453.76,111) .. (451,111) .. controls (448.24,111) and (446,108.76) .. (446,106) -- cycle ;
%Shape: Circle [id:dp3380843088771198] 
\draw  [color={rgb, 255:red, 74; green, 144; blue, 226 }  ,draw opacity=1 ] (396,41) .. controls (396,38.24) and (398.24,36) .. (401,36) .. controls (403.76,36) and (406,38.24) .. (406,41) .. controls (406,43.76) and (403.76,46) .. (401,46) .. controls (398.24,46) and (396,43.76) .. (396,41) -- cycle ;
%Shape: Circle [id:dp22287424214790597] 
\draw  [color={rgb, 255:red, 74; green, 144; blue, 226 }  ,draw opacity=1 ] (396,57) .. controls (396,54.24) and (398.24,52) .. (401,52) .. controls (403.76,52) and (406,54.24) .. (406,57) .. controls (406,59.76) and (403.76,62) .. (401,62) .. controls (398.24,62) and (396,59.76) .. (396,57) -- cycle ;
%Shape: Circle [id:dp21151024201483437] 
\draw  [color={rgb, 255:red, 74; green, 144; blue, 226 }  ,draw opacity=1 ] (396,72) .. controls (396,69.24) and (398.24,67) .. (401,67) .. controls (403.76,67) and (406,69.24) .. (406,72) .. controls (406,74.76) and (403.76,77) .. (401,77) .. controls (398.24,77) and (396,74.76) .. (396,72) -- cycle ;
%Shape: Circle [id:dp28333452248089797] 
\draw  [color={rgb, 255:red, 65; green, 117; blue, 5 }  ,draw opacity=1 ] (416,46) .. controls (416,43.24) and (418.24,41) .. (421,41) .. controls (423.76,41) and (426,43.24) .. (426,46) .. controls (426,48.76) and (423.76,51) .. (421,51) .. controls (418.24,51) and (416,48.76) .. (416,46) -- cycle ;
%Shape: Circle [id:dp2110134099533797] 
\draw  [color={rgb, 255:red, 65; green, 117; blue, 5 }  ,draw opacity=1 ] (416,66) .. controls (416,63.24) and (418.24,61) .. (421,61) .. controls (423.76,61) and (426,63.24) .. (426,66) .. controls (426,68.76) and (423.76,71) .. (421,71) .. controls (418.24,71) and (416,68.76) .. (416,66) -- cycle ;
%Straight Lines [id:da8628859873131804] 
\draw    (386,41) -- (396,71) ;
%Straight Lines [id:da07738481760942673] 
\draw    (386,71) -- (396,41) ;
%Straight Lines [id:da15674543076471048] 
\draw    (386,57) -- (396,41) ;
%Straight Lines [id:da557691718873829] 
\draw    (386,71) -- (396,57) ;
%Straight Lines [id:da15689034470719831] 
\draw    (386,71) -- (396,71) ;
%Straight Lines [id:da9152793495623923] 
\draw    (386,57) -- (396,57) ;
%Straight Lines [id:da8895419102507541] 
\draw    (386,41) -- (396,41) ;
%Straight Lines [id:da5603304253404606] 
\draw    (396,71) -- (386,57) ;
%Straight Lines [id:da8201576551785246] 
\draw    (396,56) -- (386,42) ;
%Straight Lines [id:da8200627129169666] 
\draw    (416,66) -- (406,41) ;
%Straight Lines [id:da4700737595190887] 
\draw    (416,66) -- (406,56) ;
%Straight Lines [id:da15402528076133504] 
\draw    (416,66) -- (406,71) ;
%Straight Lines [id:da04016989600829057] 
\draw    (416,46) -- (406,41) ;
%Straight Lines [id:da41403161958488877] 
\draw    (406,56) -- (416,46) ;
%Straight Lines [id:da12848380695749184] 
\draw    (416,46) -- (406,71) ;
%Curve Lines [id:da39366718975494897] 
\draw [color={rgb, 255:red, 208; green, 2; blue, 27 }  ,draw opacity=1 ]   (360,94) .. controls (345.6,93.52) and (345.48,58.02) .. (367.17,55.18) ;
\draw [shift={(370,55)}, rotate = 180] [fill={rgb, 255:red, 208; green, 2; blue, 27 }  ,fill opacity=1 ][line width=0.08]  [draw opacity=0] (5.36,-2.57) -- (0,0) -- (5.36,2.57) -- cycle    ;
%Straight Lines [id:da9962199227105886] 
\draw [color={rgb, 255:red, 74; green, 144; blue, 226 }  ,draw opacity=1 ]   (401,98) -- (401,82) ;
\draw [shift={(401,79)}, rotate = 90] [fill={rgb, 255:red, 74; green, 144; blue, 226 }  ,fill opacity=1 ][line width=0.08]  [draw opacity=0] (5.36,-2.57) -- (0,0) -- (5.36,2.57) -- cycle    ;
%Curve Lines [id:da5173143419828584] 
\draw [color={rgb, 255:red, 65; green, 117; blue, 5 }  ,draw opacity=1 ]   (440,94) .. controls (453.92,93.04) and (454.94,57.52) .. (432.88,55.12) ;
\draw [shift={(430,55)}, rotate = 358.85] [fill={rgb, 255:red, 65; green, 117; blue, 5 }  ,fill opacity=1 ][line width=0.08]  [draw opacity=0] (5.36,-2.57) -- (0,0) -- (5.36,2.57) -- cycle    ;
%Image [id:dp06715216262310708] 
\draw (25,40) node  {\includegraphics[width=37.5pt,height=37.5pt]{image/I.png}};
%Straight Lines [id:da20752232802203552] 
\draw  [dash pattern={on 0.84pt off 2.51pt}]  (25,90) -- (25,142) ;
\draw [shift={(25,145)}, rotate = 270] [fill={rgb, 255:red, 0; green, 0; blue, 0 }  ][line width=0.08]  [draw opacity=0] (5.36,-2.57) -- (0,0) -- (5.36,2.57) -- cycle    ;
%Shape: Grid [id:dp5926366725817469] 
\draw  [draw opacity=0] (86,149) -- (136,149) -- (136,198.33) -- (86,198.33) -- cycle ; \draw  [color={rgb, 255:red, 255; green, 255; blue, 255 }  ,draw opacity=1 ] (86,149) -- (86,198.33)(96,149) -- (96,198.33)(106,149) -- (106,198.33)(116,149) -- (116,198.33)(126,149) -- (126,198.33) ; \draw  [color={rgb, 255:red, 255; green, 255; blue, 255 }  ,draw opacity=1 ] (86,149) -- (136,149)(86,159) -- (136,159)(86,169) -- (136,169)(86,179) -- (136,179)(86,189) -- (136,189) ; \draw  [color={rgb, 255:red, 255; green, 255; blue, 255 }  ,draw opacity=1 ]  ;
%Shape: Grid [id:dp16171381974120214] 
\draw  [draw opacity=0] (475,15) -- (525,15) -- (525,64.33) -- (475,64.33) -- cycle ; \draw  [color={rgb, 255:red, 255; green, 255; blue, 255 }  ,draw opacity=1 ] (475,15) -- (475,64.33)(485,15) -- (485,64.33)(495,15) -- (495,64.33)(505,15) -- (505,64.33)(515,15) -- (515,64.33) ; \draw  [color={rgb, 255:red, 255; green, 255; blue, 255 }  ,draw opacity=1 ] (475,15) -- (525,15)(475,25) -- (525,25)(475,35) -- (525,35)(475,45) -- (525,45)(475,55) -- (525,55) ; \draw  [color={rgb, 255:red, 255; green, 255; blue, 255 }  ,draw opacity=1 ]  ;
%Image [id:dp9213887313206829] 
\draw (25.5,175) node  {\includegraphics[width=36.75pt,height=37.5pt]{image/J.png}};
%Shape: Ellipse [id:dp9386365533780958] 
\draw  [fill={rgb, 255:red, 0; green, 0; blue, 0 }  ,fill opacity=1 ] (135.1,38.73) .. controls (135.1,37.59) and (135.99,36.67) .. (137.08,36.67) .. controls (138.17,36.67) and (139.05,37.59) .. (139.05,38.73) .. controls (139.05,39.86) and (138.17,40.78) .. (137.08,40.78) .. controls (135.99,40.78) and (135.1,39.86) .. (135.1,38.73) -- cycle ;
%Curve Lines [id:da8106363615038371] 
\draw    (141.05,34.67) .. controls (167.66,16.6) and (192.61,16.85) .. (203.16,27.78) ;
\draw [shift={(205,30)}, rotate = 234.75] [fill={rgb, 255:red, 0; green, 0; blue, 0 }  ][line width=0.08]  [draw opacity=0] (5.36,-2.57) -- (0,0) -- (5.36,2.57) -- cycle    ;
%Shape: Grid [id:dp5800930116279384] 
\draw  [draw opacity=1] (117.08,18.73) -- (167.08,18.73) -- (167.08,68.06) -- (117.08,68.06) -- cycle ; \draw  [color={rgb, 255:red, 0; green, 0; blue, 0 }  ,draw opacity=1 ] (117.08,18.73) -- (117.08,68.06)(127.08,18.73) -- (127.08,68.06)(137.08,18.73) -- (137.08,68.06)(147.08,18.73) -- (147.08,68.06)(157.08,18.73) -- (157.08,68.06) ; \draw  [color={rgb, 255:red, 0; green, 0; blue, 0 }  ,draw opacity=1 ] (117.08,18.73) -- (167.08,18.73)(117.08,28.73) -- (167.08,28.73)(117.08,38.73) -- (167.08,38.73)(117.08,48.73) -- (167.08,48.73)(117.08,58.73) -- (167.08,58.73) ; \draw  [color={rgb, 255:red, 0; green, 0; blue, 0 }  ,draw opacity=1 ]  ;
%Straight Lines [id:da6424075592991991] 
\draw    (281,153) -- (281,125) ;
%Straight Lines [id:da839117410093384] 
\draw    (289,168) -- (289,125) ;
%Straight Lines [id:da047424452771806624] 
\draw    (300,125) -- (270,100) ;
%Straight Lines [id:da9205775769408857] 
\draw    (300,100) -- (270,125) ;

% Text Node
\draw (21,72.4) node [anchor=north west][inner sep=0.75pt]  [font=\footnotesize]  {$\mathcal{I}$};
% Text Node
\draw (107,203.4) node [anchor=north west][inner sep=0.75pt]  [font=\footnotesize]  {$J$};
% Text Node
\draw (170,157) node [anchor=north west][inner sep=0.75pt]  [font=\footnotesize] [align=left] {Encoder};
% Text Node
\draw (184,175.4) node [anchor=north west][inner sep=0.75pt]  [font=\footnotesize]  {$\mathbf{E}_{\phi }$};
% Text Node
\draw (263,146.4) node [anchor=north west][inner sep=0.75pt]  [font=\scriptsize]  {$\hat{\theta }$};
% Text Node
\draw (263,161.4) node [anchor=north west][inner sep=0.75pt]  [font=\scriptsize]  {$\hat{\tau }$};
% Text Node
\draw (263,187.4) node [anchor=north west][inner sep=0.75pt]  [font=\footnotesize]  {$z$};
% Text Node
\draw (269,30.4) node [anchor=north west][inner sep=0.75pt]  [font=\footnotesize]  {$S_{\textcolor[rgb]{0.82,0.01,0.11}{\mathbf{0,0}}}( x_{p} ,\ y_{p})$};
% Text Node
\draw (189,30.4) node [anchor=north west][inner sep=0.75pt]  [font=\footnotesize]  {$( x_{p} ,\ y_{p})$};
% Text Node
\draw (366,176) node [anchor=north west][inner sep=0.75pt]  [font=\footnotesize] [align=left] {Hypernetwork};
% Text Node
\draw (394,198.4) node [anchor=north west][inner sep=0.75pt]  [font=\footnotesize]  {$\mathbf{H}_{\psi }$};
% Text Node
\draw (366,3) node [anchor=north west][inner sep=0.75pt]  [font=\footnotesize] [align=left] {INR network};
% Text Node
\draw (379,17.4) node [anchor=north west][inner sep=0.75pt]  [font=\footnotesize]  {$\mathbf{I}( \cdot ,\cdot ;\eta )$};
% Text Node
\draw (449,32.4) node [anchor=north west][inner sep=0.75pt]  [font=\footnotesize]  {$\simeq $};
% Text Node
\draw (395,131) node [anchor=north west][inner sep=0.75pt]  [font=\footnotesize]  {$\eta $};
% Text Node
\draw (61,157.4) node [anchor=north west][inner sep=0.75pt]  [font=\scriptsize]  {$M$};
% Text Node
\draw (31,107.4) node [anchor=north west][inner sep=0.75pt]  [font=\scriptsize]  {$T \circ R$};
\draw (490,70.4) node [anchor=north west][inner sep=0.75pt]  [font=\footnotesize]  {$J^{\text{(can)}}$};

\end{tikzpicture}

\caption{Using only $z$ for reconstruction $J^{\text{can}}$. Input NOT rotated and translated coordintates to INR network for generating $J^{\text{can}}$.} 

\end{figure}

\newpage
\subsection{Reconstruction results for $J^{\text{can}}$}

\begin{figure*}[h!]
\centering
        \begin{subfigure}[b]{\linewidth}
                \centering
                \includegraphics[width=.48\linewidth]{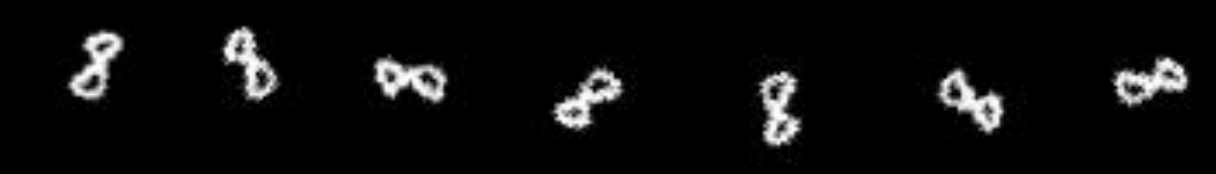}
                \hspace{2mm}
                \includegraphics[width=.48\linewidth]{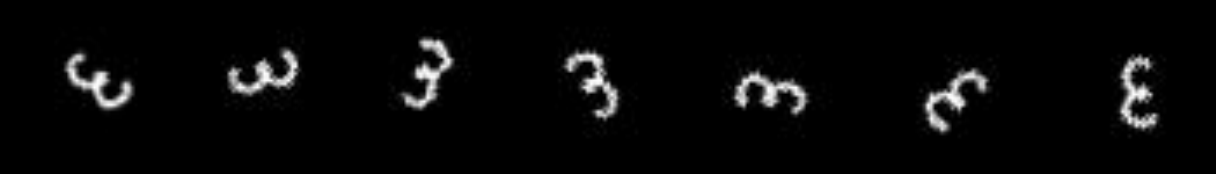}
                % \centering
                \\ \vspace{1mm}
                \includegraphics[width=.48\linewidth]{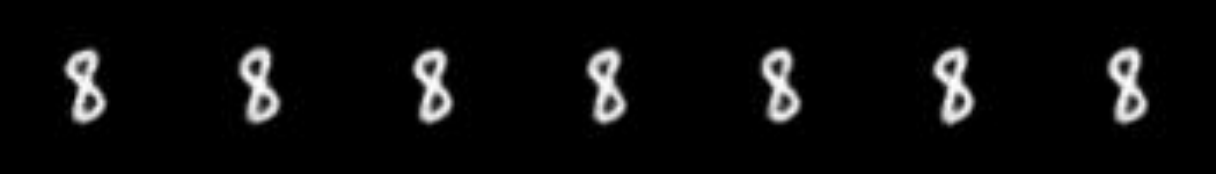}
                \hspace{2mm}
                \includegraphics[width=.48\linewidth]{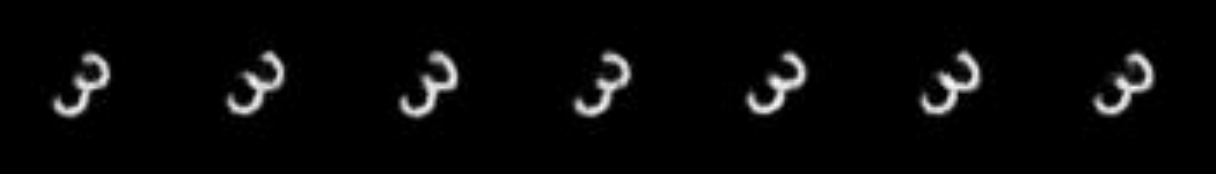}
                % \captionsetup{font=scriptsize}
                \caption{MNIST(U)}
        \end{subfigure}%
        \\ [1mm]
        \begin{subfigure}[b]{\linewidth}
                \centering
                \includegraphics[width=.48\linewidth]{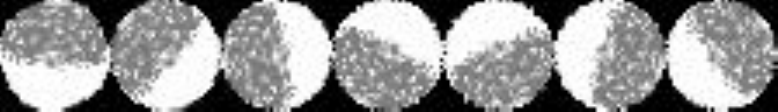}
                \hspace{2mm}
                \includegraphics[width=.48\linewidth]{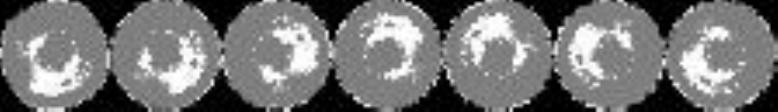}
                % \centering
                \\ \vspace{1mm}
                \includegraphics[width=.48\linewidth]{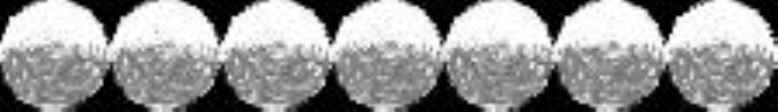}
                \hspace{2mm}
                \includegraphics[width=.48\linewidth]{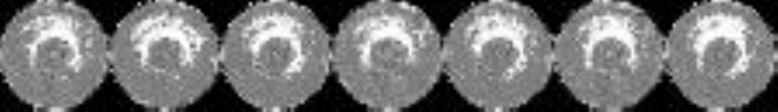}
                % \captionsetup{font=scriptsize}
                \caption{WM811k}
        \end{subfigure}%
        \\ [1mm]
        \begin{subfigure}[b]{\linewidth}
                \centering
                \includegraphics[width=.48\linewidth]{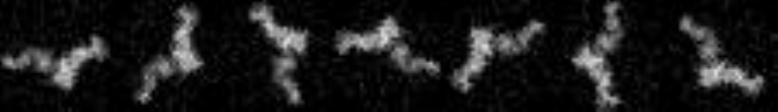}
                \hspace{2mm}
                \includegraphics[width=.48\linewidth]{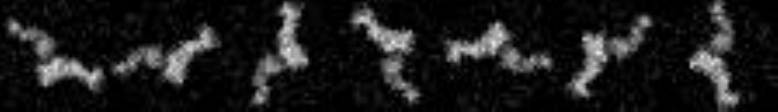}
                % \centering
                \\ \vspace{1mm}
                \includegraphics[width=.48\linewidth]{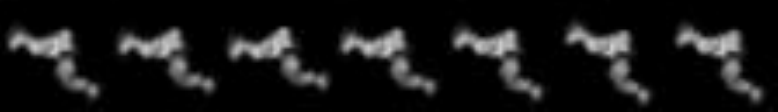}
                \hspace{2mm}
                \includegraphics[width=.48\linewidth]{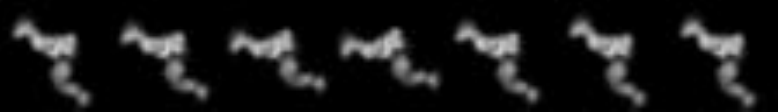}
                % \captionsetup{font=scriptsize}
                \caption{5HDB}
        \end{subfigure}%
        \\ [1mm]
        \begin{subfigure}[b]{\linewidth}
                \centering
                \includegraphics[width=.48\linewidth]{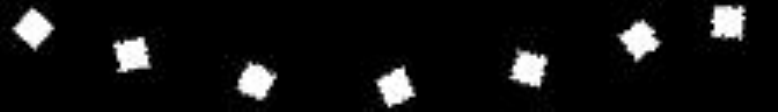}
                \hspace{2mm}
                \includegraphics[width=.48\linewidth]{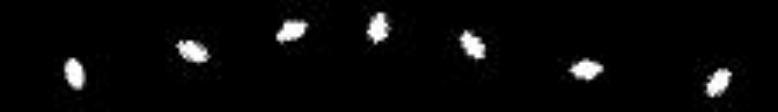}
                % \centering
                \\ \vspace{1mm}
                \includegraphics[width=.48\linewidth]{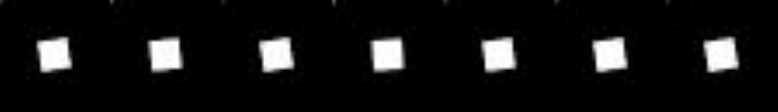}
                \hspace{2mm}
                \includegraphics[width=.48\linewidth]{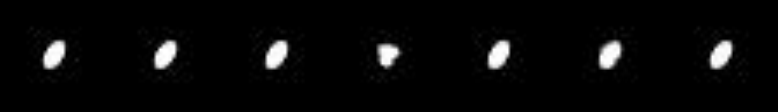}
                % \captionsetup{font=scriptsize}
                \caption{dSprites}
        \end{subfigure}%
        \\ [1mm]
        \begin{subfigure}[b]{\linewidth}
                \centering
                \includegraphics[width=.48\linewidth]{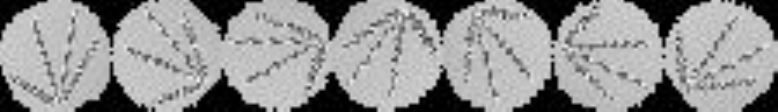}
                \hspace{2mm}
                \includegraphics[width=.48\linewidth]{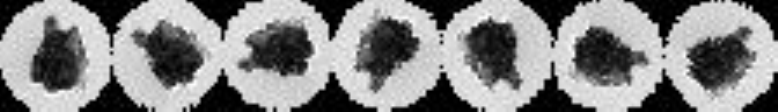}
                % \centering
                \\ \vspace{1mm}
                \includegraphics[width=.48\linewidth]{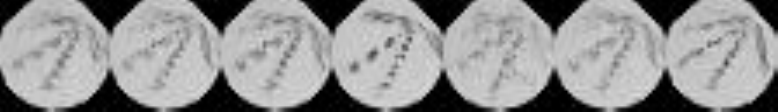}
                \hspace{2mm}
                \includegraphics[width=.48\linewidth]{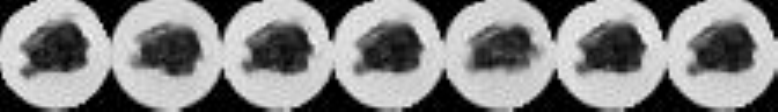}
                % \captionsetup{font=scriptsize}
                \caption{WHOI-Plankton}
        \end{subfigure}%
        \\ [1mm]
        \begin{subfigure}[b]{\linewidth}
                \centering
                \includegraphics[width=.48\linewidth]{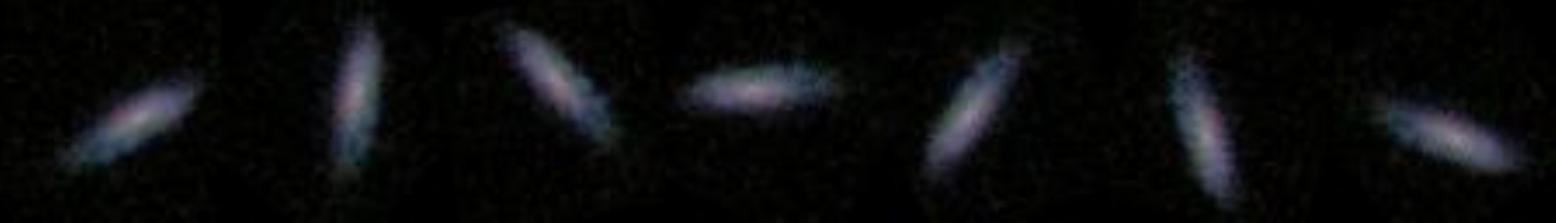}
                \hspace{2mm}
                \includegraphics[width=.48\linewidth]{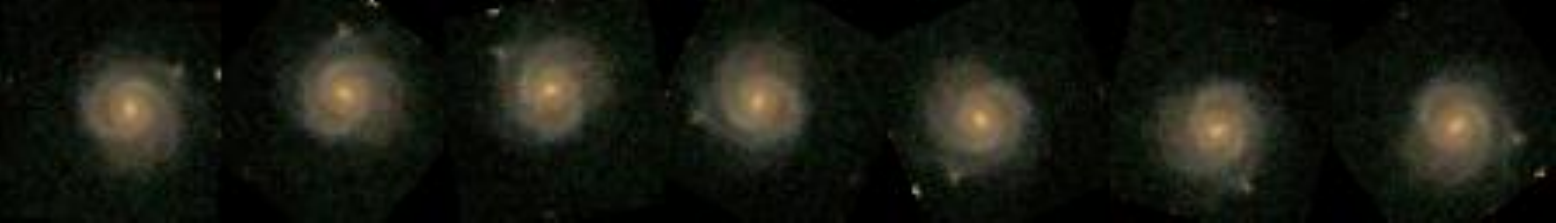}
                % \centering
                \\ \vspace{1mm}
                \includegraphics[width=.48\linewidth]{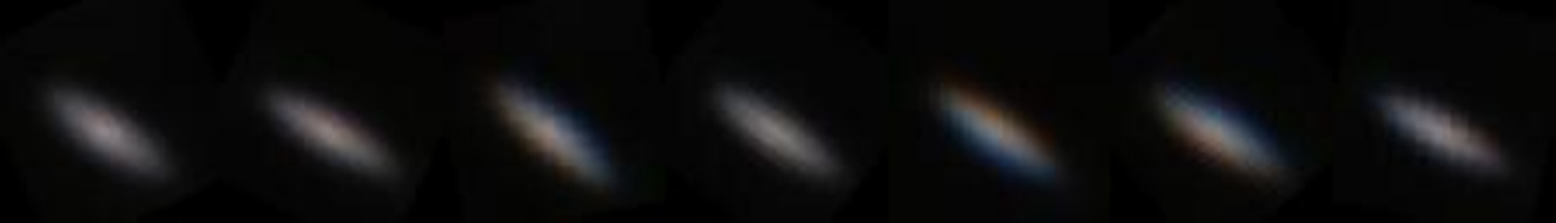}
                \hspace{2mm}
                \includegraphics[width=.48\linewidth]{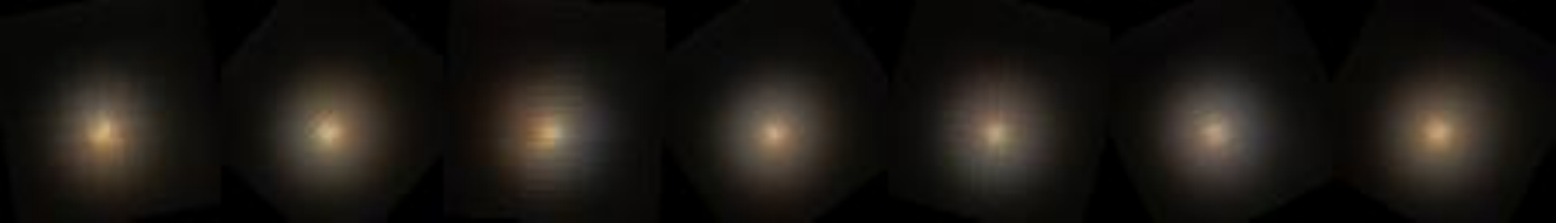}
                % \captionsetup{font=scriptsize}
                \caption{Galaxy Zoo}
        \end{subfigure}%
        \vspace{-5mm}
        \caption{To validate the disentanglement of semantic representations, we verify that the reconstructions are indeed invariant under rotation and translation. The first row of (a)--(f) are rotated by $\frac{2\pi}{7}$ degrees. The second row of (a)--(f) are reconstructions using only the semantic representation $z$, without any rotation or translation.}
        \label{Appendix:Reconstruct J can image}
\end{figure*}

\newpage
\section{Visualization of clustering results}
\begin{figure}[h!]
                \centering
                \includegraphics[width=.31\linewidth]{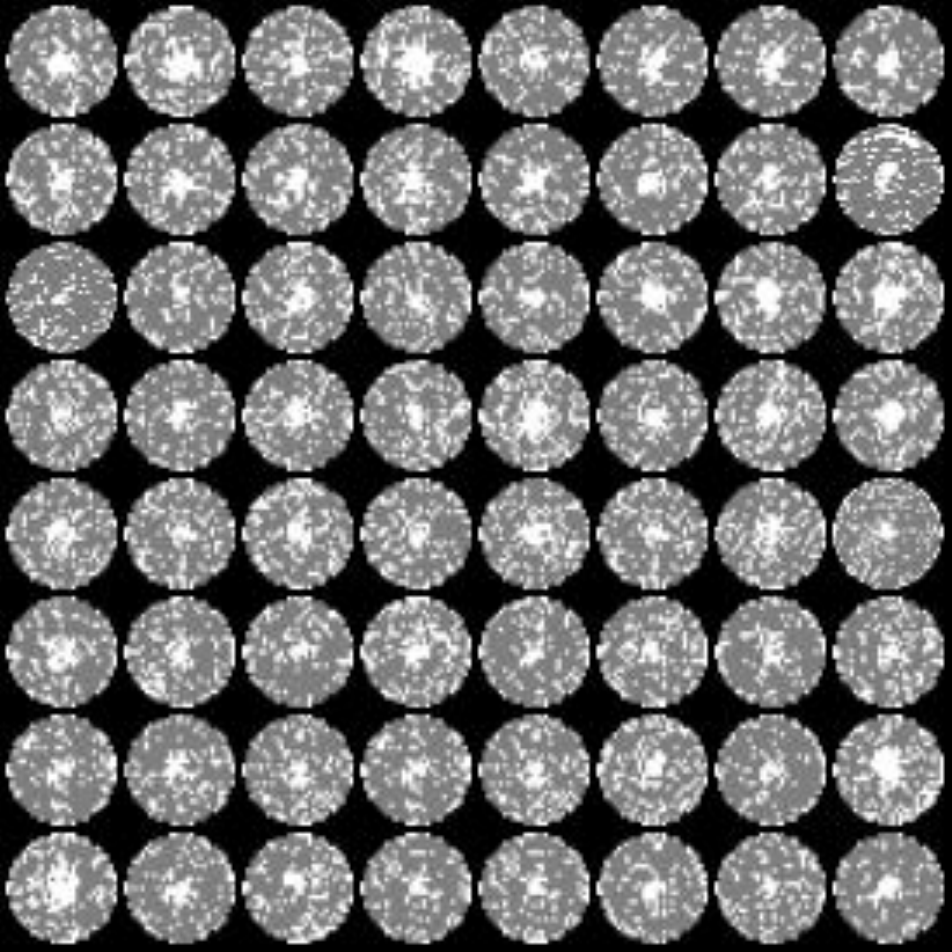}
                \includegraphics[width=.31\linewidth]{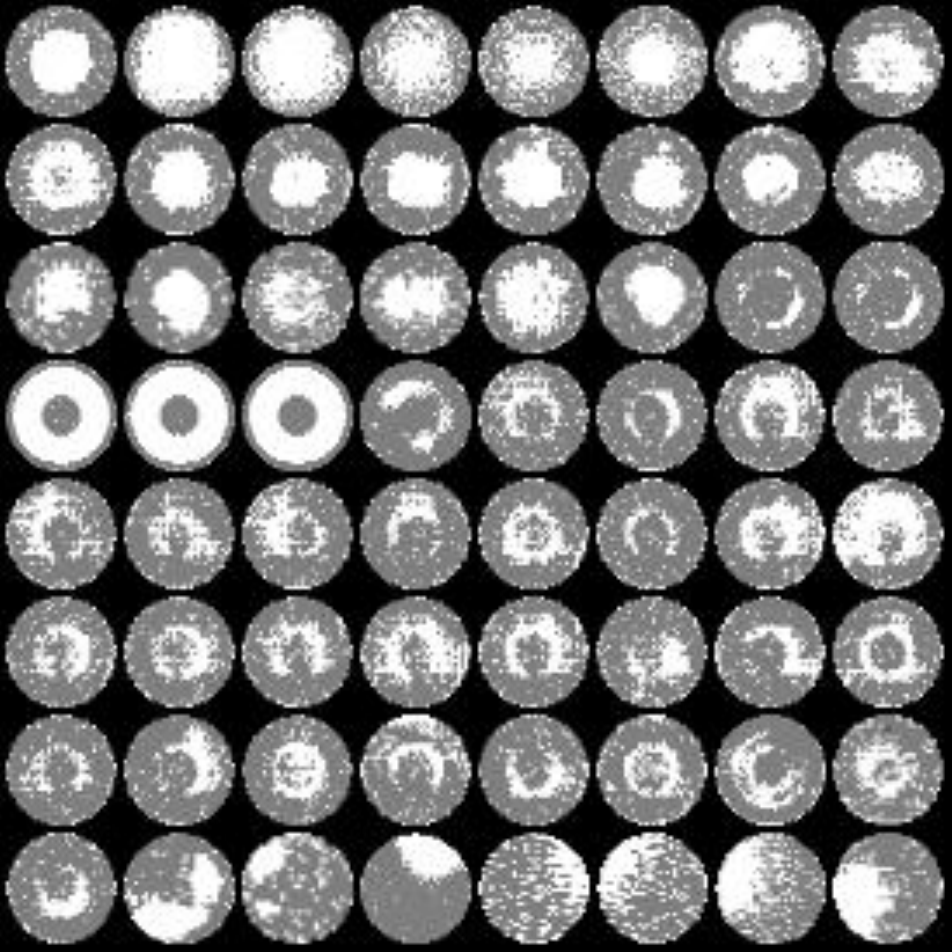}
                \includegraphics[width=.31\linewidth]{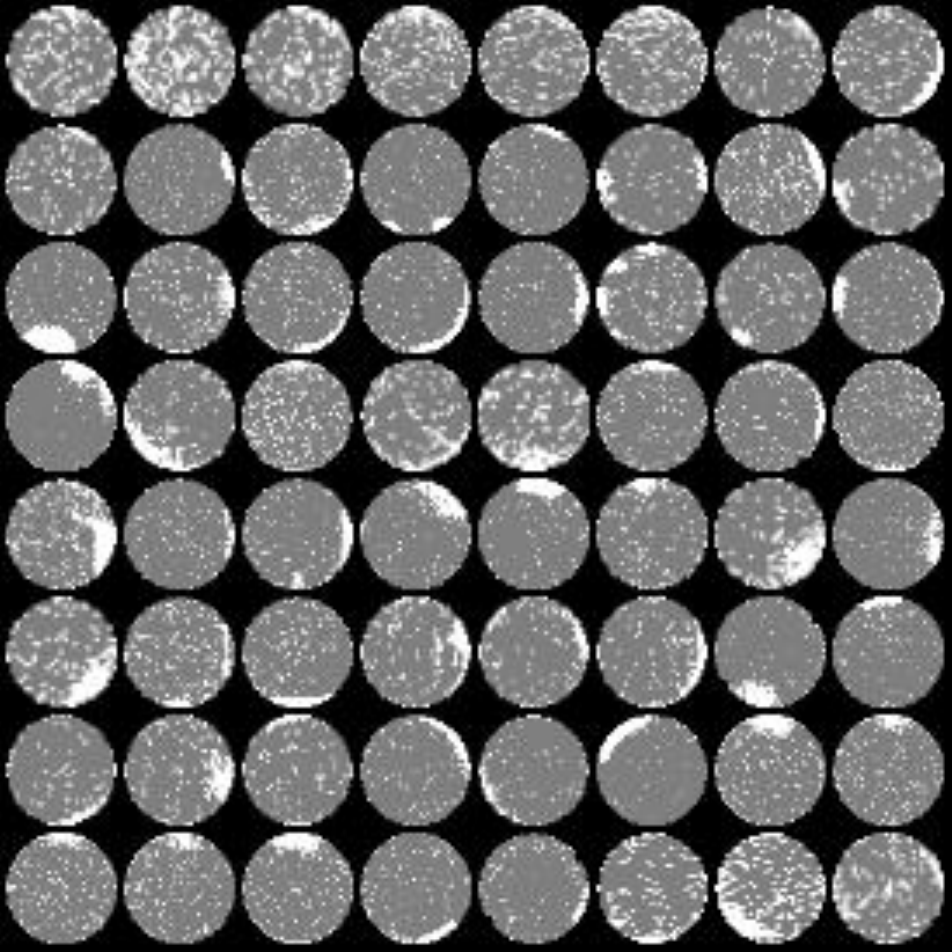}
                \\
                \includegraphics[width=.31\linewidth]{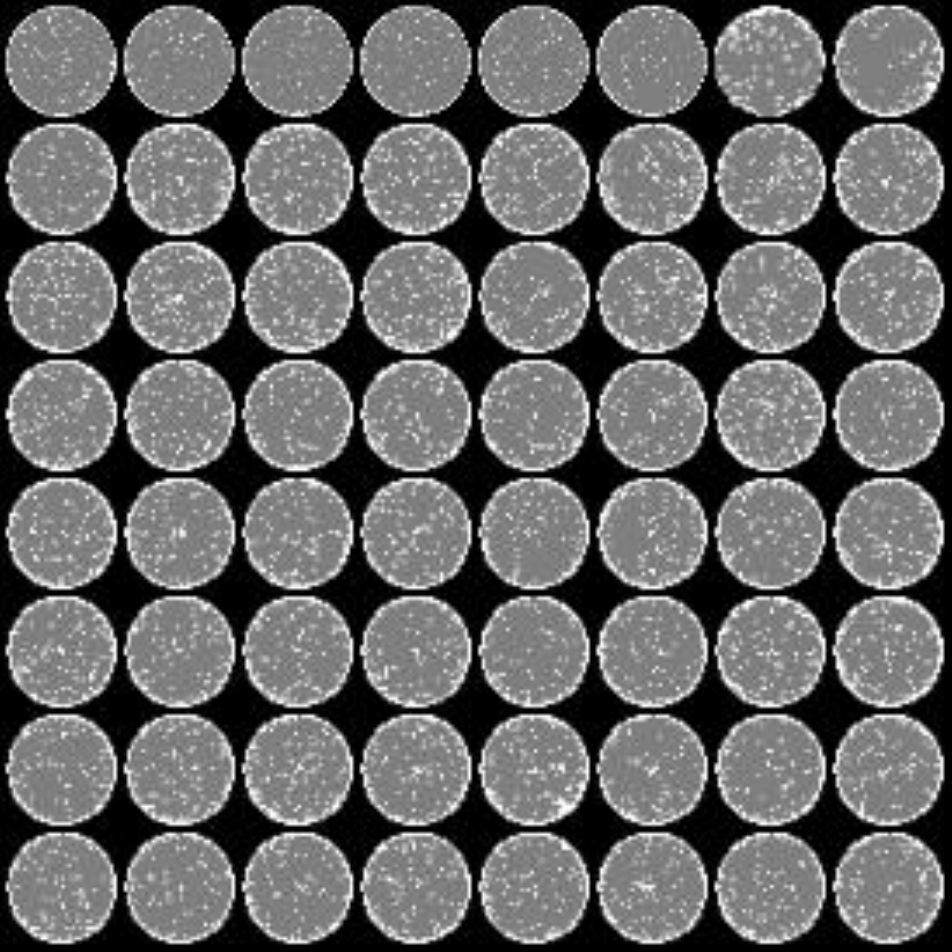}
                \includegraphics[width=.31\linewidth]{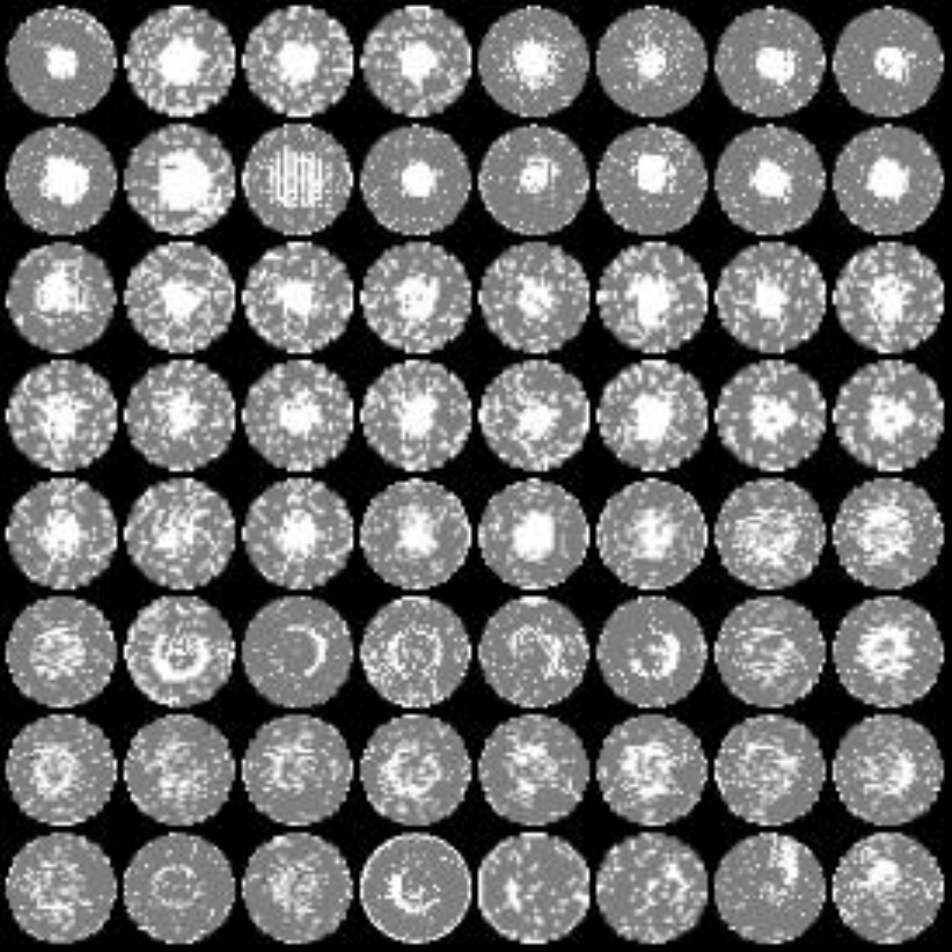}
                \includegraphics[width=.31\linewidth]{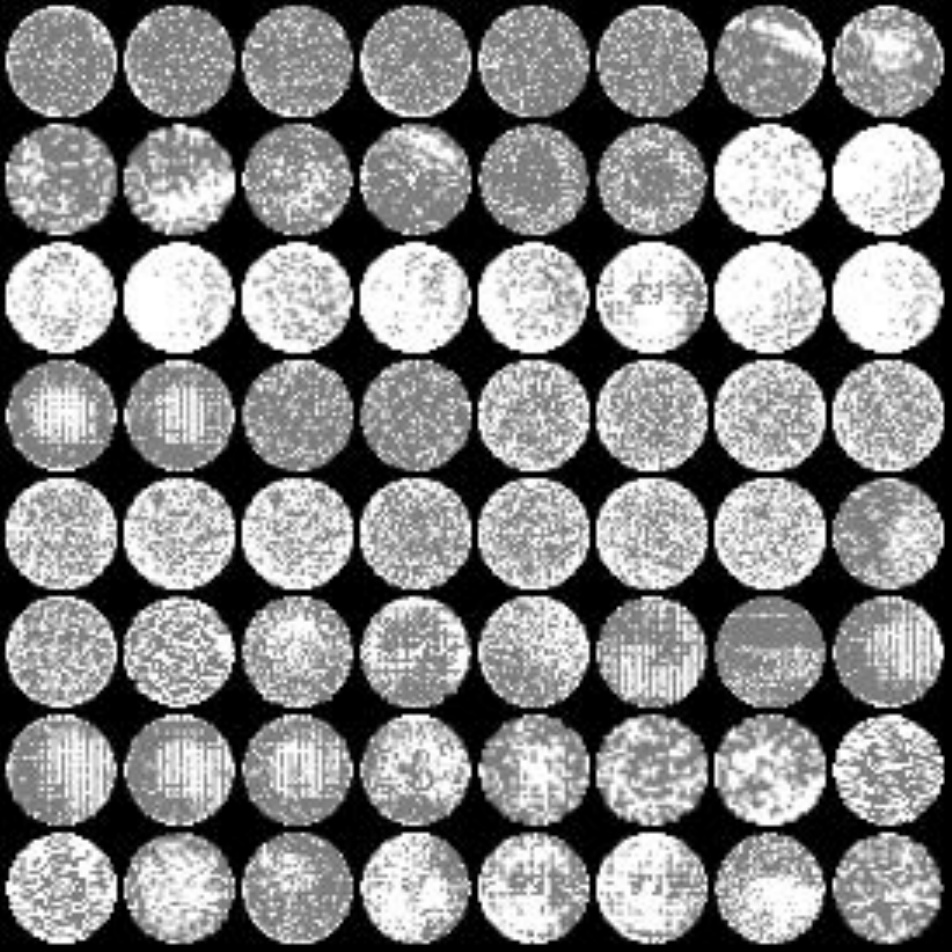}
                \\
                \includegraphics[width=.31\linewidth]{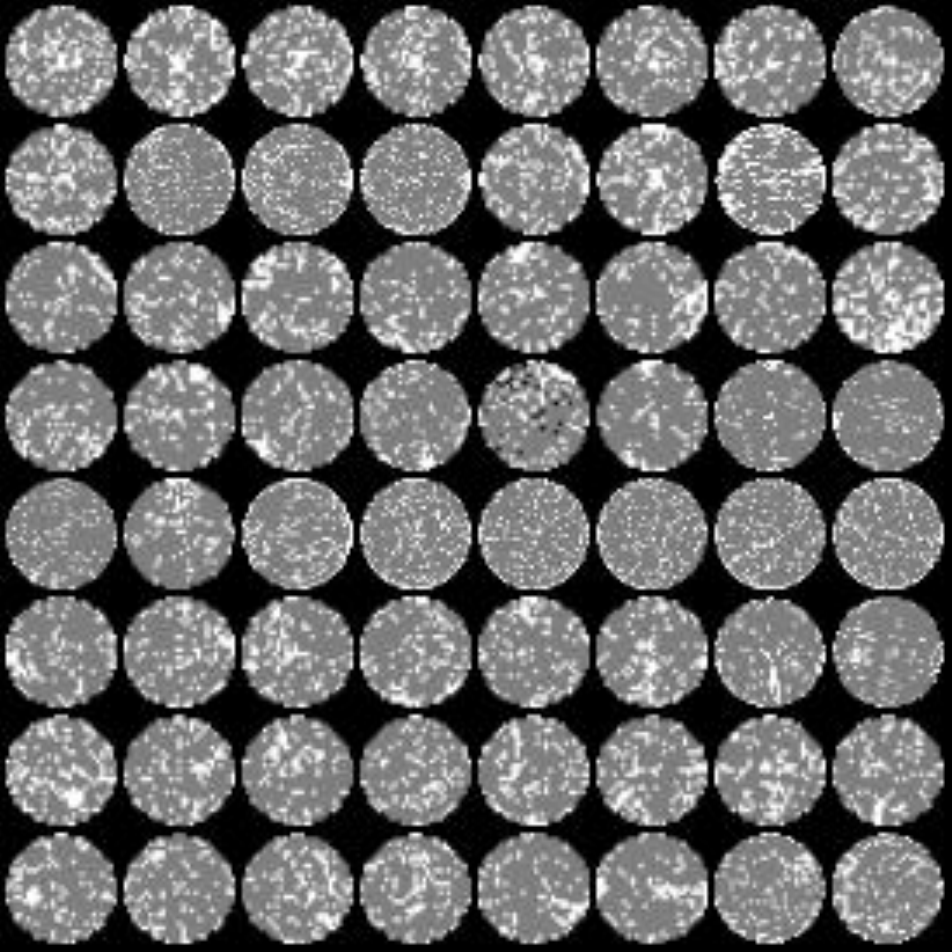}
                \includegraphics[width=.31\linewidth]{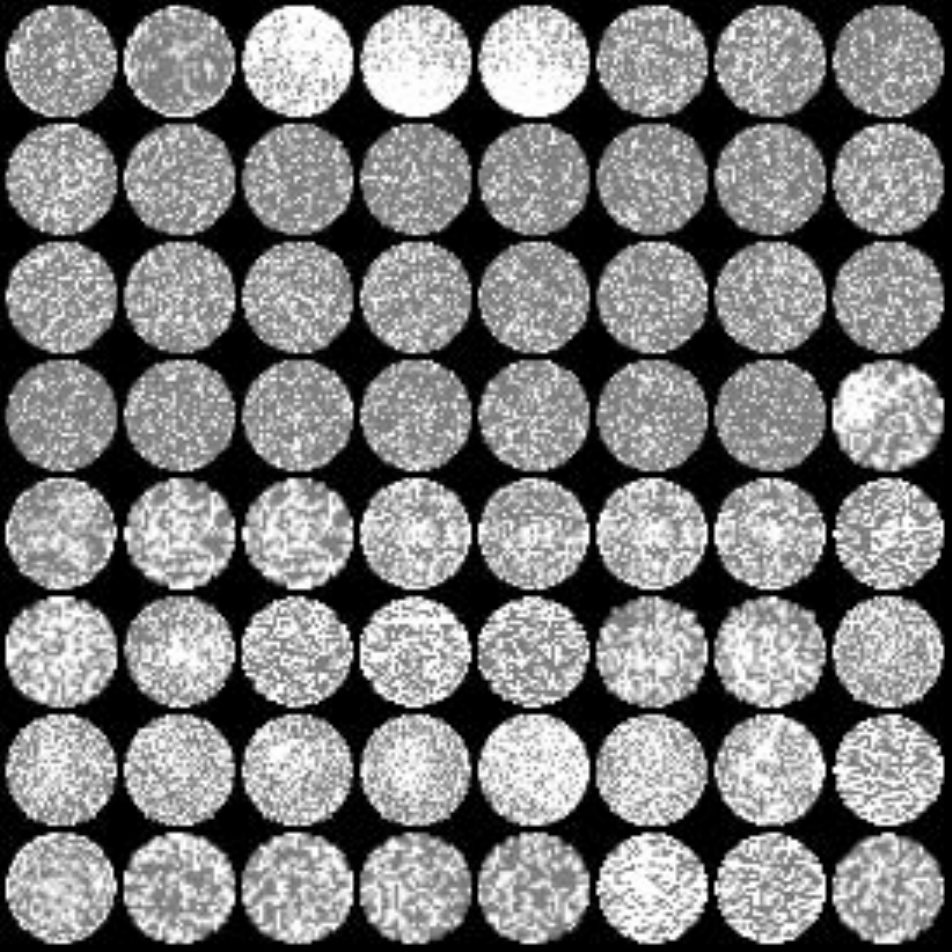}
                \includegraphics[width=.31\linewidth]{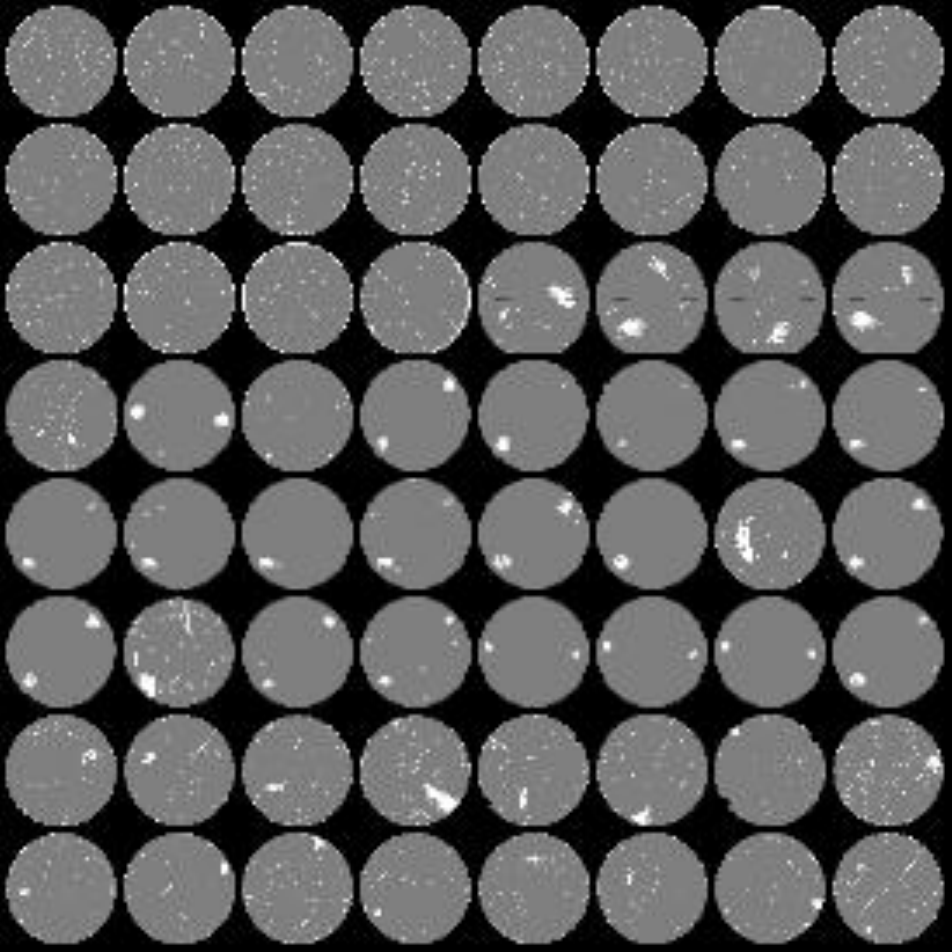}
                \vspace{-5mm}
        \caption{Visualization of WM811k clustering}
        \label{visualization of wm811k clustering}
% \end{figure}%
\label{Clustering visualization}
\end{figure}

\begin{figure}[h!]
                \centering
                \includegraphics[width=.31\linewidth]{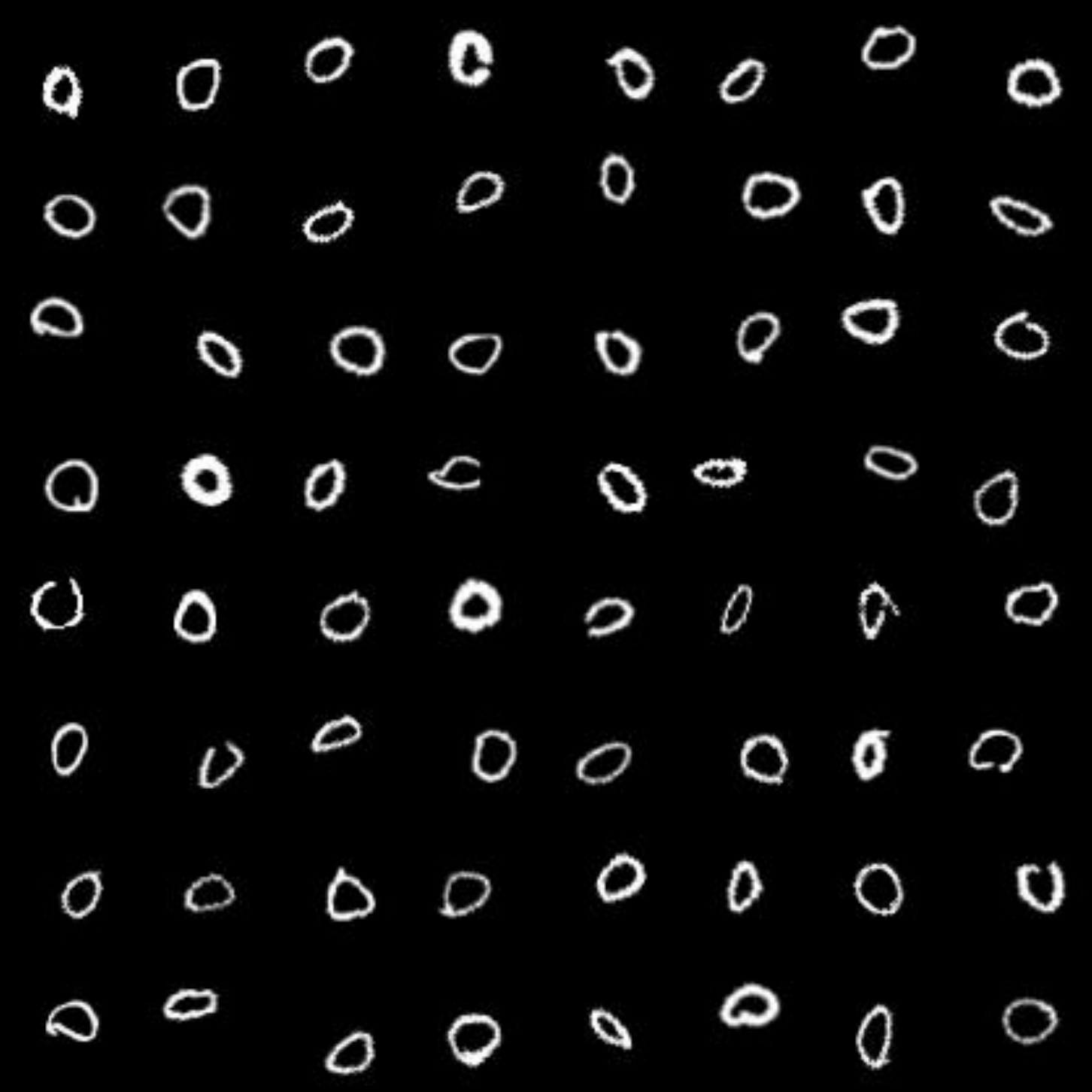}
                \includegraphics[width=.31\linewidth]{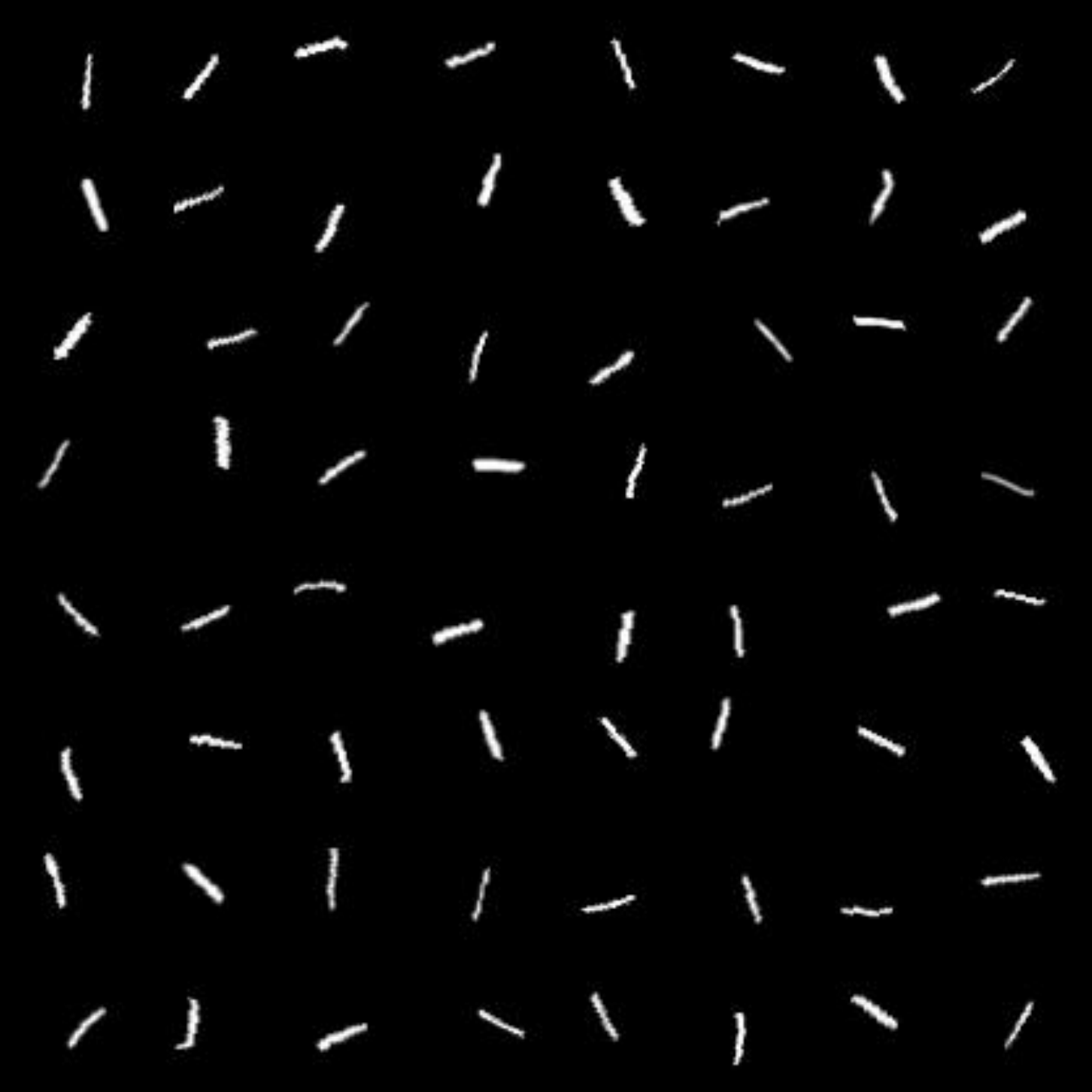}
                \includegraphics[width=.31\linewidth]{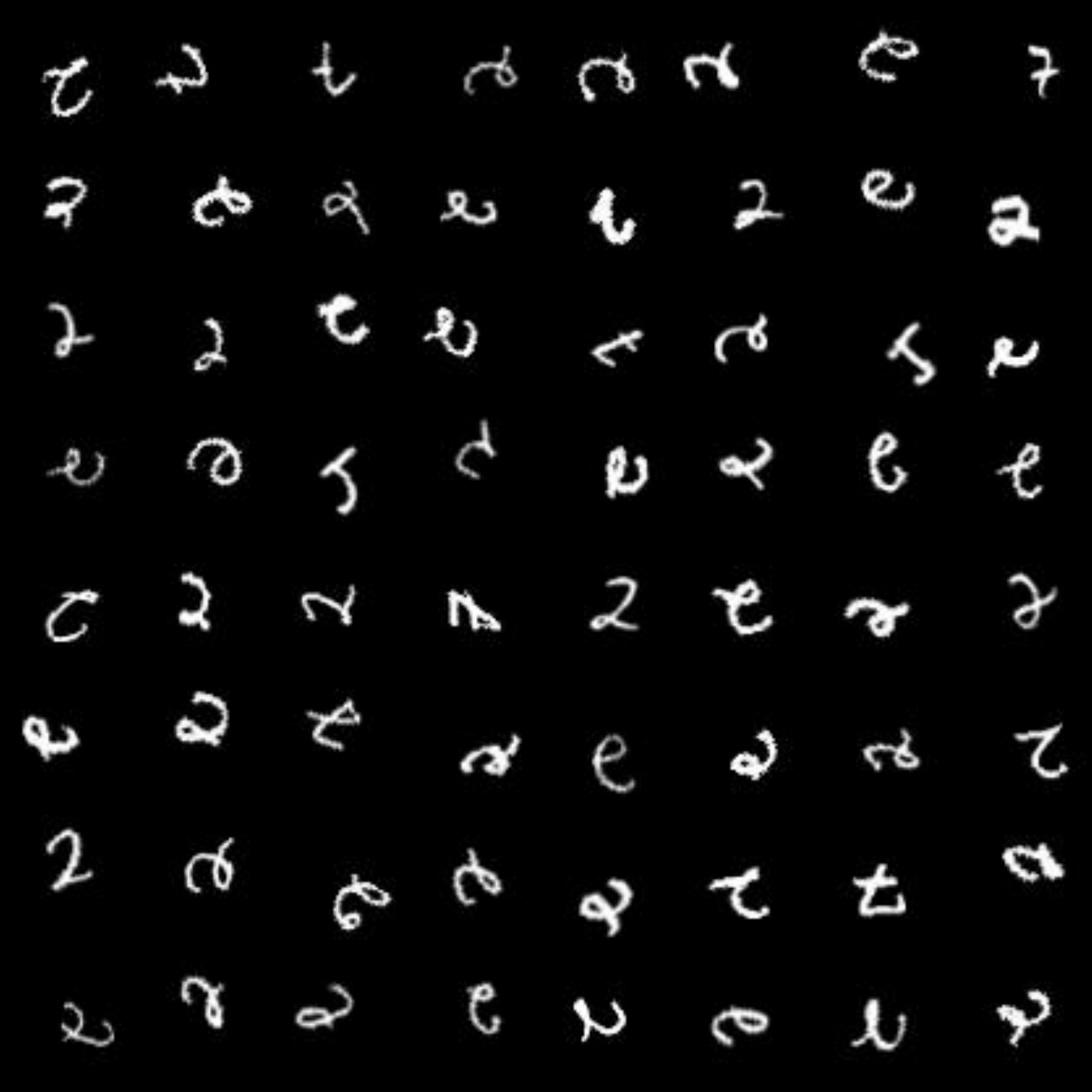}
                \\
                \includegraphics[width=.31\linewidth]{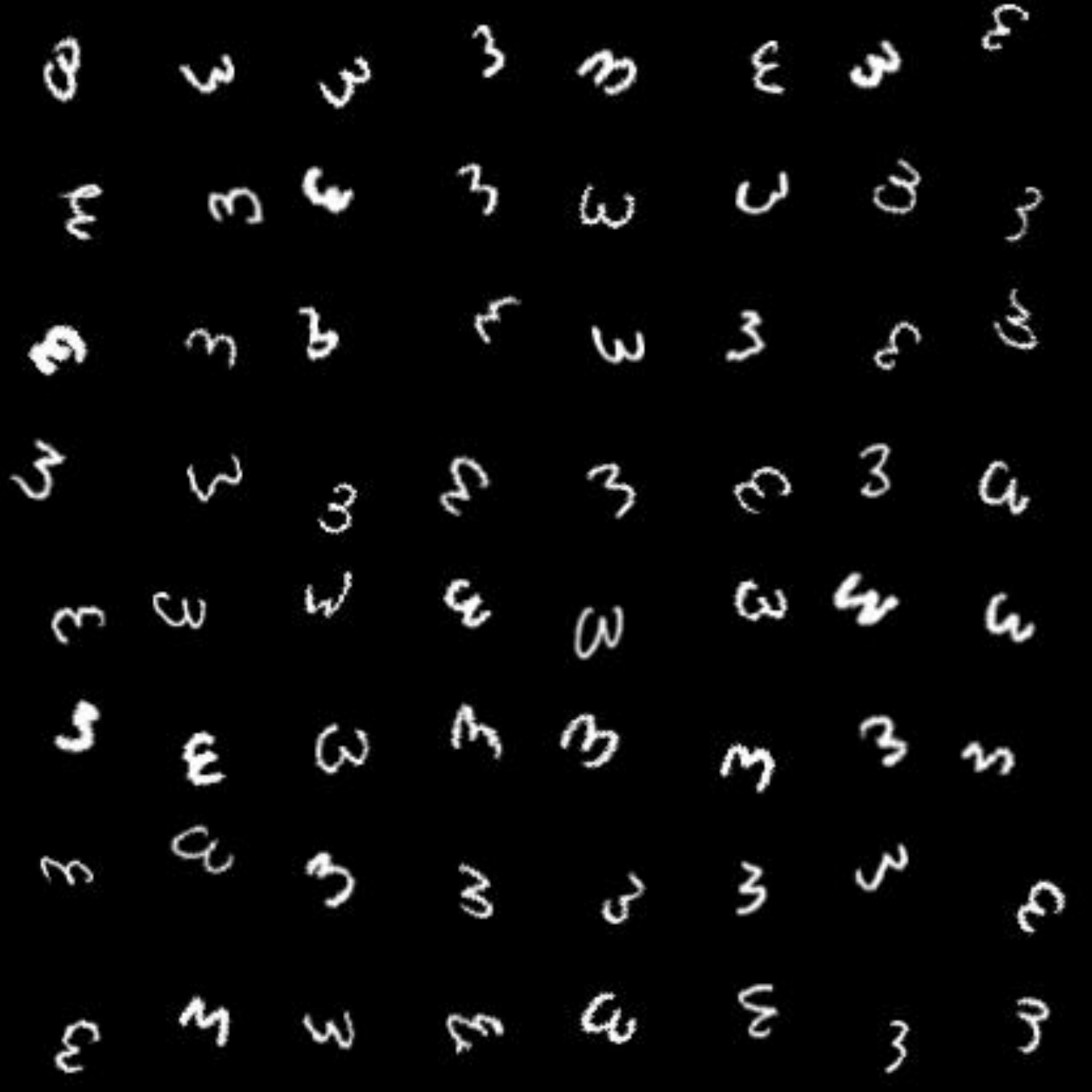}
                \includegraphics[width=.31\linewidth]{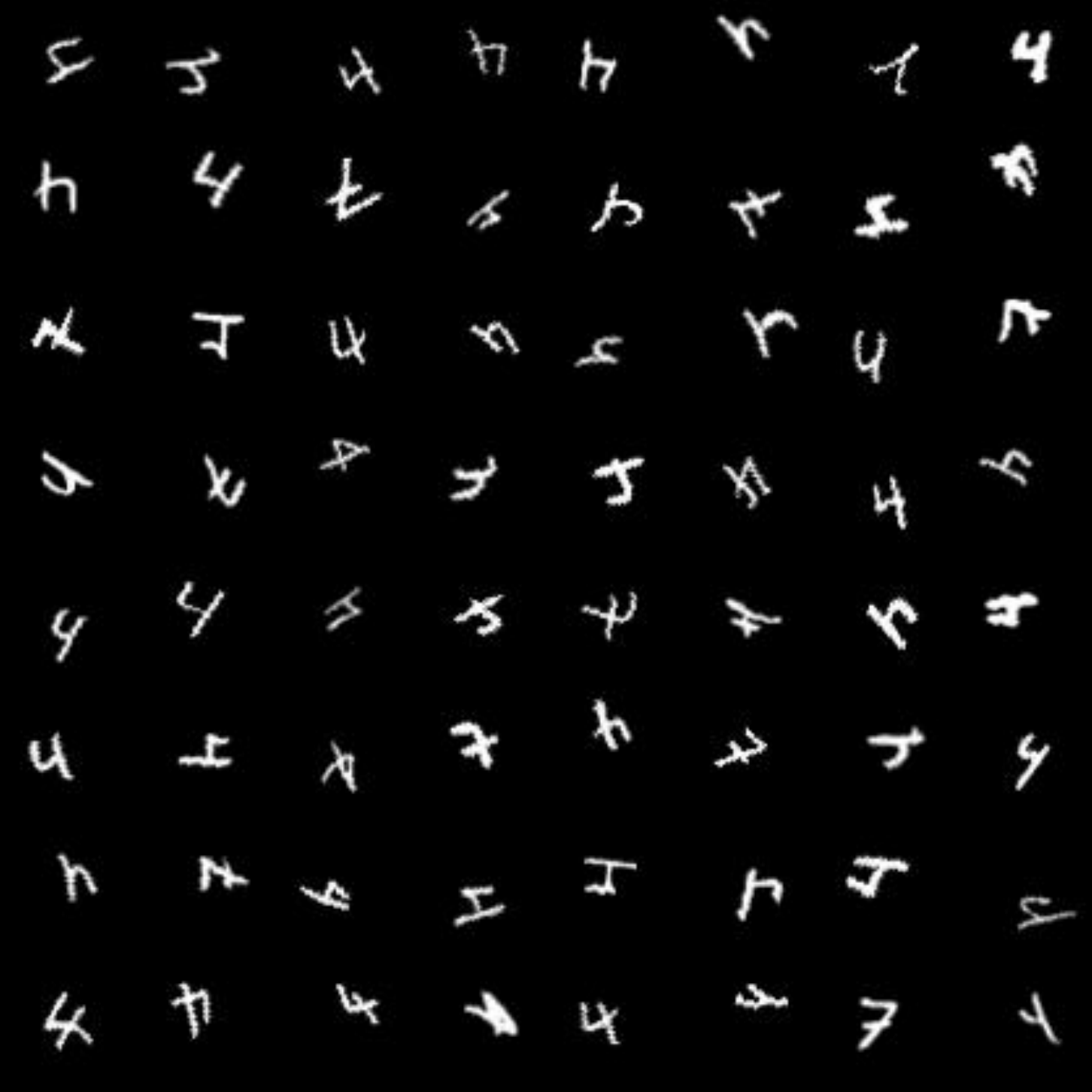}
                \includegraphics[width=.31\linewidth]{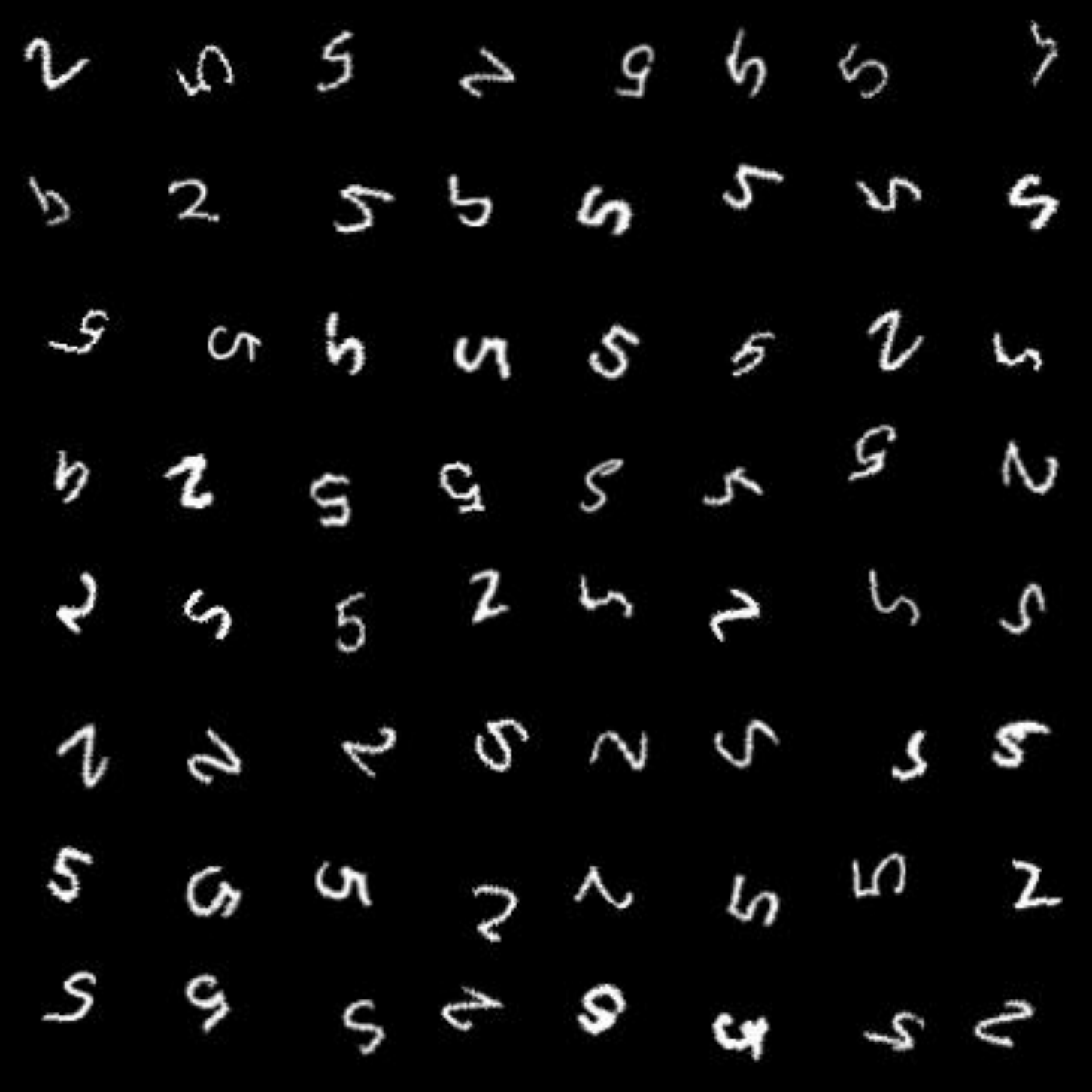}
                \\
                \includegraphics[width=.31\linewidth]{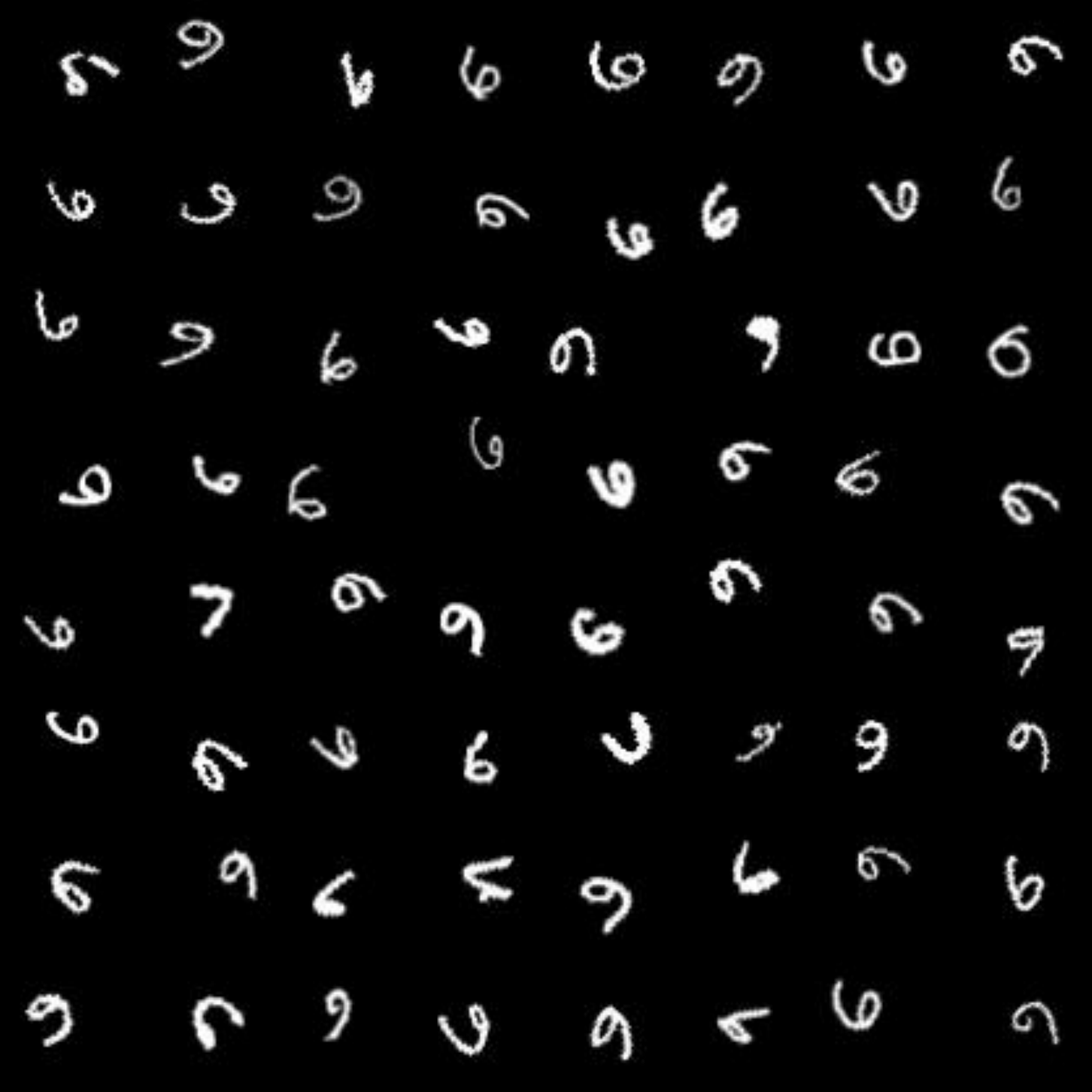}
                \includegraphics[width=.31\linewidth]{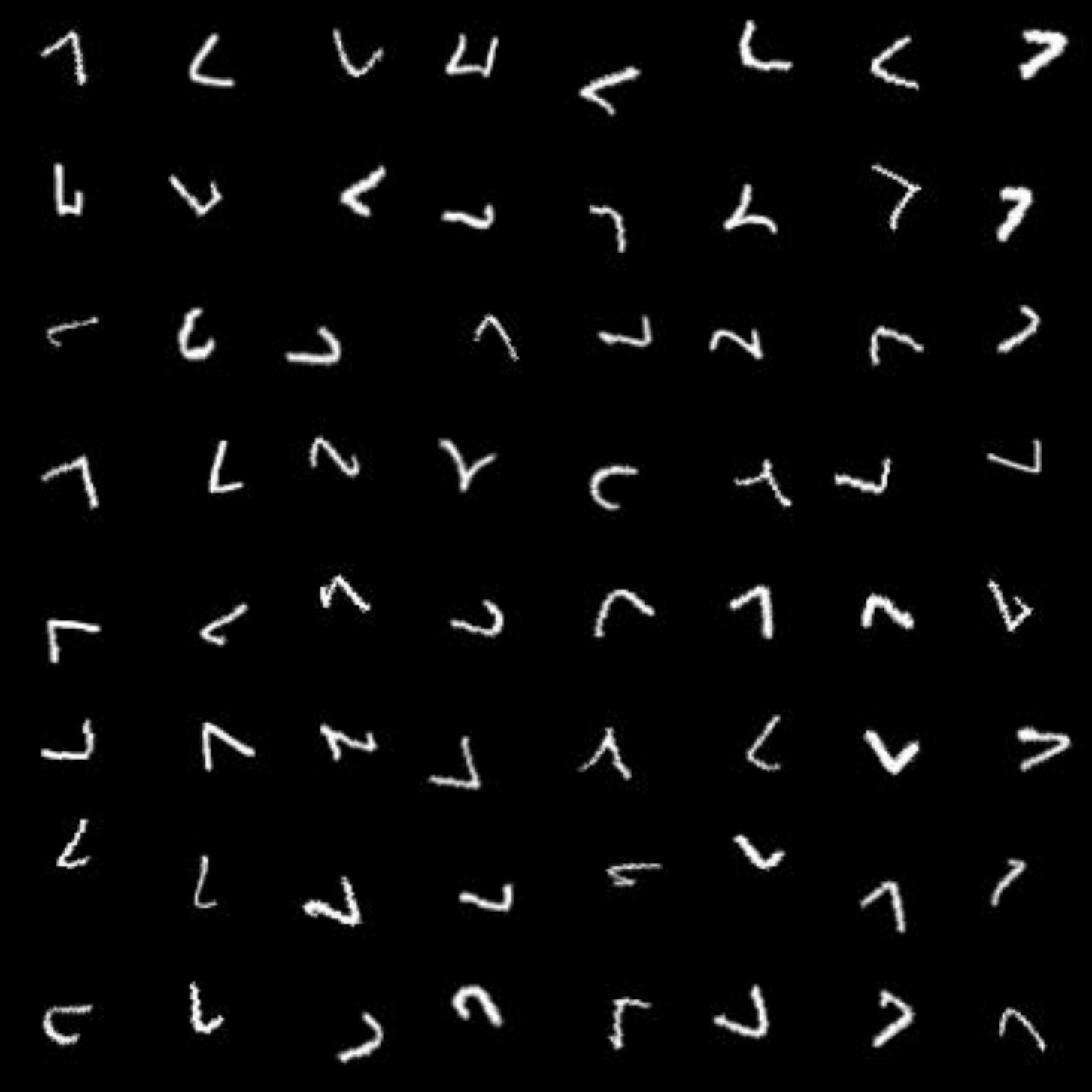}
                \includegraphics[width=.31\linewidth]{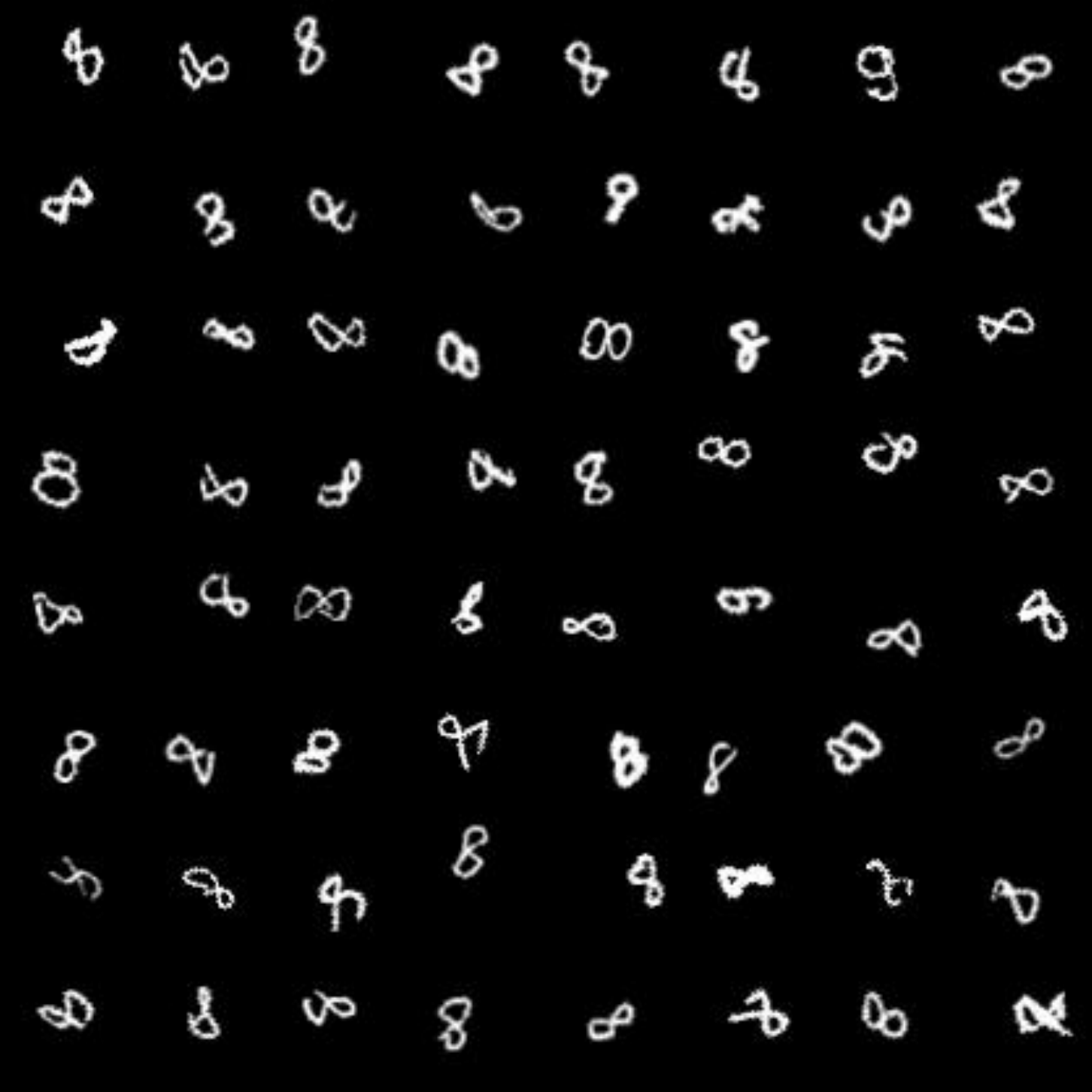}
                \\
                \includegraphics[width=.31\linewidth]{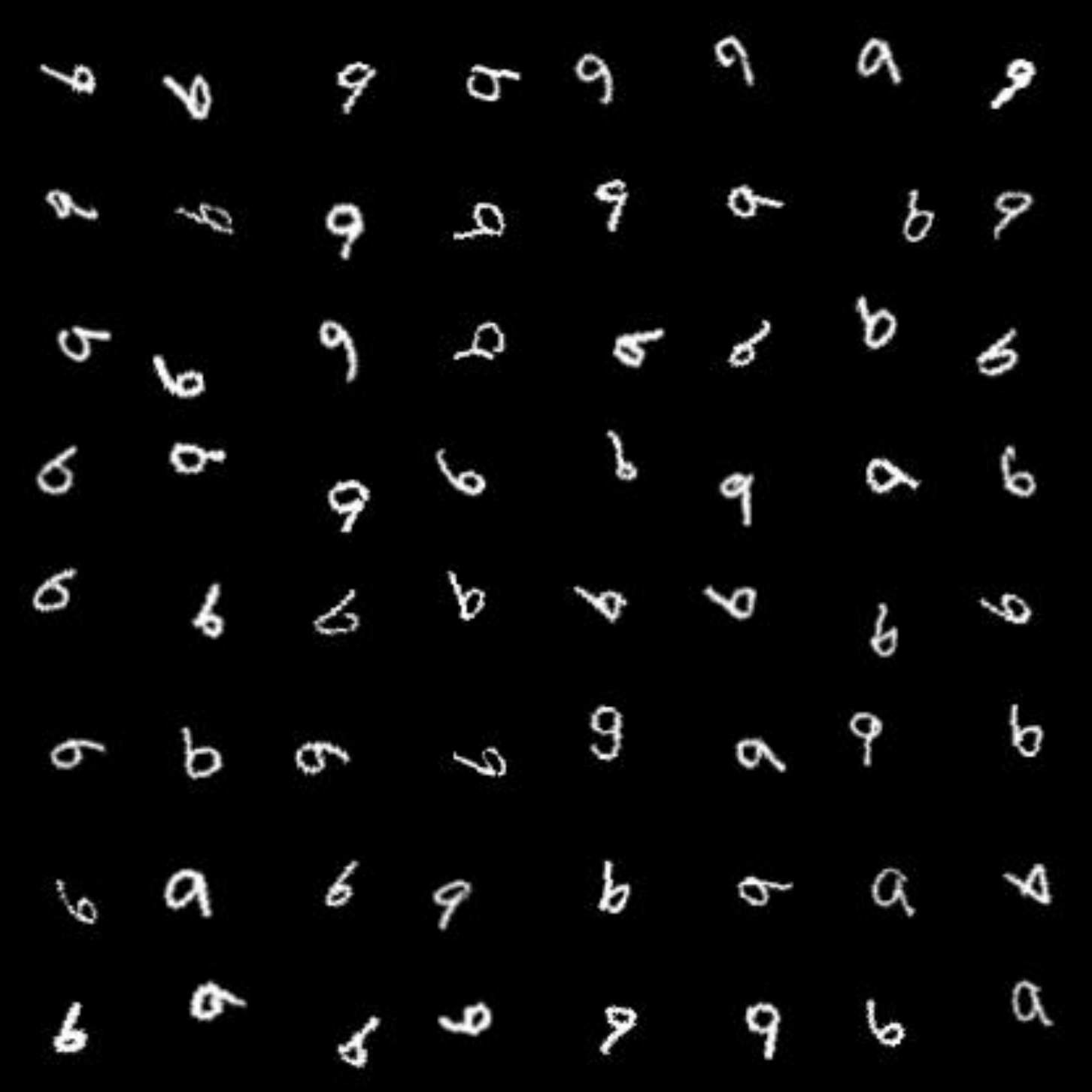}
                \vspace{-5mm}
                \caption{Visualization of MNIST(U) clustering}
                \label{visualization of mnist clustering}
\end{figure}

In this section, we show image samples from each cluster to visualize the clustering performance of IRL-INR + SCAN on WM811k and MNIST(U). Each $8\times 8$ imageset in Figure \ref{visualization of wm811k clustering} and \ref{visualization of mnist clustering} are sampled from same cluster.

\end{document}